%% file: main.tex
\documentclass{article}

 \usepackage[main, final]{neurips_2026}

\usepackage[utf8]{inputenc} 
\usepackage[T1]{fontenc}    
\usepackage{hyperref}       
\usepackage{url}            
\usepackage{booktabs}       
\usepackage{amsfonts}       
\usepackage{nicefrac}       
\usepackage{microtype}      
\usepackage{xcolor}         

\usepackage{graphicx}
\usepackage{subfigure}
\usepackage{multirow}
\usepackage{xcolor} 
\usepackage{booktabs} 
\usepackage{makecell}
\usepackage{textcomp}   
\usepackage{listings}
\usepackage[table]{xcolor}
\usepackage{colortbl}

\usepackage{amsmath}
\usepackage{amssymb}
\usepackage{mathtools}
\usepackage{amsthm}
\usepackage{wrapfig}
\usepackage{threeparttable}
\usepackage{adjustbox}
\usepackage{pifont}
\usepackage{bbm}
\usepackage{longtable}

\usepackage{listings}
\usepackage{shellesc}

\newif\ifuseminted
\usemintedfalse 

\ifuseminted
  \usepackage{minted}
\else
  \usepackage{fvextra}
  \newenvironment{minted}[2][]{%
    \VerbatimEnvironment
    \begin{Verbatim}[breaklines,fontsize=\small]%
  }{%
    \end{Verbatim}%
  }
  
  \newcommand{\inputminted}[3][]{\VerbatimInput[breaklines,fontsize=\small]{#3}}
\fi
\usepackage{tcolorbox}

 \newcommand{\camred}[1]{{\color{black}#1}}

\newcommand{\method}{\texttt{CELM}\xspace}
\newcommand{\methodscc}{\texttt{CELM-SCC}\xspace}

\input{packs}

\input{define}

\title{Neural Signals Generate Clinical Notes in the Wild}

\author{Jathurshan Pradeepkumar$^{1}$
\quad  Zheng Chen$^{2}$\thanks{Corresponding authors}
\quad Jimeng Sun$^{1*}$ \\ 
$^{1}$University of Illinois Urbana-Champaign \quad $^{2}$SANKEN, Osaka University\\
\texttt{\{jp65,jimeng\}@illinois.edu}\quad \texttt{\{chenz\}@sanken.osaka-u.ac.jp }
}

\begin{document}

\maketitle

\input{000abstract}

\section{Introduction}
    \label{sec:intro}
    \input{010intro}
\section{Related Work}
    \label{sec:related}
    \input{020related}

\section{EEG-Report Benchmark Construction}
    \label{sec:benchmark_constuction}

\input{031benchmark}
\section{CELM: Clinical EEG Language Model}
    \label{sec:clinical_elm}
    \input{032method}
\section{Experiments and Results}
    \label{sec:exp}

\input{040experiments}

\vspace{-0.1cm}
\section{Conclusion}
\vspace{-0.1cm}
    \label{sec:conclusion}
    \input{050conclusion}

\newpage

\section*{\camred{Acknowledgments, Disclosures and Ethics Statement}}
\label{sec:ethics}

\camred{
\paragraph{Funding Transparency Statement.}This project has been funded by the Jump ARCHES endowment through the Health Care Engineering Systems Center. The authors declare no competing interests.

\paragraph{Ethics Statement.}
This retrospective study uses the de-identified Harvard EEG Database v4.1~\cite{sun2025harvard,zafar2025harvard}, accessed via BDSP with credentialed access, CITI training, and a Data Use Agreement under the BDSP Credentialed Health Data License 1.5.0. All data used in this study were de-identified by the dataset providers to protect participants’ protected health information (PHI) and the authors had no access to identifiable information and no interaction with patients. All data processing and experiments were performed using locally hosted LLMs, without transmitting data to external APIs. We release only the benchmark construction pipeline and modeling code; clinical data and derived reports are not shared, and access to model checkpoints will be gated. \method{} and all released artifacts are intended for research use only and not for clinical deployment, consistent with the dataset’s data-use terms. The expert evaluation involved research collaborators and peers who voluntarily rated de-identified model outputs without compensation.

}



\bibliographystyle{unsrt}
\bibliography{main}


\newpage

\appendix
\section*{Appendix}
\DoToC
    \label{sec:appendix}
    \input{100appendix}




\newpage
\input{checklist.tex}

\end{document}

%% file: packs.tex
\usepackage{color}
\usepackage{hyperref}
\def\equationautorefname~#1\null{Equation~(#1)\null}
\usepackage{breakurl}
\usepackage{bm}
\usepackage{soul}
\usepackage{xspace}
\usepackage{algorithmic}
\usepackage{algorithm}
\usepackage{mdwlist}    
\usepackage{paralist} 

\usepackage{adjustbox}
\usepackage{array}
\usepackage{booktabs}
\usepackage{multirow}
\usepackage{enumitem}
\usepackage{subcaption}

\usepackage{array,graphicx}
\usepackage{booktabs}
\usepackage{color}

\usepackage{amsmath}
\definecolor{lightgray}{gray}{0.85}
\usepackage{multirow} 

\def\BibTeX{{\rm B\kern-.05em{\sc i\kern-.025em b}\kern-.08em
    T\kern-.1667em\lower.7ex\hbox{E}\kern-.125emX}}

\def\BibTeX{{\rm B\kern-.05em{\sc i\kern-.025em b}\kern-.08em
    T\kern-.1667em\lower.7ex\hbox{E}\kern-.125emX}}

%% file: define.tex
\newcommand{\hide}[1]{}

\newcommand{\vswd}{\vspace{0.3em}}
\newcommand{\bit}{\vswd\begin{itemize*}}
\newcommand{\eit}{\end{itemize*}\vswd}
\newcommand{\ben}{\vswd\begin{enumerate*}}
\newcommand{\een}{\end{enumerate*}\vswd}
\newcommand{\bea}{\vspace{-0.0em}\begin{eqnarray}}
\newcommand{\eea}{\end{eqnarray}\vspace{-0.0em}}
\newcommand{\beq}{\vspace{-0.0em}\begin{equation}}
\newcommand{\eeq}{\end{equation}\vspace{-0.0em}}

\renewcommand{\bit}{\vswd\begin{compactitem}}
\renewcommand{\eit}{\end{compactitem}\vswd}
\renewcommand{\ben}{\vswd\begin{compactenum}}
\renewcommand{\een}{\end{compactenum}\vswd}

\definecolor{highlightblue}{HTML}{8DA0CB}
\definecolor{highlightorange}{HTML}{FC8D62}
\definecolor{promptblue}{HTML}{516288}

\usepackage{titletoc}
\newcommand\DoToC{%
  \startcontents
  \printcontents{}{1}{\textbf{Contents}\vskip3pt\hrule\vskip5pt}
  \vskip3pt\hrule\vskip5pt
}


%% file: 000abstract.tex
\begin{abstract}



Generating clinical reports that summarize abnormal patterns, diagnostic findings, and clinical interpretations from long-term EEG recordings remains labor-intensive.
We present \method, the first clinical EEG-to-Language foundation model capable of summarizing long-duration, variable-length EEG recordings and performing end-to-end clinical report generation at multiple scales.
\method integrates pretrained EEG foundation models with language models to enable scalable multimodal learning.
We curate a large-scale clinical EEG dataset containing 9,922 reports paired with approximately 11,000 hours of EEG recordings from 9,048 patients to train \method, and \camred{release the benchmark pipeline, with automated report-structuring, to facilitate future research.}
Experimental results show that \method consistently outperforms existing methods across all evaluation settings. 
Importantly, we further conduct human evaluation with clinical experts, demonstrating that \method generates reports that are more clinically coherent, diagnostically reliable, and better aligned with expert interpretation.
\camred{We release our model code and benchmark construction pipeline at \url{https://github.com/Jathurshan0330/CELM}.}



\end{abstract}

%% file: 010intro.tex
Electroencephalography (EEG) records long-term, real-time neuronal activity with millisecond-level temporal resolution and is widely used for neurological diagnosis \cite{yang2023biot,kotoge2025evobrain,ICLR2026_aebec805}.
However, writing clinical reports to summarize diagnostic findings over EEG recordings remains a labor-intensive process.
Typically, neurologists visually inspect EEG signals to assess brain activity and then compose clinical reports that document various phenotypes and their clinical interpretations~\citep{tatum2016american}.
This task requires substantial domain knowledge and extensive clinical experience, which incur long-term training costs and require continual updates as new EEG phenotypes or waveforms are identified~\cite{biswal2019eegtotext}.

\textbf{Limitations of prior works.}
Although several deep learning approaches have been proposed for EEG report generation~\cite{biswal2019eegtotext,leereportgene2023,NeuroLex2025}, several limitations remain. 
Conceptually, many existing methods formulate report generation as a phenotype classification task followed by text decoding.
Such pipelines fail to model EEG-to-report generation in an end-to-end manner, leading to an objective misalignment between \textit{classification} and \textit{generation}.
Methodologically, they rely on short EEG contexts (spanning only several to tens of seconds) and fixed-context templates that construct checklist-style mappings between EEG phenotypes and predefined clinical notes.
They fail to model long-term reasoning required for interpreting long-duration EEG data, further preventing multi-granularity report generation.
Practically, due to the limited resource, these methods are developed as task-specific models, each tailored to a narrow reporting objective~\cite{leereportgene2023}.
However, clinical practice requires multi-level reporting, where neurologists routinely construct overall EEG summaries and section-wise outputs, including impressions, interpretations, and event or seizure annotations.

\begin{figure}[t]
    \centering
     \includegraphics[width=\textwidth]{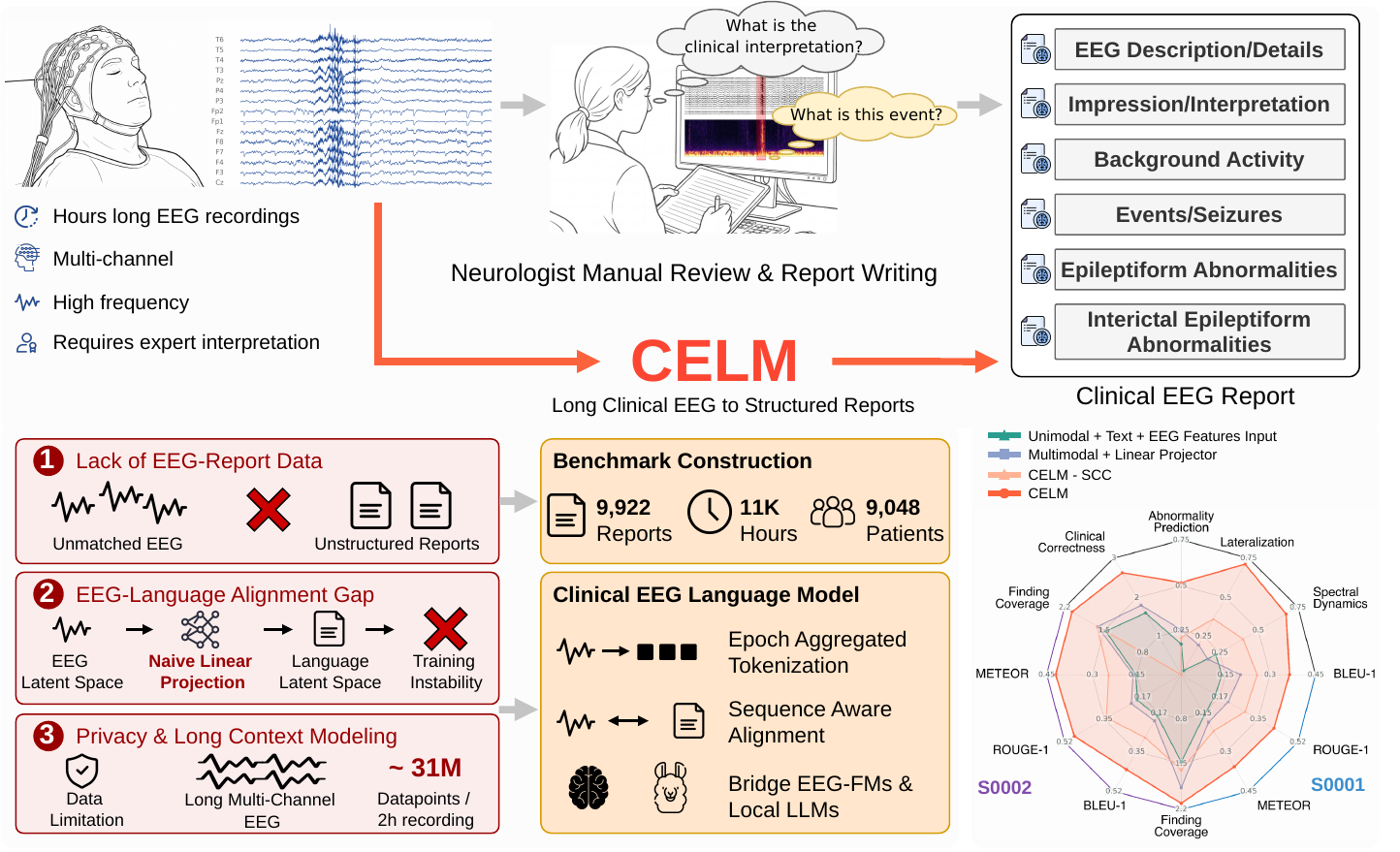}
  \caption{
Overview of \method, a clinical EEG-to-language foundation model (ELM) for translating long-duration EEG recordings into structured clinical reports.
\method addresses key challenges in EEG report generation, including limited EEG-report data, EEG-language alignment, and long-context EEG modeling.
The framework constructs a large-scale EEG-report benchmark and enables direct generation of structured clinical reports from raw EEG recordings.
}
    \label{fig:story_fig}
    \vspace{-0.5cm}
\end{figure}

\textbf{Present work.}
To fill the gap, we present \method, to the best of our knowledge, the first \textbf{C}linical EEG-to-language foundation model (\textbf{ELM}) that unlocks meaningful interpretation of raw clinical EEG signals and enable report generation at multiple scales.
Our motivation stems from recent advances in multimodal foundation models that enable text generation from non-text modalities (e.g., images and speech) \cite{VLMsurvey,xuVLM}.
\method hence shifts the paradigm from conventional EEG classification toward language-centric EEG understanding and describing, with three key innovations:
\ding{182} A report structuring pipeline that automatically decomposes clinical notes and generates hierarchical structured reports, spanning EEG description, background activity, epileptiform abnormalities, events/seizures, and impressions/interpretations.
\ding{183} A novel pretrained foundation model that processes EEG recordings of up to \textit{three hours}, learns multi-grained representations from local patterns to global temporal contexts, and translates representations into text for end-to-end report generation.
This is achieved by three core components: (i) epoch-aggregated tokenization that produces compact representations from variable-length recordings, (ii) sequence-aware alignment that captures inter-epoch temporal dependencies, and (iii) prompt fusion for conditional report generation. 
Importantly, our model design harnesses the representation capabilities of pretrained EEG foundation models and LLMs while enabling flexible generation across multiple clinical settings.
\ding{184} A large-scale EEG-report benchmark curated from the largest clinical EEG corpora to date, the Harvard Electroencephalography Database~\citep{sun2025harvard,zafar2025harvard} (HEEDB), comprising  \camred{$9{,}922$ clinical reports paired with over $11{,}000$ hours of EEG recordings from $9{,}048$ patients, enabling \method training and facilitating future model development.}


To rigorously evaluate \method, we benchmark it against state-of-the-art (SOTA) methods across five real-world report drafting tasks. 
Moreover, we conduct \textbf{human evaluation} with six clinical experts to assess the quality and clinical reliability of the generated reports. 
Extensive experiments demonstrate that \method not only achieves significant improvements across quantitative metrics, but also produces reports that are better aligned with human interpretation.
Our contributions are as follows:
\vspace{-0.2cm}

\begin{itemize}[left=0pt]
    \item We introduce a \textit{novel EEG-to-text foundation model} capable of learning representations from long-duration EEG recordings (up to three hours) and directly generating clinical reports.
    \vspace{-0.1cm}
    \item \camred{We propose \textit{an automated report structuring pipeline} for raw EEG-text clinical data, which integrates seamlessly with the HEEDB to produce our benchmark dataset.}
    \vspace{-0.1cm}
    \item Beyond benchmarking \method, we also conduct \textit{human evaluation} to assess whether the generated reports are more diagnostically reliable.
    Results show that \method is statistically significantly more clinically coherent than those produced by prior methods and general language models.

\end{itemize}

%% file: 020related.tex

\paragraph{EEG-to-Language Modeling. }
Existing work at the intersection of EEG and natural language can be broadly categorized into two paradigms: (1) EEG-to-language decoding and (2) text-enhanced EEG representation learning. 
The goal of EEG-to-language decoding is to reconstruct textual content from concurrent EEG recordings of subjects reading or imagining speech~\citep{herff2015brain,wang2022open}.
This line of work spans both invasive approaches using electrocorticography (ECoG)~\citep{makin2020machine,willett2023high}  and non-invasive methods based on scalp EEG~\citep{wang2022open,duan2023dewave,zhou2024belt}. 
However, these methods assume precise EEG–text alignment, whereas clinical EEG consists of heterogeneous events embedded in long, continuous recordings spanning hours to days~\citep{sun2025harvard,zafar2025harvard}.
The second paradigm leverages clinical notes as auxiliary supervision to enhance EEG representation learning, rather than decoding language directly from neural activity.
Inspired by vision–language pretraining frameworks~\citep{radford2021learning}, a recent work ~\citep{ndir2025eeg} aligns EEG data with clinical report text in a shared feature space.
However, it remains focused on discriminative objectives and does not support automated clinical report generation.
\\
\textbf{Clinical EEG Report Generation. }
An early attempt at EEG report generation, EEGtoText~\citep{biswal2019eegtotext}, proposed a two-stage pipeline that first classifies EEG phenotypes, and then generates report text via a text decoder conditioned on the classified labels. 
However, this relies on intermediate phenotyping as a bottleneck, limiting the capabilities to capture nuanced clinical findings beyond predefined categories. 
While some attempts jointly learn EEG encoders and text decoders~\cite{leereportgene2023}, they rely on fixed segmentation and template-based generation~\cite{NeuroLex2025}.
As a result, these methods fall short of end-to-end clinical report generation from long-duration EEG recordings.
We address the above challenges by introducing the first family of EEG–Language foundation models.
\\
\textbf{EEG Foundation Models. } 
Recent years have witnessed rapid advances in EEG foundation models.
BENDR~\citep{kostas2021bendr}, BIOT~\citep{yang2023biot}, LaBraM~\citep{jiang2024large}, TFM-Tokenizer~\citep{pradeepkumar2025single}, EEGPT~\citep{wang2024eegpt}, LUNA~\citep{doner2025luna}, REVE~\citep{ouahidi2025reve}, and CBraMod~\citep{wang2024cbramod} introduce increasingly scalable tokenization and representation learning frameworks that achieve strong transfer on diverse EEG tasks. 
These models are predominantly encoder-only and are optimized for classification tasks, leaving generative objectives largely unexplored.
Our \method is designed for clinical EEG report generation, and notably, it is fully compatible with existing foundation models, which can be directly leveraged as EEG encoders.

%% file: 031benchmark.tex
\begin{figure}[t]
    \centering
     \includegraphics[width=\textwidth]{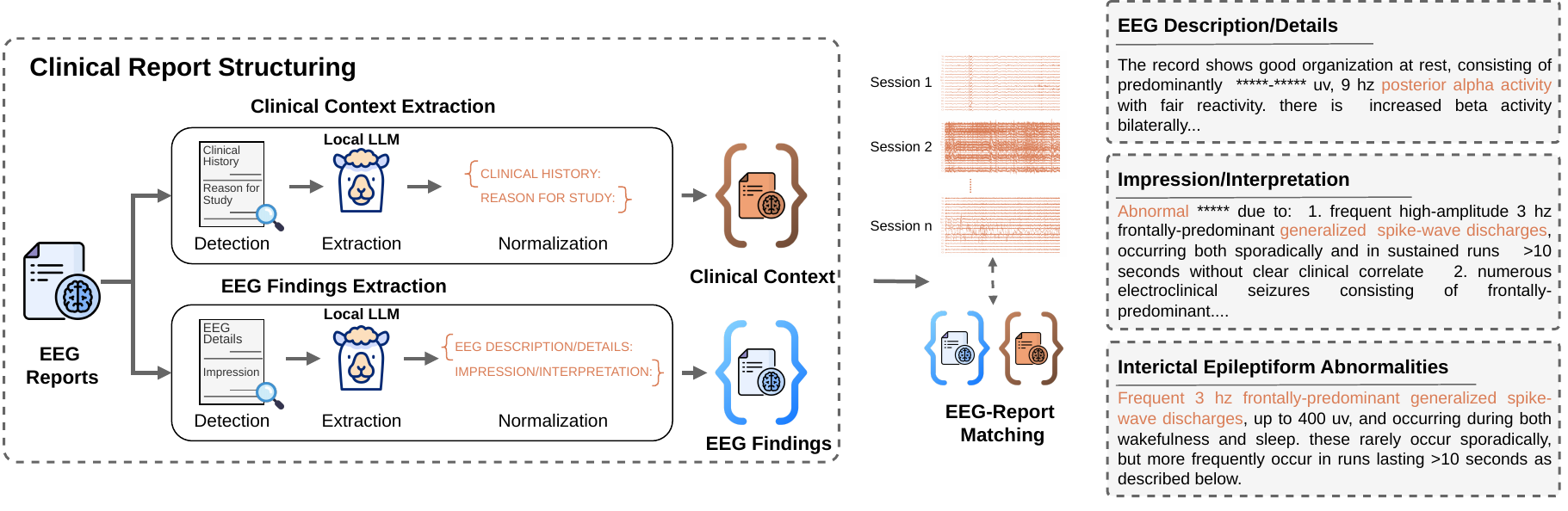}
     \vspace{-0.6cm}
    \caption{EEG–Report benchmark construction pipeline, including clinical report structuring (Section~\ref{subsection:3.1}), matching reports to EEG sessions, and examples of standardized report sections. 
    }
    \label{fig:benchmark}
    \vspace{-0.3cm}
\end{figure}

Currently, no publicly available benchmark provides structured EEG–report pairs for training and evaluating EEG-to-text models.
In this work, we leverage two resources from the \emph{Brain Data Science Platform}: 
(1) the Harvard Electroencephalography Database v4.1~\cite{zafar2025harvard,sun2025harvard}, which contains approximately \camred{$12$K} clinical reports paired with about $100$K hours of EEG recordings from $10{,}886$ patients across multiple hospital sites, and 
(2) the Electronic Health Records (EHR) Repository with corresponding unstructured neurology reports.
The EHR repository links reports to one or more EEG sessions via temporal alignment between EDF start times and report timestamps.
Therefore, we construct an EEG–report benchmark using a dedicated processing pipeline (Figure~\ref{fig:benchmark}).

\textbf{Clinical Report Structuring. }
\label{subsection:3.1}
Neurology reports can be broadly decomposed into two types of sections:
(1) \emph{Clinical Context}, which contains information such as patient history and monitoring indications that cannot be inferred from EEG signals, and 
(2) \emph{EEG Findings}, which describe EEG observations and clinical interpretations that are directly inferable from EEG recordings.
A key challenge lies in reliably extracting and structuring these sections without information loss.
Substantial variability in reporting styles and formatting makes manual extraction infeasible at scale.
We therefore propose a three-stage automated pipeline to extract and structure reports.
\begin{itemize}[left=0pt]
    \item \textbf{Section Detection:}
We curate a comprehensive list of candidate section headers under each category and detect their presence using string matching. 
\item \textbf{Section-wise Extraction:}
For each detected section, we employ a local LLM (Meta-Llama-3-8B-Instruct) to extract the corresponding text using simple copy-based prompts. 
This detect-then-segment strategy avoids hallucinations and cross-section contamination that arise when extracting multiple sections jointly.
\item \textbf{Canonical Normalization:}
We standardize the extracted sections into canonical categories, including EEG description/details, impressions/interpretations, background activity, epileptiform abnormalities, and events/seizures.
\end{itemize}

\textbf{EEG Preprocessing. }\label{subsection:3.2}
Matched EEG recordings are preprocessed using a standard pipeline. Signals are first band-pass filtered between 0.1–75 Hz and notch filtered at 60 Hz to remove power-line interference. The recordings are then resampled to 200 Hz to ensure compatibility with existing EEG-only foundation models. Each recording is segmented into non-overlapping 10-second windows, which is the typical temporal context used by current EEG foundation architectures. Finally, EEG channels are standardized to a 22-channel montage following the International 10-20 system. While the dataset contains reports matched to multiple EEG sessions, we filter to samples with single-session matches for training, validation, and evaluation. 

%% file: 032method.tex
This section introduces our \method, consisting of three key components: \emph{Epoch-Aggregated Tokenization}, \emph{Sequence-Aware Alignment}, and \emph{Prompt Fusion and Generation}.
An effective model must identify and aggregate clinically relevant information from long-term EEG recordings to support report generation at multiple levels.
To this end, Section~\ref{sec:forward} discusses the unique challenges of clinical ELMs and provides an overview of the model architecture and forward process.
Sections~\ref{sec:EAT}, \ref{sec:SAA}, and \ref{sec:prompt_fusion} then present the details of the three components.

\begin{figure}[t]
    \centering
     \includegraphics[width=0.98\textwidth]{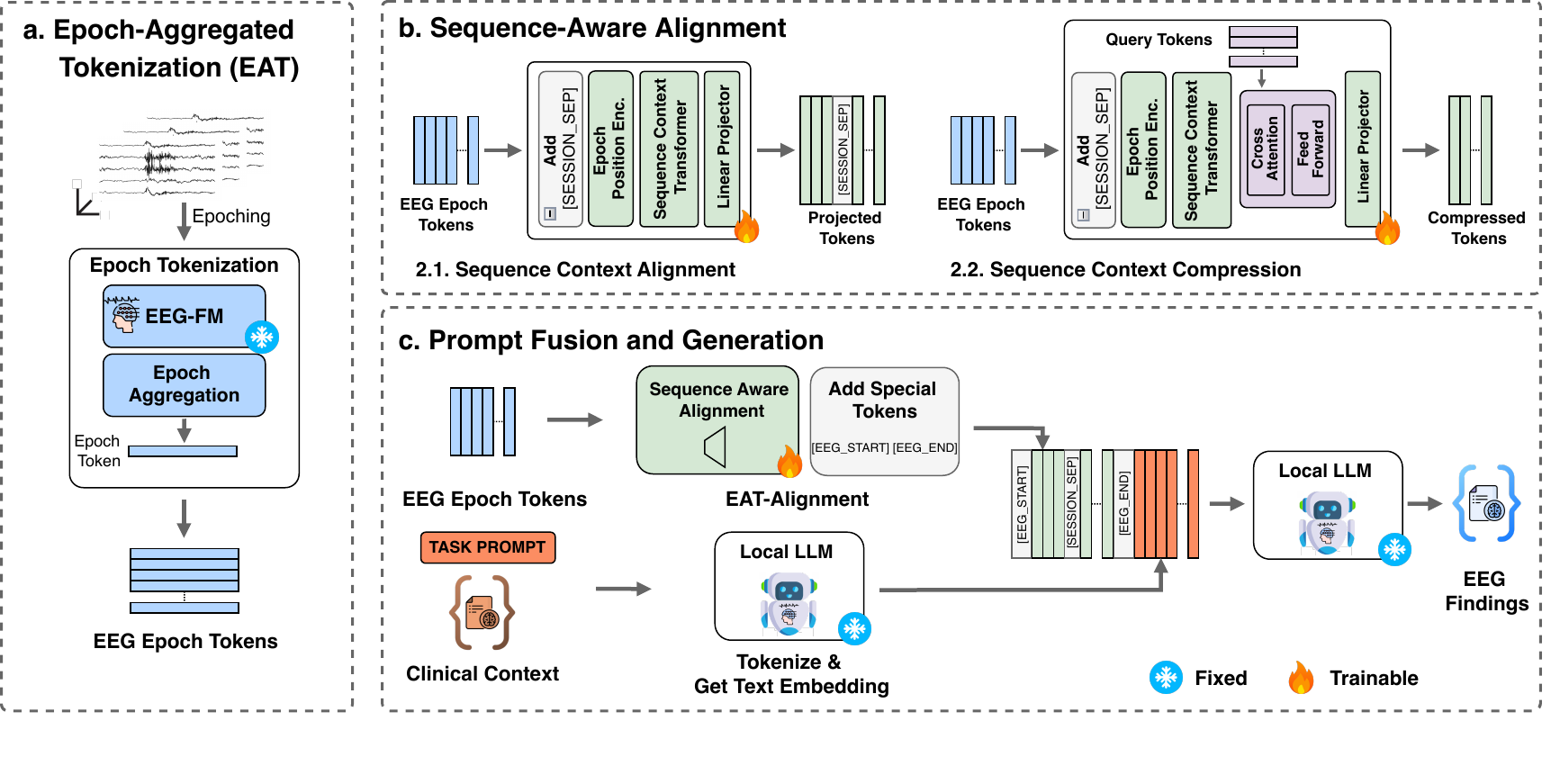}
     \vspace{-0.6cm}
    \caption{The proposed Clinical EEG Language Model (CELM) comprises (a) Epoch-Aggregated Tokenization, (b) Sequence-Aware Alignment, and (c) Prompt Fusion and Generation.
    }
    \label{fig:method_celm}
    \vspace{-0.5cm}
\end{figure}

\vspace{-0.1cm}
\subsection{Architecture and Forward Process}\label{sec:forward}

Clinical EEG introduces three unique challenges that prevent the direct adoption of existing multimodal paradigms, such as dictionary-based alignment or contrastive learning with fixed-length epochs \cite{ndir2025eeg}.
Here, we design \method with three corresponding steps:
\vspace{-0.1cm}

\textbf{Epoch-Aggregated Tokenization} addresses the extreme temporal scale of clinical EEG.
EEG sessions span hours of high-frequency, multi-channel signals, resulting in at least $\sim31.7\mathrm{M}+$ data points per recording, far exceeding the context limits of modern LLMs~\cite{grattafiori2024llama,team2025gemma,yang2025qwen3}.\\
\textbf{Sequence-Aware Alignment} mitigates the loss of temporal context during alignment.
EEG signals are inherently sequential, and naive projection into the LLM embedding space fails to preserve long-range temporal dependencies required for clinical interpretation.\\
\textbf{Prompt Fusion and Generation} addresses weak EEG–text correspondence.
Clinical reports aggregate findings over entire recordings without explicit temporal grounding, requiring the model to synthesize coherent clinical narratives from distributed EEG evidence.

Overall, given an input EEG recording, \method processes the signal through above three stages.
The recording is first transformed into compact epoch-level tokens via \emph{Epoch-Aggregated Tokenization}, which reduces the temporal scale.
These tokens are then projected into the language space through \emph{Sequence-Aware Alignment}, enabling long-range temporal dependencies to be maintained across epochs.
Finally, \emph{Prompt Fusion and Generation} conditions the language model on the aligned EEG representations and optional clinical context to generate structured reports.

\subsection{Epoch-Aggregated Tokenization}
\label{sec:EAT}
The fundamental bottleneck in developing ELMs is representing long EEG recordings within the LLM context constraints. A typical two-hour clinical EEG recording at $200$ Hz across $22$ channels yields $7,200\times200\times22\approx31.7$M data points, far exceeding current LLM capacities, making raw sample-level tokenization infeasible. A common alternative is mini-window tokenization, leveraging pretrained EEG encoders that typically tokenize 1-second segments. However, naively applying this granularity to full recordings produces $\approx158$K tokens per session, still well beyond the context limits of most local LLMs~\cite{grattafiori2024llama, team2025gemma, yang2025qwen3}. 
To overcome this limitation while harnessing pretrained EEG representations, we propose Epoch-Aggregated Tokenization (see Figure~\ref{fig:method_celm}a). 

Given an EEG recording session $\mathbf{X}\in \mathbb{R}^{N\times C\times T}$, where $N, C$, and $T$ denote the number of epochs, channels, and time points per epoch (typically 10 seconds) respectively, we first tokenize $1$s windows within each epoch using a pretrained EEG encoder. The resulting representations are then aggregated across windows and channels to produce a single token per epoch, yielding a compact sequence of epoch tokens $\mathbf{E}_{\text{eeg}} \in \mathbb{R}^{N \times D_{\text{eeg}}}$, where $D_{\text{eeg}}$ is the embedding dimension. The aggregation strategy is encoder-dependent. For example, CBraMod applies pooling~\cite{wang2024cbramod}, whereas TFM-Tokenizer~\cite{pradeepkumar2025single} uses \texttt{[CLS]} pooling. This tokenization compresses the tokens by up to $C\times T$ ($\sim220\times$ compared to window tokenization), enabling EEG recordings to fit within LLM context limits.

\subsection{Sequence-Aware Alignment}
\label{sec:SAA}

To enable EEG-conditioned clinical report generation, EEG representations must be aligned with the text embedding space of LLMs. Existing vision-language models such as LLaVA achieve this through simple linear projection~\cite{liu2023visual}, but this strategy does not generalize well to EEG (Section~\ref{sec:alignment_module_ablation}) due to its sequential and long-range temporal structure. Unlike static images, EEG signals contain temporal dynamics across epochs that must be preserved during alignment. This introduces two key challenges: modeling long temporal dependencies and scaling to very long sequences. To address these challenges, we propose two sequence-aware alignment strategies (Figure~\ref{fig:method_celm}b): (1) Sequence Context Alignment, which preserves full temporal structure, and (2) Sequence Context Compression, which compresses EEGs into fixed-length latent representations for memory-efficient alignment.

\paragraph{Sequence Context Alignment (SCA).} Given a sequence of tokens $\mathbf{E}_{\text{eeg}}$, we augment the sequence with position encodings and \texttt{[SESSION\_SEP]} tokens to differentiate EEG sessions. A lightweight linear-attention transformer~\citep{wang2020linformer} is then applied to capture the temporal structure across epochs. The resulting representations are subsequently projected through a linear layer into the language embedding space, yielding $\mathbf{H}_{\text{eeg}}\in \mathbb{R}^{N \times D_{\text{llm}}}$, where $D_{\text{llm}}$ denotes the LLM embedding dimension.

\paragraph{Sequence Context Compression (SCC).} To address the memory and scalability challenges of long EEG sequences, we explore a compression strategy inspired by Perceiver architectures~\citep{jaegle2021perceiver, jaegle2021perceiverIO}. After encoding temporal structure among epoch tokens, a fixed set of learnable query tokens $\mathbf{Q}\in  \mathbb{R}^{L \times D_{\text{eeg}}} (L<N)$ attend to the full sequence of epoch tokens via cross-attention. This compresses the variable-length EEG sequence into a fixed number of tokens, which are subsequently projected into the language embedding space, enabling efficient alignment for recordings of arbitrary duration.

\subsection{Prompt Fusion and Generation}
\label{sec:prompt_fusion}
\method generates clinical reports by conditioning a local LLM on EEG-derived representations jointly with clinical context when available (see Figure~\ref{fig:method_celm}c). Given projected EEG tokens $\mathbf{H}_{\text{eeg}}$, we augment the sequence with special tokens \texttt{[EEG\_START]} and \texttt{[EEG\_END]} to distinguish EEG representations from text inputs. A task-specific prompt that specifies the target report sections, together with optional clinical context (e.g., reason for study or clinical history), is tokenized using the LLM tokenizer to obtain text embeddings 
$\mathbf{H}_{\text{prompt}}$. 
The final input sequence is formed by concatenation,
\begin{equation}
\mathbf{H}_{\text{in}} = [\texttt{[EEG\_START]}; \mathbf{H}_{\text{eeg}}; \texttt{[EEG\_END]}; \mathbf{H}_{\text{text}}]
\end{equation}
which is processed autoregressively to generate a structured EEG report text. We employ instruction tuned Qwen3-4B~\citep{yang2025qwen3} as our base LLM, CBraMod~\citep{wang2024cbramod} as our EEG encoder and perform supervised fine-tuning using the next-token prediction objective:
\vspace{-0.2cm}
\begin{equation}
\mathcal{L} = -\sum_{t=1}^{T} \log P_\theta(y_t \mid H_{\text{input}}, y_{<t})
\end{equation}
During training, both the EEG encoder and LLM remain frozen, only the Sequence-Aware Alignment module is updated, enabling efficient adaptation while preserving pretrained representations.




%% file: 040experiments.tex
\paragraph{Datasets. }
The EEG-Report benchmark is constructed using data from two hospital sites in the Harvard Electroencephalography Database v4.1~\citep{zafar2025harvard,sun2025harvard}: Massachusetts General Hospital (MGH, S0001) and Brigham and Women's Hospital (BWH, S0002). In total, through our pipeline, we curated $12,290$ clinical EEG reports matched to one or more EEG sessions from $10,886$ patients. For model training and evaluation, we restrict our analysis to reports paired with a single EEG session.\\
\ding{112} \emph{S0001:} This site includes diverse EEG visit types (routine EEG, epilepsy monitoring unit (EMU), portable EEG, complex EEG). 
We filtered only routine EEG and EMU samples paired with single sessions, excluding long recordings exceeding $10,000$ seconds to prevent memory issues during training. 
We curated $5{,}049$ reports from $4{,}669$ patients.\\
\ding{112} \emph{S0002:} From the available visit types (EEG and portable EEG), we retained only standard EEG visits with single corresponding sessions, yielding $4,873$ reports from $4,379$ patients.
\\
For both sites, we performed patient-wise 60/20/20 splits into training, validation, and test sets. Additional dataset details are provided in Appendix~\ref{app:dataset_details}.

\input{TABLES/main_results_with_patient_history_bertscore}

\paragraph{Baselines.} 
Due to the lack of existing EEG multimodal models for clinical EEG report generation, we compare \method against three types of baselines:  (1)\emph{ Unimodal + text-only input}: only clinical context is provided as input without EEG data, providing a lower bound that quantifies hallucination and isolates performance when the model does not attend to neural signals, (2) \emph{Unimodal + text and EEG features input}: the clinical context is augmented with channel-wise spectral band power features (delta, theta, alpha, beta, gamma, Appendix~\ref{app:band_power}) extracted from each EEG session, overcoming context length limitation that prevents providing raw EEG arrays directly, (3) \emph{Multimodal + linear projector}: a linear projector maps pretrained EEG embeddings from CBraMod~\cite{wang2024cbramod} into the language model's embedding space, enabling direct conditioning on continuous EEG signals. We evaluate LLMs from three model families, including LLaMA-3~\citep{grattafiori2024llama}, Gemma-3~\citep{team2025gemma}, and Qwen-3~\citep{yang2025qwen3}, across multiple parameter scales, and include MedGemma~\citep{sellergren2025medgemma} as a domain-specific baseline (more details at Appendix~\ref{app:add_exp_details}). We report lexical and semantic  metrics~\cite{lin-2004-rouge,banerjee2005meteor,zhang2019bertscore} along with manual qualitative analysis, LLM-as-judge and a expert evaluation study to evaluate report generation performance.

\subsection{Report Generation Performance}

Table~\ref{tab:main_results_with_patient_history} compares the report generation performance of \method on test samples from both sites containing patient history. It compares against unimodal baselines that use either text-only or text with handcrafted EEG features as input, and finetuned multimodal baselines that use EEG foundation model embeddings. We report results for two \method variants: \textbf{\methodscc}, which uses sequence context 

\input{TABLES/on_all_data_main_results_without_patient_history}
compression to compress and produce a fixed number of projected EEG tokens, and \textbf{\method}, which uses non-compressed EEG representations via sequence context alignment. Incorporating EEG features yields modest gains over text-only on S0002 ($0.1622 \rightarrow0.2017$) except for LLaMA-3.1-8B, which fails due to its 8K context limit. On S0001, performance degrades due to greater multi-section report complexity. Finetuned multimodal baselines, which align EEG foundation embeddings with the language space via linear projection, achieve the best baseline performance. However, the marginal gains relative to the additional training overhead underscore the fundamental challenge of EEG-language alignment.



\method outperforms all baselines across all metrics, with statistically significant gains
($p<0.001$, paired Wilcoxon signed-rank test) and an approximately two-fold improvement in
ROUGE-1 and METEOR over the strongest baseline on both sites (e.g., S0002: $0.3084 \rightarrow 0.6408$).
Although \methodscc outperforms the baselines, its performance gap relative to the non-compressed model ($0.4487$ vs $0.6408$)
highlights a key trade-off between compression and performance for ELMs. This motivates future research in compression and alignment strategies that preserve clinically relevant information while scaling to longer EEG sequences.

\begin{figure}[t]
    \centering
     \includegraphics[width=\linewidth]{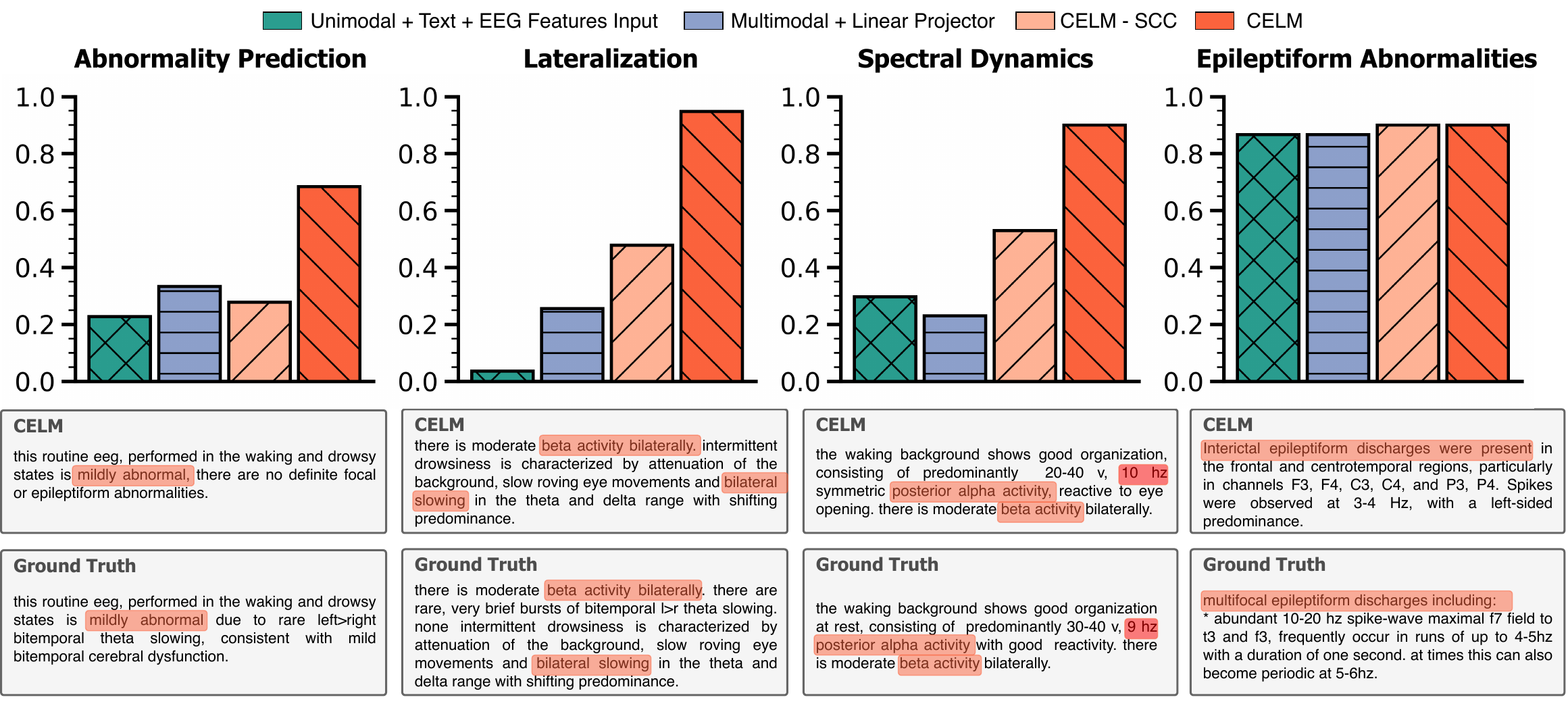}
    \caption{Manual qualitative analysis across four clinically relevant dimensions: abnormality prediction, lateralization, spectral dynamics, and epileptiform abnormalities. Representative example cases are provided below each dimension. }
    
    \label{fig:manual_inspect}
    \vspace{-0.4cm}

\end{figure}

\subsection{Zero-Context Report Generation}
\label{sec:zero-context}

To assess our model's ability to generate clinical reports solely from EEG signals, without relying on clinical context, we define a zero-context report generation task in which only EEG data is provided as input. This experiment also studies the potential language-model bias, in which reports could be generated by inferring plausible clinical context while ignoring EEG inputs. We conduct this experiment on both sites, using only the EEG signal as input, and summarize the results in Table~\ref{tab:main_results_without_patient_history}. \method outperforms all baselines across all metrics, demonstrating that it effectively extracts and utilizes clinically relevant information from EEG recordings.


\subsection{Manual Qualitative Analysis}
\label{sec:manual_qualitative_analysis}

While lexical metrics capture overlap, they fail to assess clinical correctness, and no reliable automated alternative exists.  To address this, we manually inspect 30 randomly sampled reports from both sites and evaluate correctness across four clinically relevant dimensions.  Results and representative examples are presented in Figure~\ref{fig:manual_inspect}. \ding{112} \textbf{Abnormality prediction.} We evaluate whether the generated report correctly classifies the recording as normal/abnormal.  \method achieves higher accuracy than all baselines. However, it frequently misclassifies mildly abnormal recordings as normal. One important observation is that even when the model accurately describes abnormal signal patterns (e.g., ``frequent bilateral but right-predominant temporal sharply contoured theta and delta slowing''), it often fails at decision-making. This suggests a lack of domain knowledge, in which injecting neurology guidelines presents a promising approach. 
\ding{112} \textbf{Lateralization.} We assess whether the generated report correctly identifies the hemisphere in which brain activities occur. \method performs significantly better than all baselines on this task. Notably, the unimodal baseline with EEG features and text input captures frequency information but fails to localize brain activity spatially. In contrast, \method and other baselines benefit from the learned channel embeddings of EEG foundation models, which enable better lateralization. \ding{112} \textbf{Spectral dynamics.} We study whether the generated report accurately captures frequency band behavior, which is a critical aspect of EEG interpretation. Interestingly, the unimodal baseline with handcrafted EEG features outperforms the multimodal baseline using linearly projected CBraMod embeddings, indicating that task-specific features can partially enable LLMs to infer specific EEG-related tasks. However, crafting such features is labor-intensive, motivating the use of foundation model embeddings. While naive linear projection fails to leverage these embeddings effectively, \method achieves strong performance, demonstrating the importance of proper alignment strategies. \ding{112} \textbf{Epileptiform abnormalities.} We evaluate whether the generated report correctly identifies epileptiform activity. While all models perform reasonably well on this dimension, we note that epileptiform events are rare and highly imbalanced in the dataset.


\subsection{Human Expert Evaluation Study}
\label{sec:human_expert_eval_study}

\begin{figure}[t]
    \centering
     \includegraphics[width= \linewidth]{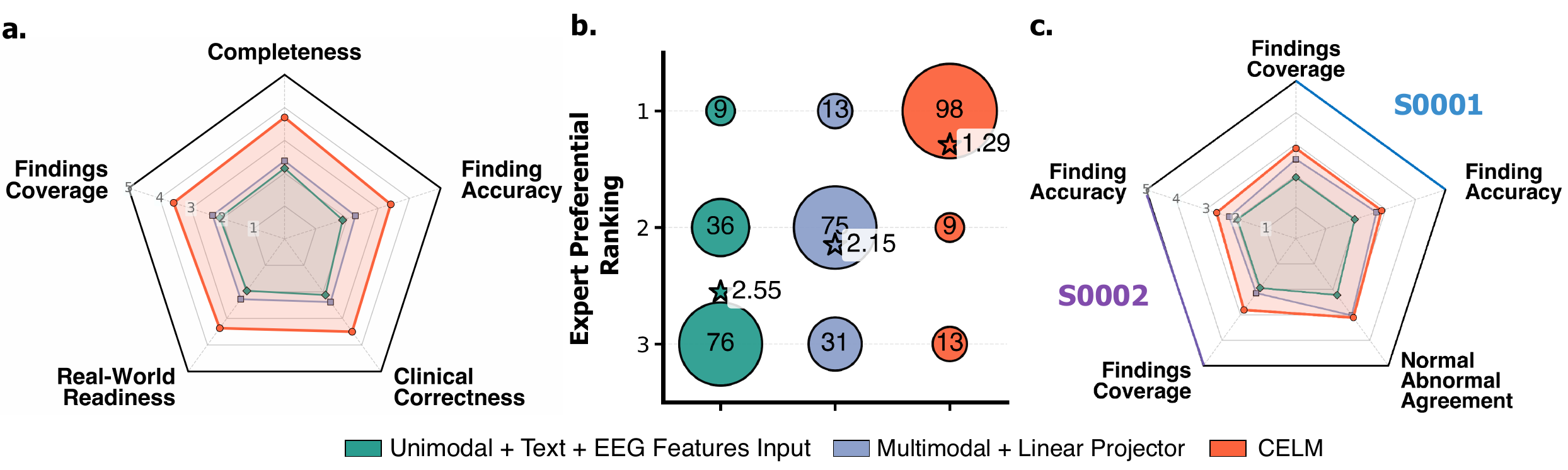}
    \caption{(a) Human expert evaluation study results,
(b) Human expert preferential ranking for each report. (c) LLMs-as-a-judge results 
    }
    \label{fig:humanexpert_study}
    \vspace{-0.7cm}

\end{figure}


To further strengthen our evaluation, we conducted an expert study in which six domain experts (clinicians and BME researchers) assessed report quality through an anonymized survey. Each expert evaluated 20 randomly sampled cases, with each case presenting the ground-truth clinical report alongside three anonymized generated reports from two different baseline categories and \method.
Experts rated each report on a 5-point scale (1: very poor to 5: very good) across five clinically relevant dimensions: (1) Clinical Correctness -  accuracy compared to the ground truth, (2) Findings Coverage - coverage of key findings identified relative to ground truth, (3) Finding Accuracy - accuracy of attributes such as lateralization, frequency, and morphology, (4) \textit{Real-World Readiness} - suitability for clinical use, and (5) \textit{Completeness} - completeness of the report, whether all relevant aspects are covered. Results are summarized in Figure~\ref{fig:humanexpert_study}a, where \method significantly outperforms both baselines across all five dimensions (Wilcoxon signed-rank test, $p<0.001$ for all comparisons). Experts also provided their preferred rankings across the reports, as shown in Figure~\ref{fig:humanexpert_study}b, where \method obtains a weighted-average rank of 1.29, with 98 votes for rank 1. 
\camred{ Inter-rater reliability was assessed using three measures: (1) ICC(2,k) was 0.79 overall (95\% CI 0.74–0.82), with per-dimension values of 0.82 for Finding Accuracy (95\% CI 0.75–0.89), 0.79 for Findings Coverage (95\% CI 0.70–0.87), 0.79 for Clinical Correctness (95\% CI 0.70–0.87), 0.76 for Real-World Readiness (95\% CI 0.65–0.84), and 0.76 for Completeness (95\% CI 0.65–0.84), all in the good range~\cite{koo2016guideline}, (2) Adjacent agreement (within $\pm$1 point on the 5-point scale) was 74\%-80\% across dimensions, and (3) Kendall's $W$ indicated statistically significant concordance across all dimensions ($W$ = 0.40–0.49, $p < 0.001$). }

\textbf{LLM-as-judge.} Human expert evaluation is labor-intensive and difficult to scale. We conducted an LLM-as-judge evaluation of generated reports using the Qwen2.5-72B-Instruct-AWQ model (Figure~\ref{fig:humanexpert_study}c). Although closed LLMs (e.g., GPT-4, Claude) correlate well with human judgments, data-usage agreement prohibits their use. We define three evaluation metrics, as in the expert study, and \method consistently outperforms.

\vspace{-0.2cm}
\subsection{\camred{EEG Grounding Control Analysis}}
\label{sec:eeg_control_experiments}

\input{TABLES/eeg_grounding_control_experiments}
Clinical EEG reports often follow standardized structures and templates, raising the possibility that \method learns site-specific reporting conventions rather than clinically relevant EEG representations. While Section~\ref{sec:zero-context} demonstrates that \method can generate reports without clinical context, we further investigate its dependence on EEG signals through five inference-time perturbation experiments on the S0002 test set, without any retraining (Table~\ref{tab:controls}):
\ding{112} \textbf{Mismatch}: pairing each reference report with a randomly selected EEG recording from a  different patient, 
\ding{112} \textbf{Channel permutation}: randomly permuting EEG channels across the montage,
\ding{112} \textbf{Shuffle epochs}: randomly permuting the temporal order of EEG epochs,
\ding{112} \textbf{Gaussian noise}: replacing EEG signals with Gaussian noise scaled to the original signal's standard deviation,  and
\ding{112} \textbf{Repeat single epoch}: repeating a randomly selected epoch to match the original recording length.

All five perturbations significantly degrade performance ($p<0.01$, Wilcoxon signed-rank test), with relative reductions of 32-39\% in ROUGE-1 and 32-40\% in BERTScore. The substantial decline in BERTScore indicates that the degradation extends beyond lexical overlap. Epoch shuffling produces performance losses comparable to replacing EEG signals with noise, suggesting that \method is sensitive to temporal organization. Similarly, channel permutation causes substantial degradation, indicating reliance on spatial channel information, consistent with its lateralization performance (Figure~\ref{fig:manual_inspect}). Collectively, these results provide evidence that \method relies on EEG recordings rather than site-specific reporting templates alone.

\subsection{Alignment Module Ablation}
\label{sec:alignment_module_ablation}
\begin{wrapfigure}{r}{0.465\textwidth}
\vspace{-0.5cm}
    \centering
     \includegraphics[width=\linewidth]{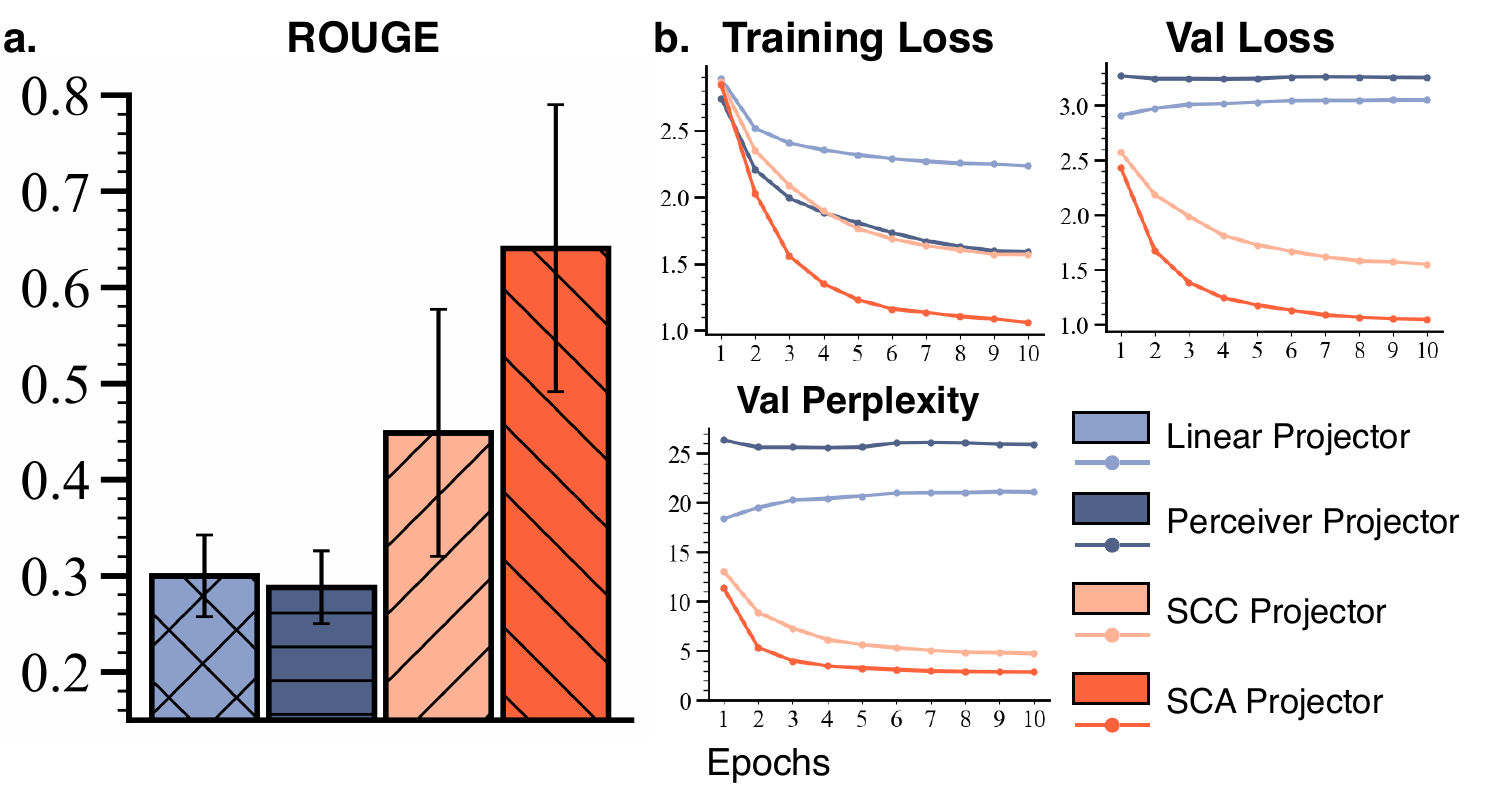}
    \caption{(a) Report generation performance of different alignment module variants.
(b) Training dynamics of each variant, including training loss, validation loss, and perplexity curves.
    }
    \label{fig:projector_ablation}
    \vspace{-0.3cm}

\end{wrapfigure}

The alignment module is critical for ELMs, and naive linear projection has been shown to be insufficient. We ablate four alignment strategies: (1) \textit{Linear projector}, projection via a single linear layer, (2) \textit{Perceiver projector}, cross-attention with learnable queries followed by feedforward \camred{(similar to Perceiver resampler in
Flamingo~\cite{alayrac2022flamingo}, and analogous to the Q-Former in BLIP-2~\cite{li2023blip})} and our proposed (3) \textit{SCA} and (4) \textit{SCC} projectors (Section~\ref{sec:SAA}). As shown in Figure~\ref{fig:projector_ablation}, SCA achieves the best performance. Training dynamics (Figure~\ref{fig:projector_ablation}b) reveal that Linear and Perceiver projectors overfit, with increasing validation loss despite decreasing training loss, whereas SCA and SCC exhibit stable convergence. This confirms that modeling inter-epoch temporal dependencies prior to projection is essential for effective alignment.

\vspace{-0.2cm}

\section{\camred{Broader Impact, Limitations and Implications for Future Work}}
\label{sec:discussion}
\vspace{-0.2cm}
\camred{\paragraph{Limitations and Implications for Future work.} In this work, we demonstrate the feasibility of end-to-end clinical EEG–language modeling for generating structured reports directly from long-duration EEG recordings. The strong performance gains over competitive baselines highlight the potential of ELMs for clinical report generation. However, several limitations remain. 
\ding{43}\emph{Failure modes.} \method{}  often misclassifies mildly abnormal EEGs as normal, even when correctly describing abnormalities, and struggles with rare findings such as interictal epileptiform abnormalities.
\ding{43}\emph{Evaluation limitations.} ELM development is constrained by the lack of rigorous benchmarks and clinically grounded evaluation protocols, as common text generation metrics mainly capture lexical similarity rather than clinical correctness. 
\ding{43}\emph{Memory scalability.} Memory remains a key bottleneck. While our approach supports EEG recordings of up to approximately three hours, further advances in memory-efficient representations are needed to scale to longer recordings. 
\ding{43}\emph{Human-in-the-loop potential.} ELMs offer opportunities for human-in-the-loop workflows, where clinicians guide generation via targeted prompts instead of fixed templates, enabling better alignment with clinical intent.}

\vspace{-0.2cm}
\camred{
\paragraph{Broader Impact and Intended Use Statement.}
This work advances machine learning research by introducing the first clinical EEG–language model that integrates long-duration clinical EEG signals with large language models to generate structured clinical reports. The proposed approach has the potential to support clinical workflows. However, this work is intended only as a research contribution rather than a deployable clinical system. \method{} is an assistive drafting tool that needs neurologist review, and must not be used for autonomous reporting or diagnosis. Any real-world application would require extensive validation, regulatory approval, and careful oversight by experts.  }


%% file: TABLES/main_results_with_patient_history_bertscore.tex
\begin{table}[t]

\centering
\caption{\camred{Report generation performance comparison on samples with clinical context across two hospital sites (S0001 and S0002).
We report mean $\pm$ standard deviation for ROUGE-1, METEOR and BERTScore. 
The best results are highlighted in \textcolor{highlightorange}{orange}, and the best baseline per category is highlighted in \textcolor{highlightblue}{blue}.
Relative improvements are also reported, and $^\star$ indicates finetuning.}}

\label{tab:main_results_with_patient_history}
\begin{adjustbox}{max width=\textwidth}

\setlength{\tabcolsep}{4pt} 
\begin{tabular}{lcccccc}
\toprule
&\multicolumn{3}{c}{\textbf{S0001}} & \multicolumn{3}{c}{\textbf{S0002}}\\

\cmidrule(lr){2-4} 
\cmidrule(lr){5-7} 
\textbf{Method}   & \textbf{ROUGE-1}  &\textbf{METEOR}  & 
\textbf{BERTScore}  &\textbf{ROUGE-1}  &\textbf{METEOR}  & 
\textbf{BERTScore}\\
\midrule
 &\multicolumn{6}{c}{\textbf{Unimodal + Text Only Input}} \\
\cmidrule(lr){2-7}

Gemma-3-1b-it& 0.2112 $\pm$ 0.1408& 0.1567 $\pm$ 0.1238 & 0.6461 $\pm$ 0.3014 
& 0.2330 $\pm$ 0.0876& 0.1319 $\pm$ 0.0565 & 0.7058 $\pm$ 0.2423 \\

Gemma-3n-e2b-it& 0.2410 $\pm$ 0.1722& 0.1875 $\pm$ 0.1725 & 0.7848 $\pm$ 0.0765
& 0.2318 $\pm$ 0.0600& 0.1181 $\pm$ 0.0472 & 0.7683 $\pm$ 0.0560 \\

Gemma-3-4b-it& 0.2857 $\pm$ 0.1590& 0.2147 $\pm$ 0.1668& 0.7941 $\pm$ 0.0922
& \cellcolor{highlightblue!35} 0.2775 $\pm$ 0.0558& 0.1506 $\pm$ 0.0514 &0.7851 $\pm$ 0.0730\\

Gemma-3n-e4b-it& 0.2416 $\pm$ 0.1778& 0.1816 $\pm$ 0.1786 & 0.7847 $\pm$ 0.0747
& 0.1998 $\pm$ 0.0649& 0.0993 $\pm$ 0.0490& 0.7635 $\pm$ 0.0525\\

Medgemma-4b-it& 0.2196 $\pm$ 0.1924& 0.1529 $\pm$ 0.1835 &  0.6683 $\pm$ 0.2994
& 0.2280 $\pm$ 0.0734& 0.1198 $\pm$ 0.0542 & 0.7576 $\pm$ 0.1375\\

Llama-3.2-1b-instruct& 0.0919 $\pm$ 0.0744& 0.0529 $\pm$ 0.0519 & 0.5580 $\pm$ 0.3284
& 0.0954 $\pm$ 0.0869& 0.0494 $\pm$ 0.0504 & 0.5139 $\pm$ 0.3501\\

Llama-3.2-3b-instruct& 0.2224 $\pm$ 0.1755& 0.1843 $\pm$ 0.1864 & 0.7116 $\pm$ 0.2311
& 0.2346 $\pm$ 0.0877& 0.1430 $\pm$ 0.0689 & 0.7261 $\pm$ 0.1919\\

Llama-3.1-8b-instruct& 0.2511 $\pm$ 0.1872& 0.1839 $\pm$ 0.1788 & 0.6994 $\pm$ 0.2605
& 0.2655 $\pm$ 0.0766& 0.1480 $\pm$ 0.0581 &0.7502 $\pm$ 0.1531\\

Meta-llama-3-8b-instruct& 0.2673 $\pm$ 0.1731& 0.1946 $\pm$ 0.1809 & 0.7392 $\pm$ 0.2055
& 0.2588 $\pm$ 0.0586& 0.1416 $\pm$ 0.0481 & 0.7702 $\pm$ 0.0817\\

Qwen3-1.7b & 0.1378 $\pm$ 0.1557& 0.1102 $\pm$ 0.1316 & 0.3856 $\pm$ 0.3941
& 0.1189 $\pm$ 0.1296& 0.0735 $\pm$ 0.0821 &  0.3650 $\pm$ 0.3893\\

Qwen3-4b-instruct-2507& \cellcolor{highlightblue!35} 0.3038 $\pm$ 0.1460& \cellcolor{highlightblue!35} 0.2418 $\pm$ 0.1440 & \cellcolor{highlightblue!35}0.7975 $\pm$ 0.0852

& 0.2690 $\pm$ 0.0448& \cellcolor{highlightblue!35} 0.1622 $\pm$ 0.0434 & \cellcolor{highlightblue!35} 0.7867 $\pm$ 0.0178 \\

\midrule

& \multicolumn{6}{c}{\textbf{Unimodal + Text + EEG Features Input}} \\
\cmidrule(lr){2-7}

Gemma-3-1b-it & 0.0086 $\pm$ 0.0524& 0.0064 $\pm$ 0.0462 & 0.0311 $\pm$ 0.1512
& 0.1755 $\pm$ 0.1157& 0.0954 $\pm$ 0.0692 & 0.5785 $\pm$ 0.3384\\

Gemma-3n-e2b-it & 0.0116 $\pm$ 0.0605& 0.0088 $\pm$ 0.0543 & 0.0415 $\pm$ 0.1756
& 0.2190 $\pm$ 0.0840& 0.1082 $\pm$ 0.0546 &0.7175 $\pm$ 0.2153\\

Gemma-3-4b-it & 0.0093 $\pm$ 0.0594& 0.0078 $\pm$ 0.0558 & 0.0247 $\pm$ 0.1384
& \cellcolor{highlightblue!35} 0.2886 $\pm$ 0.0458& 0.1738 $\pm$ 0.0476 & \cellcolor{highlightblue!35} 0.7883 $\pm$ 0.0418\\

Gemma-3n-e4b-it & 0.0397 $\pm$ 0.0606& 0.0231 $\pm$ 0.0392 & 0.2498 $\pm$ 0.3570
& 0.0357 $\pm$ 0.0903& 0.0180 $\pm$ 0.0462 &0.1073 $\pm$ 0.2678\\

Medgemma-4b-it & 0.2100 $\pm$ 0.1919& 0.1529 $\pm$ 0.1832 & 0.6172 $\pm$ 0.3319
& 0.1829 $\pm$ 0.1128& 0.1000 $\pm$ 0.0702 &0.5979 $\pm$ 0.3260\\

Llama-3.2-1b-instruct & 0.0820 $\pm$ 0.0727& 0.0473 $\pm$ 0.0479 & 0.4978 $\pm$ 0.3562
& 0.1016 $\pm$ 0.0773& 0.0517 $\pm$ 0.0448 &0.5903 $\pm$ 0.3062\\

Llama-3.2-3b-instruct & 0.1844 $\pm$ 0.1611& 0.1751 $\pm$ 0.1871 & 0.6441 $\pm$ 0.2898
& 0.2309 $\pm$ 0.0903& 0.1540 $\pm$ 0.0743 &0.7377 $\pm$ 0.1557\\

Llama-3.1-8b-instruct & \cellcolor{highlightblue!35} 0.2151 $\pm$ 0.1693& 0.1844 $\pm$ 0.1647 & 0.6930 $\pm$ 0.2289
& 0.2256 $\pm$ 0.0904& 0.1509 $\pm$ 0.0682 &0.6895 $\pm$ 0.2171\\

Qwen3-1.7b & 0.2051 $\pm$ 0.1453& 0.1683 $\pm$ 0.1304 & 0.6063 $\pm$ 0.3215
& 0.2049 $\pm$ 0.1021& 0.1413 $\pm$ 0.0724 &0.6345 $\pm$ 0.2944\\

Qwen3-4b-instruct-2507 & 0.2058 $\pm$ 0.1192& \cellcolor{highlightblue!35} 0.2067 $\pm$ 0.1079 & \cellcolor{highlightblue!35}0.7249 $\pm$ 0.1840
& 0.2638 $\pm$ 0.0341& \cellcolor{highlightblue!35} 0.2017 $\pm$ 0.0285 & 0.7704 $\pm$ 0.0125\\

\midrule
& \multicolumn{6}{c}{\textbf{Multimodal + Linear Projector + CBraMod$^\star$}} \\
\cmidrule(lr){2-7}

Gemma-3-4b-it & 0.2688 $\pm$ 0.1448& 0.2020 $\pm$ 0.1579 & 0.7904 $\pm$ 0.0798
& \cellcolor{highlightblue!35} 0.3084 $\pm$ 0.0491& 0.2002 $\pm$ 0.0482 & \cellcolor{highlightblue!35}0.7879 $\pm$ 0.0729\\

Medgemma-4b-it  & 0.0494 $\pm$ 0.1806& 0.0441 $\pm$ 0.1719 & 0.0886 $\pm$ 0.2601
& 0.1491 $\pm$ 0.1349& 0.0729 $\pm$ 0.0701 &0.4492 $\pm$ 0.3899\\

Llama-3.2-3b-instruct  & 0.0568 $\pm$ 0.1311& 0.0396 $\pm$ 0.1172 & 0.2402 $\pm$ 0.3583
& 0.2425 $\pm$ 0.0837& 0.1398 $\pm$ 0.0610 &0.7333 $\pm$ 0.1900\\

Llama-3.1-8b-instruct  & 0.0063 $\pm$ 0.0428& 0.0050 $\pm$ 0.0369 & 0.0189 $\pm$ 0.1169
& 0.0385 $\pm$ 0.0823& 0.0209 $\pm$ 0.0458 &0.1533 $\pm$ 0.3011\\

Qwen3-4b-instruct-2507  & \cellcolor{highlightblue!35}  0.2819 $\pm$ 0.1132& \cellcolor{highlightblue!35}  0.2444 $\pm$ 0.1093 & \cellcolor{highlightblue!35} 0.7921 $\pm$ 0.0417
& 0.2998 $\pm$ 0.0425& \cellcolor{highlightblue!35} 0.2143 $\pm$ 0.0349 &0.7852 $\pm$ 0.0153\\

\midrule




\textbf{\methodscc}   &  \textbf{ 0.3843} $\pm$ 0.1876& \textbf{ 0.2889} $\pm$ 0.1866 & \textbf{0.7949} $\pm$ 0.1783

& \textbf{0.4487} $\pm$ 0.1283& \textbf{0.3232} $\pm$ 0.1261 & \textbf{0.8229} $\pm$ 0.1341\\

\textbf{\method}  & \cellcolor{highlightorange!35}\textbf{ 0.5565} $\pm$ 0.1683& \cellcolor{highlightorange!35}\textbf{ 0.4734} $\pm$ 0.1941 & \cellcolor{highlightorange!35}\textbf{0.8697} $\pm$ 0.0901 

& \cellcolor{highlightorange!35}\textbf{ 0.6408} $\pm$ 0.1494& \cellcolor{highlightorange!35}\textbf{ 0.5597} $\pm$ 0.1728&

\cellcolor{highlightorange!35}\textbf{ 0.8755} $\pm$ 0.1521

\\

\midrule

\textbf{Improvement}  &$+$83.18$\%$ &$+$93.70$\%$ & $+$9.05$\%$
 &$+$107.78$\%$ &$+$161.18$\%$ &  $+$11.06$\%$\\






\bottomrule

\end{tabular}
\end{adjustbox}

\vspace{-0.6cm}
\end{table}

%% file: TABLES/on_all_data_main_results_without_patient_history.tex
\begin{wraptable}{r}{0.48\textwidth}

\centering
\caption{Zero-context report generation performance on S0001 and S0002.
The best results are highlighted in \textcolor{highlightorange}{orange}, and the best baseline is highlighted in \textcolor{highlightblue}{blue}. 
}
\label{tab:main_results_without_patient_history}
\begin{adjustbox}{max width=\linewidth}

\setlength{\tabcolsep}{4pt} 
\renewcommand{\arraystretch}{1.0}
\begin{tabular}{lcccc}
\toprule

&\multicolumn{2}{c}{\textbf{S0001}} & \multicolumn{2}{c}{\textbf{S0002}}\\

\cmidrule(lr){2-3} 
\cmidrule(lr){4-5} 
\textbf{Method}   &\textbf{ROUGE-1}  &\textbf{METEOR}    &\textbf{ROUGE-1}  &\textbf{METEOR} \\
\midrule

& \multicolumn{4}{c}{\textbf{Unimodal + Text + EEG Features Input}} \\
\cmidrule(lr){2-5}

Gemma-3-4b-it& \cellcolor{highlightblue!35} 0.20 $\pm$ 0.09& 0.15 $\pm$ 0.07
& 0.09 $\pm$ 0.13& 0.05 $\pm$ 0.08\\

Medgemma-4b-it& 0.17 $\pm$ 0.12& 0.10 $\pm$ 0.07
& 0.13 $\pm$ 0.10& 0.07 $\pm$ 0.06\\

Llama-3.2-3b-it& 0.14 $\pm$ 0.11& 0.10 $\pm$ 0.08
& 0.20 $\pm$ 0.09& 0.12 $\pm$ 0.06\\

Qwen3-4b-it& 0.16 $\pm$ 0.07& \cellcolor{highlightblue!35} 0.16 $\pm$ 0.06
& \cellcolor{highlightblue!35} 0.23 $\pm$ 0.08& \cellcolor{highlightblue!35} 0.17 $\pm$ 0.06\\

\midrule
& \multicolumn{4}{c}{\textbf{Multimodal + Linear Projector + CBraMod}} \\
\cmidrule(lr){2-5}

Gemma-3-4b-it & 0.24 $\pm$ 0.05& 0.17 $\pm$ 0.05
& \cellcolor{highlightblue!35} 0.29 $\pm$ 0.06& 0.19 $\pm$ 0.05\\

Medgemma-4b-it & 0.03 $\pm$ 0.09& 0.02 $\pm$ 0.06
& 0.18 $\pm$ 0.11& 0.08 $\pm$ 0.05\\

Llama-3.2-3b-it & 0.02 $\pm$ 0.06& 0.01 $\pm$ 0.04
& 0.13 $\pm$ 0.12& 0.07 $\pm$ 0.07\\

Qwen3-4b-it& \cellcolor{highlightblue!35}  0.25 $\pm$ 0.05 & \cellcolor{highlightblue!35}  0.22 $\pm$ 0.04
& 0.29 $\pm$ 0.03& \cellcolor{highlightblue!35} 0.21 $\pm$ 0.03\\

\midrule



\textbf{\methodscc} &  \textbf{ 0.27} $\pm$ 0.17& \textbf{ 0.21} $\pm$ 0.14
& \textbf{0.39} $\pm$ 0.12 & \textbf{0.26} $\pm$ 0.10
\\

\textbf{\method} & \cellcolor{highlightorange!35}\textbf{ 0.51} $\pm$ 0.16 &  \cellcolor{highlightorange!35}\textbf{ 0.41} $\pm$ 0.17
& \cellcolor{highlightorange!35}\textbf{ 0.56} $\pm$ 0.20 & \cellcolor{highlightorange!35}\textbf{ 0.49} $\pm$ 0.21
\\

\midrule
\textbf{Improvement} &  $+$104$\%$ &$+$86$\%$ &$+$93$\%$ & $+$133$\%$ \\




\bottomrule

\end{tabular}
\end{adjustbox}

\vspace{-0.5cm}
\end{wraptable}

%% file: TABLES/eeg_grounding_control_experiments.tex
\begin{wraptable}{r}{0.55\textwidth}

\vspace{-0.5cm}
\centering
\caption{Report generation performance under inference time EEG perturbations in S0002. We report mean $\pm$ standard deviation with the relative drop \%.  
}
\label{tab:controls}
\begin{adjustbox}{max width=\linewidth}

\setlength{\tabcolsep}{3pt} 
\renewcommand{\arraystretch}{1.0}
\begin{tabular}{lcccc}
\toprule

\textbf{Perturbation}  &\textbf{ROUGE-1}  & \textbf{Drop \%}  &\textbf{BERTScore} & \textbf{Drop \%}\\
\midrule

\method & \cellcolor{highlightorange!35}\textbf{0.641} $\pm$ 0.149 & & \cellcolor{highlightorange!35}\textbf{0.876} $\pm$ 0.152	& \\

Mismatch & 0.390 $\pm$ 0.322 & \textcolor{green}{$\downarrow$} 39.2\% &  0.544 $\pm$ 0.437 & \textcolor{green}{$\downarrow$} 37.8\% \\

Channel permutation & 0.437 $\pm$ 0.328 & \textcolor{green}{$\downarrow$} 31.7\% &  0.593 $\pm$ 0.430 & \textcolor{green}{$\downarrow$} 32.3\% \\

Shuffle epochs & 0.394 $\pm$ 0.341 & \textcolor{green}{$\downarrow$} 38.5\% &  0.531 $\pm$ 0.447 & \textcolor{green}{$\downarrow$}39.4\% \\

Gaussian noise & 0.390 $\pm$ 0.341 & \textcolor{green}{$\downarrow$}39.1\%&  0.526 $\pm$ 0.447 & \textcolor{green}{$\downarrow$}39.9\% \\

Repeat single epoch & 0.423 $\pm$ 0.326  & \textcolor{green}{$\downarrow$} 33.8\% &  0.581 $\pm$ 0.432 & \textcolor{green}{$\downarrow$}33.7\% \\

\bottomrule

\end{tabular}
\end{adjustbox}

\vspace{-0.5cm}
\end{wraptable}


%% file: 050conclusion.tex
We introduced CELM, the first clinical EEG-to-language foundation model for automated report generation from long-duration EEG recordings. CELM addresses three core challenges in this setting: (1) representing hour-scale EEG within LLM context limits via Epoch-Aggregated Tokenization, (2) preserving long-range temporal dependencies through Sequence-Aware Alignment, and (3) enabling flexible, multi-scale report generation via Prompt Fusion. We further presented a scalable EEG–Report benchmark construction pipeline and conducted extensive evaluations across hospital sites, report sections, and zero-context settings. Empirical results show that CELM consistently outperforms strong baselines. Overall, this work establishes EEG-to-language modeling as a distinct and promising research direction at the intersection of multimodal learning and clinical neurophysiology, and we hope that CELM and the accompanying benchmark construction pipeline will facilitate future advances in long-context EEG modeling and clinically grounded EEG language systems.

%% file: 100appendix.tex
\section{Details of EEG-Report Benchmark}
\label{app:dataset_details}

\begin{figure}[ht]
    \centering
    \includegraphics[width=0.95\linewidth]{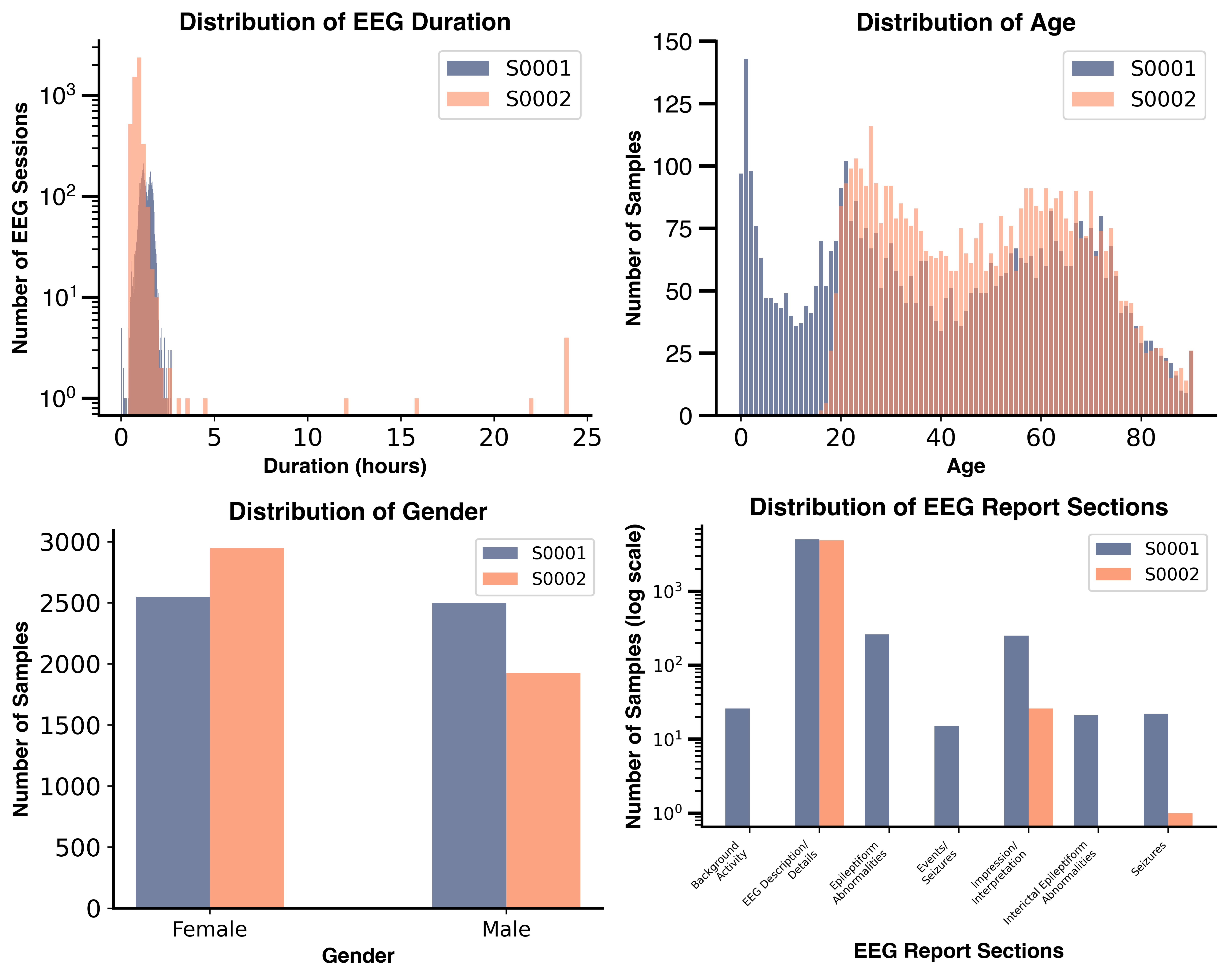}
    \caption{Dataset statistics of the filtered and constructed EEG-Report Benchmark used in our study
    }
    \label{fig:dataset_statistics}

\end{figure}

This section provides detailed information on the EEG–Report benchmark constructed in our study. Table~\ref{tab:dataset_stats} summarizes key statistics, including the number of EEG–report pairs, patient counts, and total EEG recording duration for each site. Figure~\ref{fig:dataset_statistics} further illustrates dataset distributions across EEG session duration, patient demographics (age and gender), and EEG report sections.
\input{TABLES/dataset_statistics}

\vspace{-0.2cm}

\section{Additional Experiment Details}
\label{app:add_exp_details}

\subsection{Band Power Features}
\label{app:band_power}
This section describes the extraction of handcrafted EEG features used in the unimodal text + EEG feature baselines.  To construct EEG features that fit current context-length constraints, we extract band-power features from EEG recordings. For each 10-second multichannel EEG segment, we compute channel-wise band power using Welch’s method by first estimating the power spectral density and then integrating it within five EEG frequency bands: delta (0.5–4 Hz), theta (4–8 Hz), alpha (8–13 Hz), beta (13–30 Hz), and gamma (30–80 Hz).
\input{TABLES/llm_baselines_and_context_length}
To further reduce input sequence length for LLMs with limited context windows, we optionally pool consecutive epochs by concatenating their time axes prior to spectral estimation, yielding a more compact representation. The resulting band power values are formatted as structured text and provided as input to the LLMs (see example in Figure~\ref{fig:unimodal_text_and_eeg_features_prompt}).

\subsection{Baselines and Prompts}

In this section, we provide additional details on the baseline models and the prompts used in our experiments. Table~\ref{tab:llm_baselines} summarizes all LLMs considered in our study along with their respective context lengths\footnote{Context length information is extracted from their respective technical reports and Hugging Face \url{https://huggingface.co/}}, further underscoring the practical challenges of directly conditioning LLMs on long-duration EEG recordings. Figures~\ref{fig:unimodal_text_only_prompt} and~\ref{fig:unimodal_text_and_eeg_features_prompt} present the prompts used for the unimodal text-only and unimodal text + EEG feature baselines, respectively. For clarity, we provide examples for certain prompt sections using bounding boxes.

\input{FIG/unimodal_text_only_prompts}
\input{FIG/unimodal_text_EEG_features_prompt}

\subsection{Evaluation Metrics}

We evaluate report generation performance using a comprehensive set of natural language generation metrics to evaluate different aspects of generation quality. In the main manuscript, we report ROUGE-1, METEOR and BERTScore. In Appendix~\ref{app:add_experiment_results}, we additionally report smoothed BLEU-1 and BLEU-4, ROUGE-2, ROUGE-L, and ROUGE-LSUM. We further include per-sample score distributions for all metrics in Appendix~\ref{app:score_distribution}.

\clearpage

\section{Implementation Details}
\label{app:implementation_details}

To facilitate reproducibility, we provide the complete source code for constructing the EEG–report benchmark from the Harvard Electroencephalography Database and the EHR database hosted on the Brain Data Science Platform, together with our model implementation, training scripts, and data loaders at \url{https://github.com/Jathurshan0330/CELM}. Additionally, we provide the detailed implementation details of our approach in this section and the prompts used for the baselines in Appendix~\ref{app:add_exp_details}. The dataset utilized for this study is publicly available and can be accessed at \url{https://bdsp.io/content/harvard-eeg-db/4.1/} after obtaining credentialed access. Experiments were conducted using multiple NVIDIA RTX 6000 and NVIDIA A100 GPUs with 49GB and 80GB of memory. 

\subsection{Model Details and Hyperparameters} 
We use CBraMod~\cite{wang2024cbramod} as the EEG encoder and Qwen-4B-Instruct-2507 as the local LLM backbone in our framework. CBraMod is selected due to its SOTA performance among EEG foundation models and its pretraining on large, diverse EEG datasets. In addition, our encoder ablation study comparing CBraMod with LaBraM shows that CBraMod consistently achieves better performance. We chose Qwen-4B-Instruct-2507 as the LLM backbone because it outperforms other local LLMs in unimodal baselines and reliably follows instruction prompts to generate outputs with the required structure. In Table~\ref{tab:elm_params}, we provide the hyperparameters of the alignment modules used in our study.

\input{TABLES/ELM_hyperparameters}

\input{TABLES/ELM_training_parameters}
\subsection{Training Details}
In our approach, the EEG encoder and LLM backbone are frozen, and only the alignment module is trained via supervised learning using a next-token prediction objective. Training details are summarized in Table~\ref{tab:elm_training}.

\subsection{Prompts}
Figure~\ref{fig:ELM_prompt} illustrates the prompt used in \\ our approach along with the EEG tokens. \\ We also provide examples of the \\sections in the prompt. 
\input{FIG/ELM_prompts}
\clearpage

\section{Additional Experiment Results}
\label{app:add_experiment_results}

\subsection{Extended Report Generation Performance on S0001}
Table~\ref{tab:app_baseline_results_S0001_with_PH} presents a comprehensive performance comparison of EEG report generation on the S0001 site for samples with clinical context. Both proposed variants consistently outperform all baselines across all evaluation metrics. Among them, \method achieves better performance compared to the memory-efficient \method-SCC variant across all metrics.
\input{TABLES/app_baseline_results_S0001_with_patient_history}

\subsection{Extended Report Generation Performance on S0002}
Tables~\ref{tab:app_baseline_results_S0002_with_PH} and~\ref{tab:app_baseline_results_S0002_without_PH} present a comprehensive performance comparison of EEG report generation on the S0002 site under settings with and without patient clinical context (subset of S0002 samples without patient clinical context). Both proposed variants consistently outperform all baselines across all evaluation metrics. Among them, \method achieves better performance compared to the memory-efficient \methodscc variant across all metrics.
\input{TABLES/app_baseline_results_S0002_with_patient_history}

\input{TABLES/app_baseline_results_S0002_without_patient_history}

\clearpage
\subsection{Additional Results on Alignment Module Ablation}
Comprehensive results of the alignment-module ablation study on the S0002 dataset, covering settings with clinical context, without clinical context, and all cases, are summarized in Table~\ref{tab:app_results_projector_ablation}. We report results on three groups of S0002 test recordings:
\textbf{(1) EEG Only}, the recordings whose reports contain no clinical-context section, 
\textbf{(2) With patient history}, the recordings with clinical context provided, and
\textbf{(3) All}, the full test set of recordings. Across most evaluation settings, the SCA projector achieves significantly stronger performance than other alignment variants, highlighting the importance of modeling dependencies among EEG epoch tokens before projecting them into the LLM embedding space.
\input{TABLES/main_eeg_projector_ablation}

\subsection{EEG Encoder Ablation}

We conducted an ablation study to analyze report generation performance across different EEG encoders. Specifically, we compared CBraMod~\cite{wang2024cbramod} and LaBraM~\cite{jiang2024large}, both EEG foundation models pretrained on large-scale EEG datasets. The results show that CBraMod consistently outperforms LaBraM, highlighting the critical role of high-quality EEG representations in effective clinical report generation.
\input{TABLES/app_eeg_encoder_ablation}

\subsection{Performance Analysis by Report Section}
Clinical EEG reports comprise multiple sections, each serving a distinct purpose: EEG description/details provides a detailed narrative of observed waveforms and patterns, impression/interpretation summarizes the clinical significance of findings, background activity characterizes baseline rhythms, and events/seizures documents seizure episodes. To evaluate whether \method can reliably generate these diverse sections, we conduct a section-wise analysis on S0001, which exhibits richer section diversity compared to S0002 (where reports predominantly contain only EEG description/details; see Figure~\ref{fig:dataset_statistics} in Appendix). Figure~\ref{fig:section_wise_results} presents ROUGE-1 scores, comparing \method against the best-performing unimodal baseline (text + EEG features) from each LLM family.  \method achieves the highest performance in 6 out of 7 report sections. Performance degrades in the interictal epileptiform abnormalities section, highlighting a key limitation and a challenge for ELMs in modeling rare and clinically complex events, an important direction for future work. Detailed results are provided in Table~\ref{tab:app_section_wise_performance}, which reports the complete section-wise performance across all models and evaluation metrics. These results further demonstrate the consistency of CELM across diverse components of clinical reports.

\begin{figure}[t]
    \centering
     \includegraphics[width=0.95\linewidth]{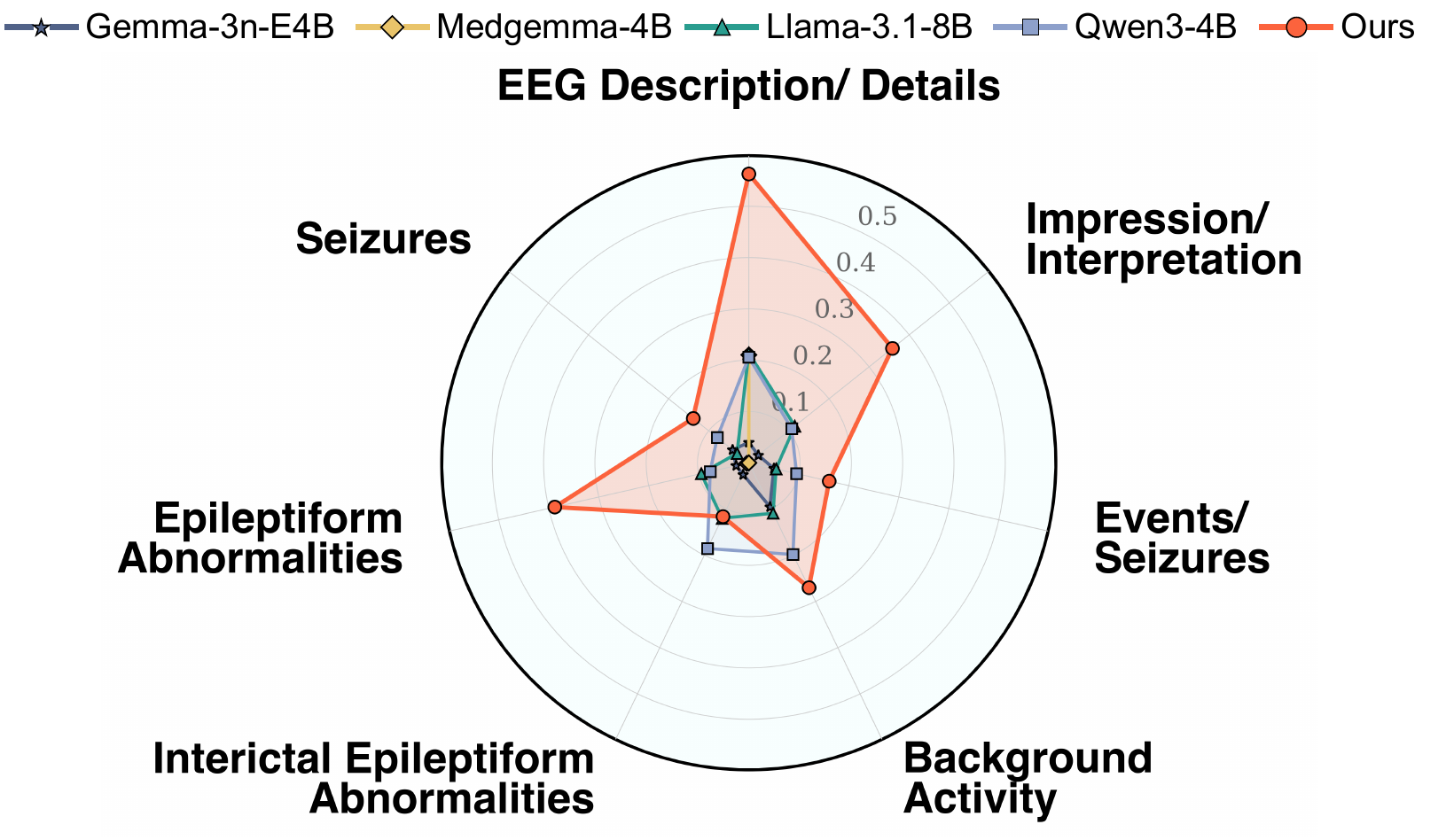}
    \caption{Section-wise comparison of report generation performance between CELM and the best-performing baselines from different LLM families.
    }
    
    \label{fig:section_wise_results}
    \vspace{-0.5cm}
\end{figure}

\input{TABLES/app_section_wise_performace}



\clearpage

\subsection{\camred{Cross-Site Generalization}}
\label{app:cross_site}

To assess whether \method{} generalizes beyond the reporting conventions and patient distributions of the site it
was trained on, we train \method{} on S0002 and evaluate it on the S0001 test set without
any fine-tuning on S0001. We compare it with the in-site model trained and tested on S0001
(Table~\ref{tab:cross_site}). Lexical metrics decrease substantially in the cross-site
setting (ROUGE-1: 0.5565 $\rightarrow$ 0.3585), whereas BERTScore decreases by only 7.8\%
(0.8697 $\rightarrow$ 0.8015). This gap between lexical and semantic metrics is consistent
with the two sites differing in reporting style as S0001 includes more diverse visit types
(routine EEG and epilepsy monitoring unit) and more multi-section report structures that
the S0002-trained model has not seen. The cross-site model
therefore retains much of the semantic content of the reports but does not reproduce
S0001's reporting conventions. 


\input{TABLES/cross_site_generalization_experiment}

\subsection{Score Distribution}
\label{app:score_distribution}

\begin{figure}[ht]
\vspace{-0.2cm}
    \centering
    \includegraphics[width=\linewidth]{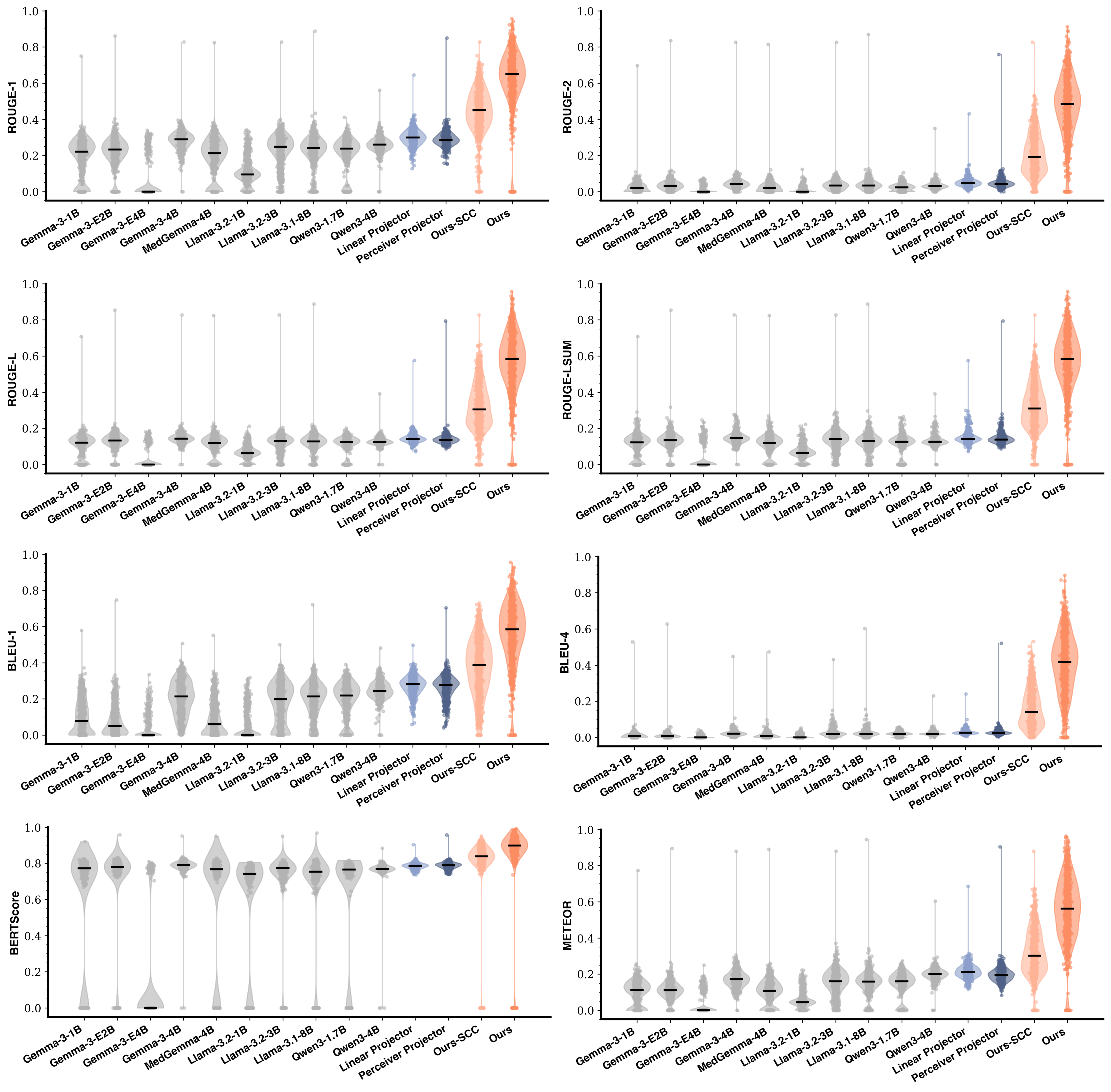}
    \caption{Distribution of various metrics for overall report generation in the S0002 dataset.
        }
    \label{fig:S0002_score_distribution}
    \vspace{-2cm}
\end{figure}

\begin{figure}[ht]
    \centering
    \includegraphics[width=\linewidth]{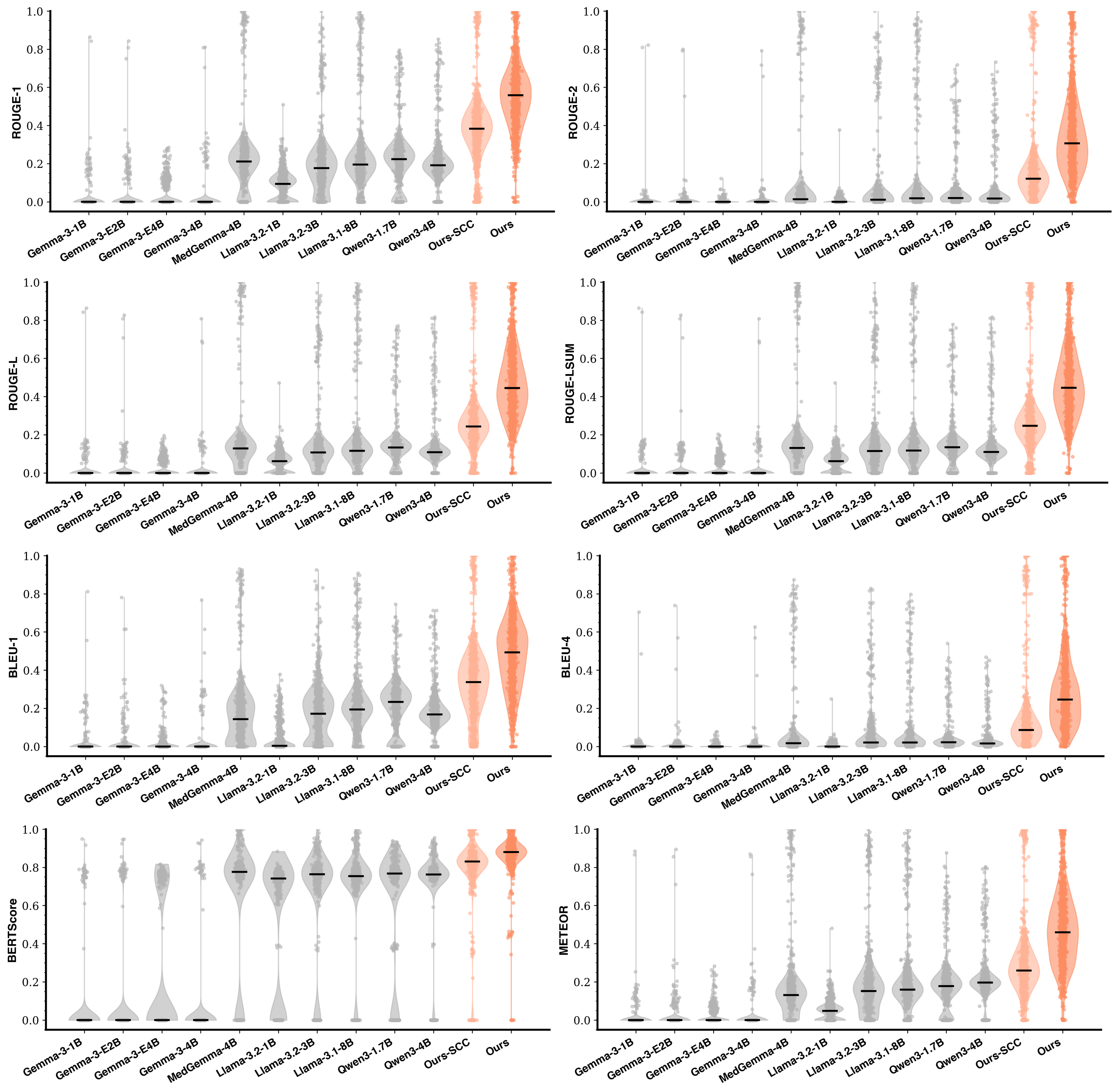}
    \caption{Distribution of various metrics for overall report generation in the S0001 dataset.
    }
    \label{fig:S0001_score_distribution}

\end{figure}

\clearpage

\clearpage




\subsection{\camred{SCA vs.\ SCC by Recording Duration}}
\label{app:sca_scc_duration}
To study how the performance of SCA and SCC projectors vary with recording length, we split each test set into duration quartiles and compare SCA and SCC within each quartile (Table~\ref{tab:sca_scc_duration}). SCA outperforms SCC in every quartile on both sites (Wilcoxon signed-rank test, $p<0.001$ within each quartile). On S0001, SCC degrades more with duration than SCA as ROUGE-1 falls from 0.4031 to
0.3479 from Q1 to Q4 ($-$14\%), compared with 0.5522 to 0.5173 for SCA ($-$6\%). This attributes to the fixed set of query tokens becoming an information bottleneck as recordings grow longer. On S0002, where recordings are comparatively shorter, both models improve with duration, and the SCA-SCC gap widens  as SCA benefits more from longer inputs.

\begin{table}[h]
\centering
\small
\caption{Report generation performance of SCA and SCC by recording-duration quartile.}
\label{tab:sca_scc_duration}
\setlength{\tabcolsep}{4pt}
\begin{tabular}{lcccccc}
\toprule
\textbf{Quartile} & \textbf{Dur. (min)} & \textbf{Range (min)} & \textbf{Proj.} & \textbf{ROUGE-1} & \textbf{METEOR} & \textbf{BERTScore} \\
\midrule
 & \multicolumn{6}{c}{\textbf{S0001}} \\
\cmidrule(lr){2-7}
\multirow{2}{*}{Q1} & \multirow{2}{*}{54.4} & \multirow{2}{*}{0.7--65.2} & SCA & \textbf{0.5522} & \textbf{0.4553} & \textbf{0.8709} \\
 &  &  & SCC & 0.4031 & 0.2895 & 0.8112 \\
\cmidrule(lr){1-7}
\multirow{2}{*}{Q2} & \multirow{2}{*}{71.5} & \multirow{2}{*}{65.3--77.8} & SCA & \textbf{0.5940} & \textbf{0.5089} & \textbf{0.8824} \\
 &  &  & SCC & 0.4041 & 0.3081 & 0.8013 \\
\cmidrule(lr){1-7}
\multirow{2}{*}{Q3} & \multirow{2}{*}{86.4} & \multirow{2}{*}{78.0--94.2} & SCA & \textbf{0.5618} & \textbf{0.4828} & \textbf{0.8758} \\
 &  &  & SCC & 0.3816 & 0.2899 & 0.7939 \\
\cmidrule(lr){1-7}
\multirow{2}{*}{Q4} & \multirow{2}{*}{103.4} & \multirow{2}{*}{94.3--161.5} & SCA & \textbf{0.5173} & \textbf{0.4461} & \textbf{0.8494} \\
 &  &  & SCC & 0.3479 & 0.2680 & 0.7729 \\
\midrule
 & \multicolumn{6}{c}{\textbf{S0002}} \\
\cmidrule(lr){2-7}
\multirow{2}{*}{Q1} & \multirow{2}{*}{37.1} & \multirow{2}{*}{24.5--46.3} & SCA & \textbf{0.5521} & \textbf{0.4845} & 0.8030 \\
 &  &  & SCC & 0.3861 & 0.2401 & \textbf{0.8048} \\
\cmidrule(lr){1-7}
\multirow{2}{*}{Q2} & \multirow{2}{*}{49.9} & \multirow{2}{*}{46.5--52.3} & SCA & \textbf{0.6603} & \textbf{0.5635} & \textbf{0.8962} \\
 &  &  & SCC & 0.4645 & 0.3414 & 0.8232 \\
\cmidrule(lr){1-7}
\multirow{2}{*}{Q3} & \multirow{2}{*}{55.4} & \multirow{2}{*}{52.5--58.2} & SCA & \textbf{0.6761} & \textbf{0.5996} & \textbf{0.8991} \\
 &  &  & SCC & 0.4801 & 0.3676 & 0.8372 \\
\cmidrule(lr){1-7}
\multirow{2}{*}{Q4} & \multirow{2}{*}{66.4} & \multirow{2}{*}{58.3--128.0} & SCA & \textbf{0.6757} & \textbf{0.5917} & \textbf{0.9044} \\
 &  &  & SCC & 0.4646 & 0.3444 & 0.8266 \\
\bottomrule
\end{tabular}
\end{table}

\subsection{\camred{Benchmark Construction Quality Validation}}
\label{app:benchmark_validation}

To assess the reliability of our benchmark construction pipeline, we conducted a systematic manual validation of section extraction and canonical normalization using 50 randomly sampled clinical EEG reports from both sites. \textbf{(1) Section extraction quality.}
We manually compared 150 extracted sections against their corresponding original report text using a four-point scale: 3 (faithful extraction), 2 (minor boundary errors), 1 (major extraction errors), and 0 (failed or empty extraction). The mean score was 2.9 out of 3.0, demonstrating high extraction fidelity using our detect-then-extract pipeline with a local LLM. \textbf{(2) Canonical normalization accuracy.}
We evaluated whether extracted EEG section headers were correctly mapped to their canonical categories. Across 80 evaluated sections spanning all seven canonical categories, the normalization procedure achieved 100\% accuracy. These results support the reliability of the section extraction and normalization procedures used in constructing our benchmark.

\subsection{Qualitative Analysis and Case Studies}

\begin{figure}[thpb]
    \centering
     \includegraphics[width=\linewidth]{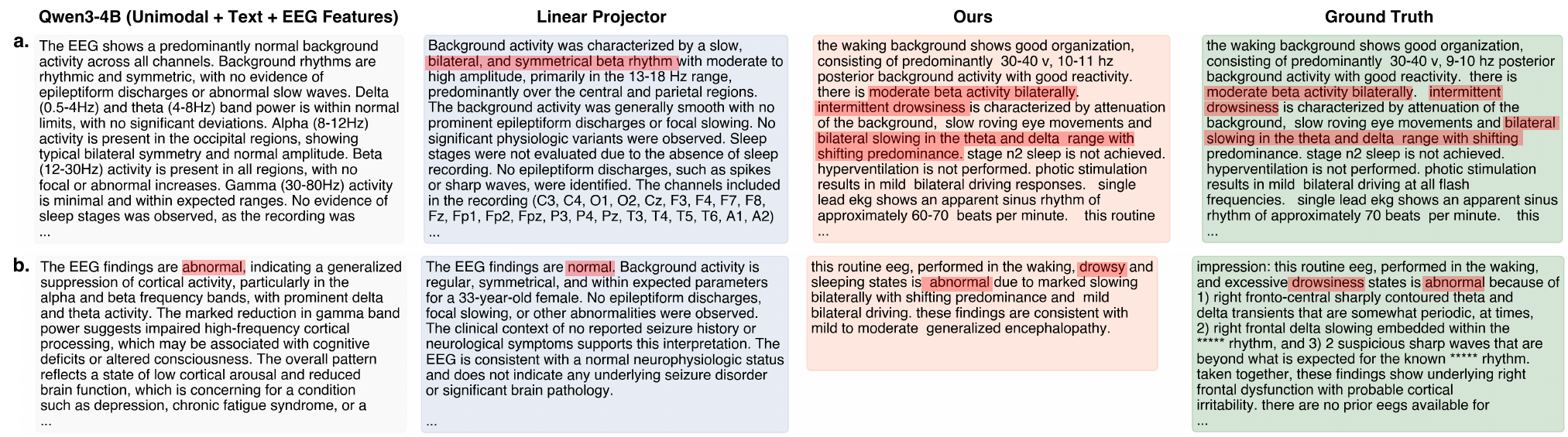}
    \caption{Examples of generated reports on S0002.
Comparisons between the unimodal baseline (text + EEG features), the linear-projector alignment variant, CELM, and the ground-truth reports. (a) EEG description/details; (b) Impression/interpretation.
    }
    \label{fig:qualitative_main_1}
    \vspace{-0.3cm}

\end{figure}
\begin{figure}[ht]
    \centering
     \includegraphics[width=\linewidth]{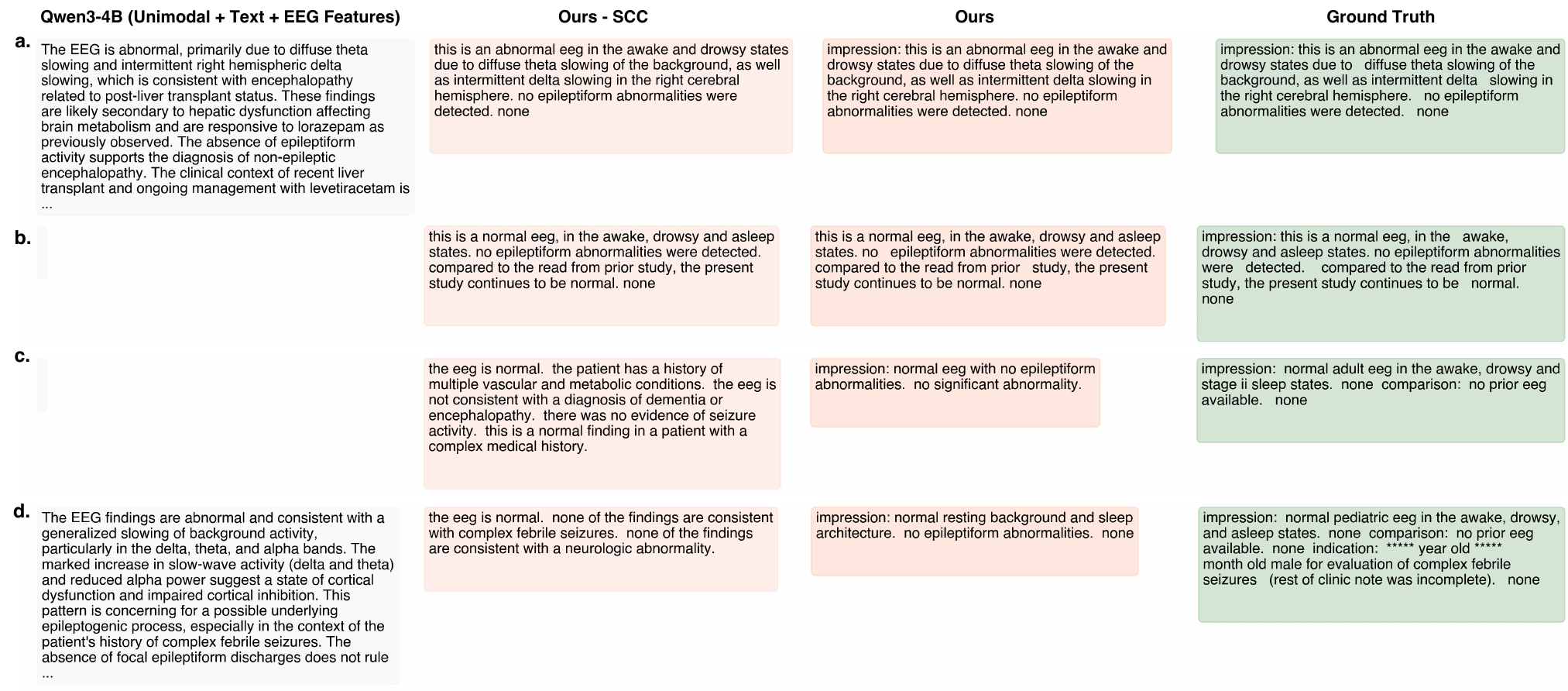}
    \caption{Impression/interpretation generation examples on S0001. Examples are ordered from (a) to (d) by decreasing ROUGE-1 score. We compare outputs from the unimodal baseline (text + EEG features), CELM-SCC, CELM, and the ground-truth reports.
    }
    \label{fig:2_S0002_impression_diff_rouge_scores}

\end{figure}
\begin{figure}[h]
    \centering
     \includegraphics[width=\linewidth]{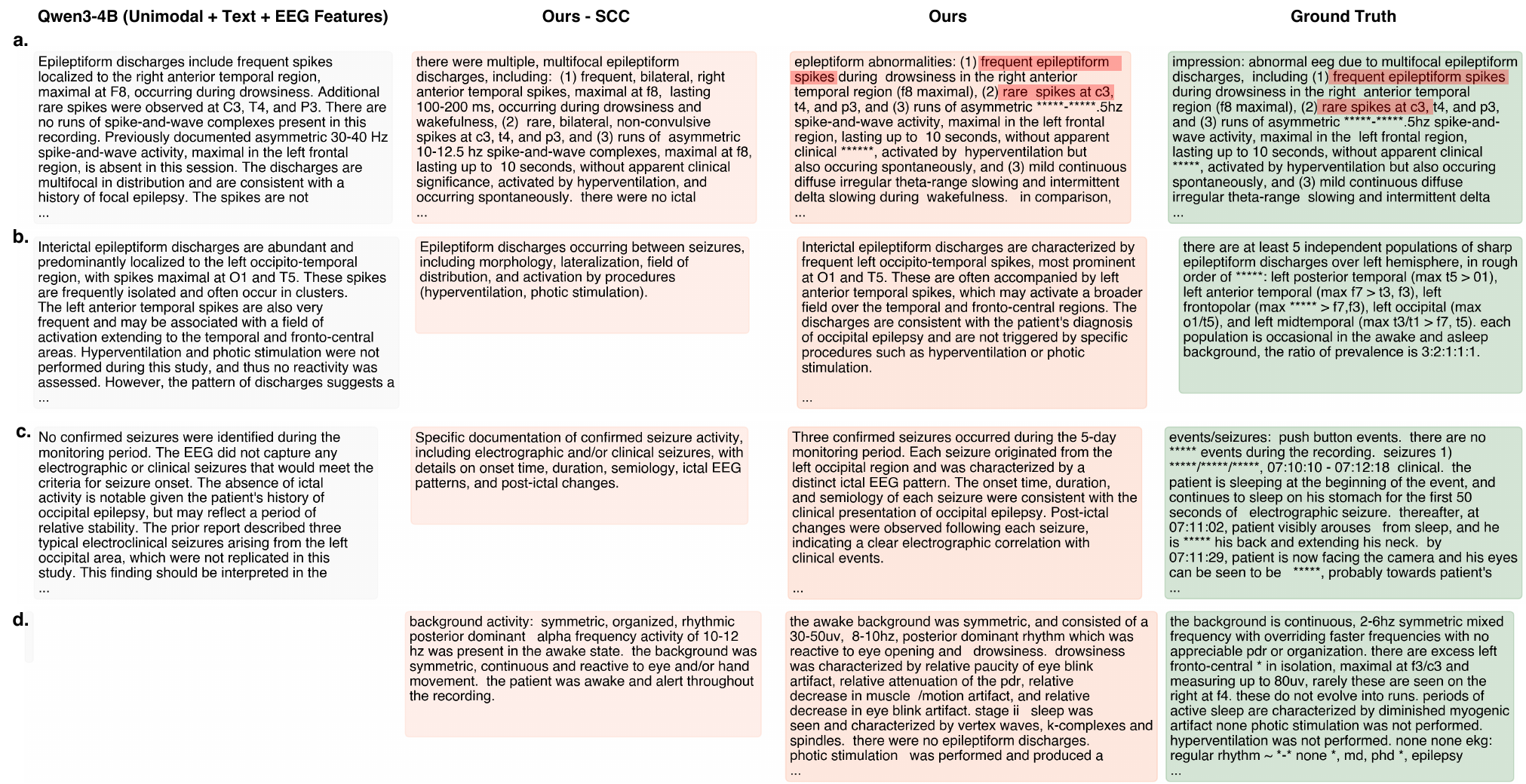}
    \caption{Section-wise generation examples on S0001.
(a) Epileptiform abnormalities, (b) Interictal epileptiform abnormalities, (c) Seizures, and (d) Background activity. We compare outputs from the unimodal baseline (text + EEG features), CELM-SCC, CELM, and the ground-truth reports.
    }
    \label{fig:3_S0001_others}

\end{figure}
\begin{figure}[ht]
    \centering
     \includegraphics[width=\linewidth]{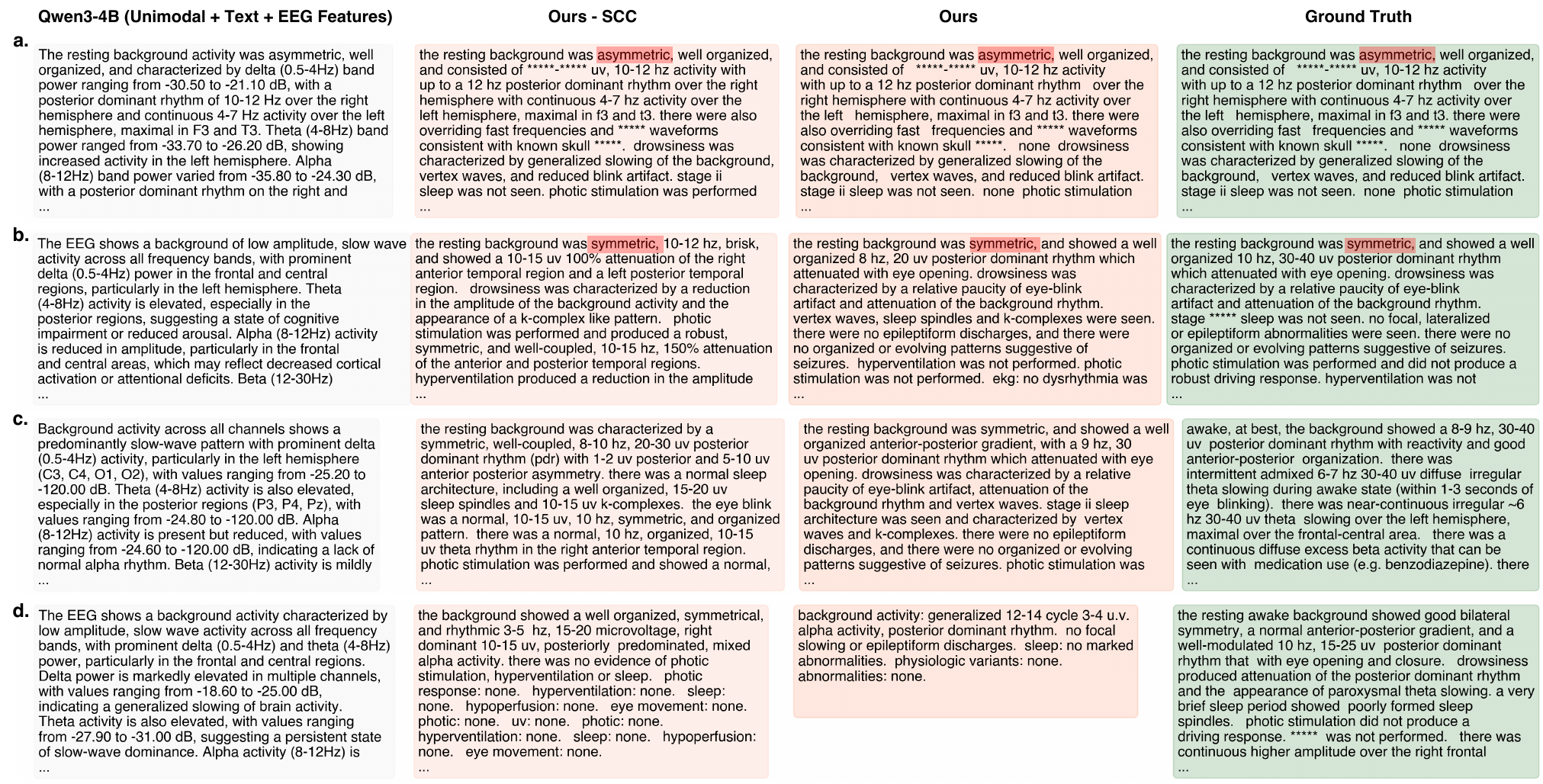}
    \caption{EEG description/details generation examples on S0001. Examples are ordered from (a) to (d) by decreasing ROUGE-1 score. We compare outputs from the unimodal baseline (text + EEG features), CELM-SCC, CELM, and the ground-truth reports.
    }
    \label{fig:1_S0001_eeg_desc_diff_rouge_scores}

\end{figure}
\begin{figure}[h]
    \centering
     \includegraphics[width=\linewidth]{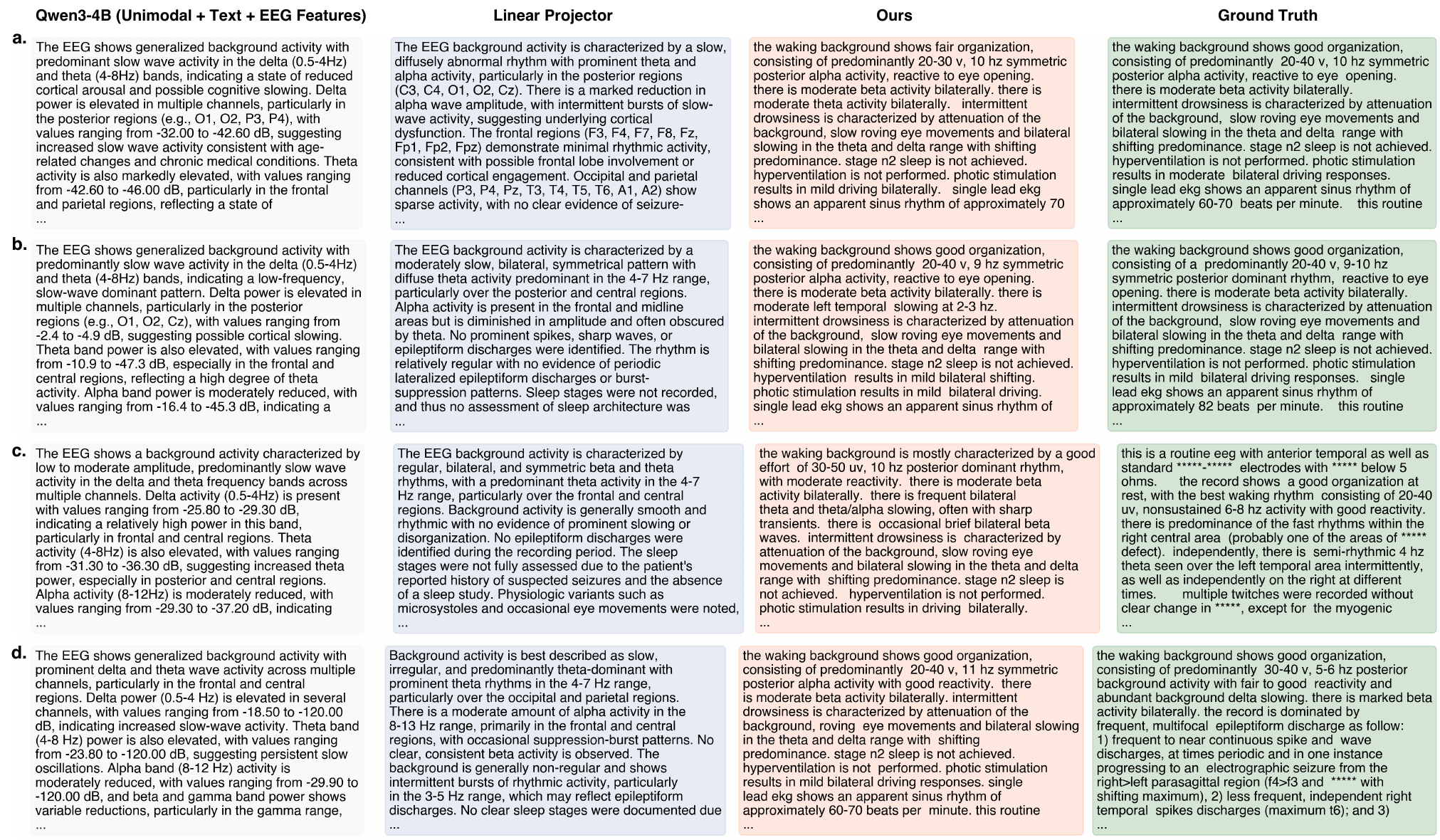}
    \caption{EEG description/details generation examples on S0002. Examples are ordered from (a) to (d) by decreasing ROUGE-1 score. We compare outputs from the unimodal baseline (text + EEG features), CELM-SCC, CELM, and the ground-truth reports.
    }
    \label{fig:2_S0002_eeg_desc_diff_rouge_scores}

\end{figure}

Figure~\ref{fig:qualitative_main_1} presents qualitative comparisons on S0002 between the best-performing unimodal baseline (text with EEG features), the linear projector variant, and \method, alongside ground truth reports. Figure~\ref{fig:qualitative_main_1}a shows the EEG description/details section, while Figure~\ref{fig:qualitative_main_1}b shows the impression/interpretation section. Across these examples, \method produces reports that are more closely aligned with the ground truth, correctly identifying clinically relevant findings such as moderate bilateral beta activity, intermittent drowsiness, and bilateral slowing in the theta and delta ranges. In contrast, the linear projector and unimodal baselines fail to capture these. In the impression section, \method identifies the recording as abnormal, whereas the linear projector variant incorrectly predicts it as normal. These results demonstrate the promise of ELMs in learning to generate clinical reports from unstructured notes paired with long EEG recordings. At the same time, they underscore the need for more rigorous and standardized benchmarking to assess these models. Additional examples spanning both sites, diverse report sections, and varying performance levels are provided below to further illustrate model strengths and failure modes. In Figure~\ref{fig:2_S0002_impression_diff_rouge_scores},~\ref{fig:3_S0001_others},~\ref{fig:1_S0001_eeg_desc_diff_rouge_scores}, and ~\ref{fig:2_S0002_eeg_desc_diff_rouge_scores}, we compare generated reports from the unimodal baseline, CELM-SCC, and CELM against ground-truth reports across multiple EEG report sections and datasets. 



\clearpage

\section{\camred{Human Expert Evaluation Protocol}}
\label{app:expert}

We describe the evaluation protocol, rating criteria, statistical analysis, and ethical oversight for the human expert evaluation presented in Section~\ref{sec:human_expert_eval_study}.

\paragraph{Evaluators.}
Six experts participated in the evaluation, comprising clinicians and BME researchers with expertise in clinical EEG interpretation, signal analysis, and neurophysiology. Evaluators were recruited from the authors' academic and clinical research networks and participated voluntarily without compensation. All evaluators completed the BDSP access requirements for the Harvard EEG Database, including credentialed access, CITI training, and a Data Use Agreement.

\paragraph{Sample Selection and Evaluation Procedure.}
We randomly selected 20 cases from the held-out test sets across two sites. For each case, evaluators reviewed the reference clinical report alongside three generated reports: CELM, the best \emph{Unimodal + Text + EEG Features Input} baseline (Qwen3-4b-instruct-2507), and the best \emph{Multimodal + Linear Projector + CBraMod} baseline (Qwen3-4b-instruct-2507). This was a blinded study where model identities were concealed. All evaluators received standardized written instructions. For each case, the ground-truth clinical report was presented alongside three anonymized generated reports to minimize evaluation bias. Evaluators reviewed only de-identified report text, without access to raw EEG recordings or patient-identifiable information. Evaluators completed three tasks:

\begin{enumerate}
    \item \textbf{Report rating:} Independently assess each generated report on the five dimensions defined in Table~\ref{tab:expert_dims}, using the specified five-point scales.
    \item \textbf{Preference ranking:} Rank the three generated reports according to overall preference.
    \item \textbf{Optional feedback:} Provide qualitative comments or suggestions for improving the generated reports.
\end{enumerate}

Each evaluator assessed all 20 cases, yielding 60 individual report assessments per evaluator and 360 ratings per dimension across the six evaluators. 

\paragraph{Rating Dimensions and Scale.}
Evaluators assessed each generated report against the corresponding ground-truth clinical report across five dimensions: clinical correctness, findings coverage, finding accuracy, real-world readiness, and completeness. Each dimension was rated on a five-point Likert-type scale, with dimension-specific anchors provided in Table~\ref{tab:expert_dims}. Evaluators additionally ranked the three generated reports by overall preference and could optionally provide qualitative feedback.

\begin{table}[h]
\centering
\small
\caption{Expert evaluation dimensions and rating criteria. All assessments were made with reference to the ground-truth clinical report.}
\label{tab:expert_dims}
\begin{tabular}{p{3.2cm}p{9.5cm}}
\toprule
Dimension & Evaluation criterion and rating scale \\
\midrule
Clinical Correctness &
Correctness of clinical content relative to the ground-truth report.
5: Very good; 4: Good; 3: Acceptable; 2: Poor; 1: Very poor. \\

Findings Coverage &
Extent to which key clinical findings in the ground-truth report are identified.
5: All found; 4: Most found; 3: Some found; 2: Few found; 1: Missed. \\

Finding Accuracy &
Accuracy of finding attributes, including lateralization, frequency, and morphology.
5: All correct; 4: Minor errors; 3: Some errors; 2: Many errors; 1: Inaccurate. \\

Real-World Readiness &
Suitability of the generated report for real-world clinical use.
5: Very good; 4: Good; 3: Acceptable; 2: Poor; 1: Very poor. \\

Completeness &
Extent to which the generated report covers important information.
5: Very good; 4: Good; 3: Acceptable; 2: Poor; 1: Very poor. \\
\bottomrule
\end{tabular}
\end{table}

\clearpage
\section{\camred{Model Card}}
\label{app:modelcard}

We document CELM following the model-card framework from ~\cite{mitchell2019model}.

\begin{small}
\begin{longtable}{p{3.3cm}p{10.0cm}}
\toprule
\endfirsthead
\toprule
\endhead
\bottomrule
\endfoot

\multicolumn{2}{l}{\textbf{Model Details}} \\
\midrule
Developers & Jathurshan Pradeepkumar, Zheng Chen, and Jimeng Sun (University of Illinois
Urbana-Champaign; SANKEN, Osaka University). \\
Model date &  April 2026. \\
Model version & v1.0. Two variants: CELM (Sequence Context Alignment, SCA) and
CELM-SCC (Sequence Context Compression). Separate models for S0001 and S0002\\
Model type & EEG-to-language model for structured clinical report generation. A frozen
pretrained EEG encoder (CBraMod) produces one token per 10-s epoch (Epoch-Aggregated
Tokenization). A trainable sequence-aware alignment module maps these tokens into the
embedding space of a frozen instruction-tuned LLM (Qwen3-4B-Instruct-2507). In SCA this
module is a linear-attention transformer followed by a linear projection and in SCC it is
Perceiver-style cross-attention with learnable queries. The LLM generates report text
conditioned on the EEG tokens, a task prompt, and optional clinical context. \\
Inputs and outputs & Input: scalp EEG resampled to 200\,Hz, 22-channel 10-20 montage in a
fixed channel order, up to $\sim$3 hours long. Optional clinical context (reason for study,
clinical history). Output: free-text report sections (EEG description/details,
impression/interpretation, background activity, epileptiform abnormalities,
events/seizures). \\
Training & Supervised fine-tuning with a next-token prediction objective. Only the alignment
module is updated and the EEG encoder and LLM stay frozen. Hyperparameters and compute are in
Appendix~\ref{app:implementation_details}. No fairness constraints were applied during training. \\
Paper & This paper. Code: \url{https://github.com/Jathurshan0330/CELM}. \\
License & Code: MIT License. Model checkpoints are not publicly
released and access will be gated and subject to the BDSP Credentialed Health Data License
1.5.0. \\
Contact & \texttt{jp65@illinois.edu}, \texttt{chenz@sanken.osaka-u.ac.jp}, or GitHub issues. \\
\midrule

\multicolumn{2}{l}{\textbf{Intended Use}} \\
\midrule
Primary intended uses & Research on EEG--language modeling and on automated drafting of
clinical EEG reports, including benchmarking new methods on the EEG--Report benchmark. \\
Primary intended users & Machine learning and clinical neurophysiology researchers with
credentialed access to the Harvard EEG Database through BDSP. \\
Out-of-scope uses & Any clinical deployment, autonomous clinical reporting, diagnostic or
treatment decisions, triage or prioritization of patient studies. use on patient data
without review by a qualified neurologist, and any use that violates the BDSP
data use agreement, which restricts use to research. \\
\midrule


\multicolumn{2}{l}{\textbf{Metrics}} \\
\midrule
Performance measures & Lexical: BLEU-1/4, ROUGE-1/2/L/Lsum, METEOR. Semantic: BERTScore.
LLM-as-judge (Qwen2.5-72B-Instruct-AWQ, run locally): clinical correctness, findings
coverage, finding accuracy. Expert ratings (1--5) on five clinical dimensions, plus
preference rankings. Manual accuracy on abnormality prediction, lateralization, spectral
dynamics, and epileptiform abnormalities. Lexical metrics are reported for comparability
but do not measure clinical correctness. \\
Decision thresholds & Not applicable: CELM generates text and has no classification
threshold. Its normal/abnormal judgment appears only in the generated text.  \\
Variation approaches & Mean $\pm$ standard deviation across test samples; paired Wilcoxon
signed-rank tests against baselines and perturbation controls; inter-rater ICC(2,$k$),
adjacent agreement, and Kendall's $W$ for expert ratings. \\
\midrule

\multicolumn{2}{l}{\textbf{Evaluation Data}} \\
\midrule
Datasets & Held-out test splits (20\%, patient-wise) of the EEG--Report benchmark built from
the Harvard EEG Database v4.1 and the linked EHR repository on BDSP: S0001 (Massachusetts
General Hospital) and S0002 (Brigham and Women's Hospital). \\
Motivation & The largest available source of paired clinical EEG and reports. The two sites
differ in visit types and report structure, which allows a within- and cross-site
evaluation. \\
Preprocessing & EEG: 0.1--75\,Hz band-pass, 60\,Hz notch, resampling to 200\,Hz, 10-s
epochs, 22-channel 10--20 montage; single-session reports only; S0001 recordings over
10{,}000\,s excluded. Reports: automated structuring (section detection, extraction with a
local Llama-3-8B-Instruct, canonical normalization; Section~\ref{sec:benchmark_constuction}). A
manual audit gave a mean extraction score of 2.9/3 over 150 sections and 100\% correct
normalization. \\
\midrule

\multicolumn{2}{l}{\textbf{Training Data}} \\
\midrule
 & Training (60\%) and validation (20\%) patient-wise splits of the same two sites, with
the same preprocessing. In total, 9{,}922 reports, $\sim$11{,}000 hours of EEG, and
9{,}048 patients (5{,}049 reports from 4{,}669 patients at S0001; 4{,}873 from 4{,}379
at S0002). Duration, age, gender, and report-section distributions are in Table~\ref{tab:dataset_stats} and
Figure~\ref{fig:dataset_statistics}. No clinical data or derived text are redistributed. \\
\midrule


\multicolumn{2}{l}{\textbf{Ethical Considerations}} \\
\midrule
 & The training data are sensitive clinical records. They were de-identified by the data
providers, are accessed under a credentialed data use agreement, and were processed only
with locally hosted models. Because model weights could in principle memorize report
text, checkpoints are not publicly released. Generated reports can omit or invent
findings, used without oversight, they could cause misdiagnosis or encourage automation
bias.  \\
\midrule

\multicolumn{2}{l}{\textbf{Caveats and Recommendations}} \\
\midrule
 & (1)~CELM often classifies mildly abnormal recordings as normal, and it is weaker on
interictal epileptiform abnormalities and other rare findings. Treat a ``normal''
impression with caution. (2)~CELM is sensitive to channel order and inputs must follow the
same 22-channel montage and ordering used in training. (3)~Recordings longer than
$\sim$3 hours are not supported, CELM-SCC handles longer inputs at lower quality. (4)~Intended only for research purposes \\
\end{longtable}
\end{small}

\clearpage







%% file: TABLES/dataset_statistics.tex
\begin{table}[ht]
\centering
\caption{Statistics of the EEG–Report benchmark.}
\label{tab:dataset_stats}

\begin{tabular}{lccc}
\toprule
\textbf{Site } & \textbf{\# of Paired EEG \& Reports }  & \textbf{\# of Patients} & \textbf{Total Duration (hrs)}\\

\midrule
S0001 & 5049 & 4669 & 6639\\
S0002 & 4873 & 4379 & 4367\\
\bottomrule

\end{tabular}

\end{table}

%% file: TABLES/llm_baselines_and_context_length.tex
\begin{wraptable}{r}{0.5\textwidth}
\vspace{-0.3cm}
\centering
\caption{Summary of LLM Baselines}
\label{tab:llm_baselines}
\begin{adjustbox}{max width=\linewidth}
\begin{tabular}{lcc}
\toprule
\textbf{LLM} & \textbf{Family}  & \textbf{Context Length}  \\
\midrule

Gemma-3-1b-it & Gemma 3 & 32K \\
Gemma-3n-E2B-it & Gemma 3 & 32K \\
Gemma-3n-E4B-it & Gemma 3 & 32K \\
Gemma-3-4b-it & Gemma 3 & 128K\\

Medgemma-4b-it & Gemma & 128K \\

Llama-3.2-1B-Instruct  & Llama-3.2 & 128K \\
Llama-3.2-3B-Instruct & Llama-3.2 & 128K\\
Llama-3.1-8B-Instruct & Llama-3.1 & 128K\\
Meta-Llama-3-8B-Instruct & Llama-3 & 8K\\

Qwen3-1.7B & Qwen 3 & 32K\\
Qwen3-4B-Instruct-2507 & Qwen 3 & 256K\\

\bottomrule

\end{tabular}
\end{adjustbox}
\vspace{-0.6cm}
\end{wraptable}

%% file: FIG/unimodal_text_only_prompts.tex
\begin{figure}[ht]
    \centering
    \begin{tcolorbox}[
    colback=gray!5,
    colframe=promptblue,
    title={\mbox{\textbf{Unimodal + Text Only Prompt}}},
    boxrule=0.8pt,
        arc=2mm,
        left=8pt, right=8pt, top=8pt, bottom=8pt,   
    ]

You are an expert clinical neurophysiologist specializing in EEG interpretation and clinical report generation.\\

\textbf{TASK}\\
Your task is to generate the specified sections of a formal clinical EEG report using only the provided:
\begin{itemize}
\vspace{-0.2cm}
    \item Patient history
    \vspace{-0.2cm}
    \item EEG description
\end{itemize}

\textbf{EEG SECTION DESCRIPTIONS} \\
\text{[STANDARDIZED\_SECTION\_DESCRIPTIONS]}
\begin{minted}[ baselinestretch=0.9, frame=single]{text}
e.g. EEG DESCRIPTION/DETAILS: Detailed narrative of EEG findings 
including background activity, sleep stages, physiologic variants, 
and abnormalities observed during the recording period.
\end{minted}

\textbf{GUIDELINES}
\begin{itemize}
\vspace{-0.2cm}
    \item Generate only the sections listed in **SECTIONS TO BE GENERATED**.
    \vspace{-0.2cm}
    \item Do NOT generate any additional sections.
    \vspace{-0.2cm}
    \item Do NOT repeat the same section more than once.
    \vspace{-0.2cm}
    \item Only generate the output in the JSON format and do not include any other text or explanation.
\end{itemize}

\textbf{OUTPUT FORMAT (STRICT)}\\
Return ONLY the following JSON structure, with no preamble, explanation, or markdown:

\begin{minted}[ baselinestretch=0.9, frame=single]{text}
json
{"report_sections": [
    {"section_name": "Name of the section as given in SECTIONS TO BE 
                      GENERATED",
     "section_text": "Generated text for the section in string"},
    ...
]}
\end{minted}

\textbf{SECTIONS TO BE GENERATED}\\
\text{[SECTION\_NAMES]}
\begin{minted}[ baselinestretch=0.9, frame=single]{text}
e.g. ['EEG DESCRIPTION/DETAILS']
\end{minted}

\textbf{PATIENT HISTORY AND EEG DESCRIPTION}\\
\text{[PATIENT\_HISTORY\_AND\_EEG\_DESCRIPTION]}
\begin{minted}[ baselinestretch=0.9, frame=single]{text}
e.g. age: 77.0, gender: Female, indication: ***** y.o. female with 
history of afib, l mca stroke, and headaches, presenting with episodes 
of altered consciousness concerning for seizure vs. syncope.   
none  pertinent medications: keppra, neurontin, seroquel.  none
\end{minted}

Now generate the EEG report.

    \medskip
    
    \end{tcolorbox}
    \caption{Prompt for unimodal + text only baselines.}
    \label{fig:unimodal_text_only_prompt}
    \vspace{-0.2cm}
\end{figure}

%% file: FIG/unimodal_text_EEG_features_prompt.tex
\begin{figure}[hb]
    \centering
    \begin{tcolorbox}[
    colback=gray!5,
    colframe=promptblue,
    title={\mbox{\textbf{Unimodal + Text + EEG Features Prompt}}},
    boxrule=0.8pt,
        arc=2mm,
        left=8pt, right=8pt, top=8pt, bottom=8pt,   
    ]

\textbf{EEG-DERIVED STATISTICS}
\begin{minted}[fontsize=\small, baselinestretch=0.9, frame=single]{text}
{"eeg_session_0":{"delta (0.5-4Hz) band power (dB)": [[-22.90, -21.60, ...]],
                  "theta (4-8Hz) band power (dB)": [[-25.30, -24.80, ...]],
                  "alpha (8-12Hz) band power (dB)": [[-27.50, -26.60, ...]],
                  "beta (12-30Hz) band power (dB)": [[-30.60, -30.20, ...]],
                  "gamma (30-80Hz) band power (dB)": [[-36.30, -36.10, ...]],
\end{minted}
\vspace{-0.2cm}
\textbf{EEG CHANNELS}
\begin{minted}[fontsize=\small, baselinestretch=0.9, frame=single]{text}
['C3', 'C4', 'O1', 'O2', 'Cz', 'F3', 'F4', 'F7', 'F8', 'Fz', 'Fp1', 'Fp2', 
'Fpz', 'P3', 'P4', 'Pz', 'T3', 'T4', 'T5', 'T6', 'A1', 'A2']
\end{minted}
You are an expert clinical neurophysiologist specializing in EEG interpretation and clinical report generation.

\textbf{TASK}\\
Your task is to generate the specified sections of a formal clinical EEG report using only the provided:
\begin{itemize}
\vspace{-0.2cm}
    \item Patient history
    \vspace{-0.2cm}
    \item EEG description
    \vspace{-0.2cm}
    \item EEG Channels
    \vspace{-0.2cm}
    \item EEG-derived statistics (provided above)
    \vspace{-0.2cm}
\end{itemize}

\textbf{EEG SECTION DESCRIPTIONS} \\
\text{[STANDARDIZED\_SECTION\_DESCRIPTIONS]}
\begin{minted}[fontsize=\small, baselinestretch=0.9, frame=single]{text}
e.g. EEG DESCRIPTION/DETAILS: Detailed narrative of EEG findings including 
background activity, sleep stages, physiologic variants, and abnormalities 
observed during the recording period.
\end{minted}
\vspace{-0.1cm}
\textbf{GUIDELINES}
\begin{itemize}
\vspace{-0.2cm}
    \item Generate only the sections listed in **SECTIONS TO BE GENERATED**.
    \vspace{-0.2cm}
    \item Do NOT generate any additional sections.
    \vspace{-0.2cm}
    \item Do NOT repeat the same section more than once.
    \vspace{-0.2cm}
    \item Only generate the output in the JSON format and do not include any other text or explanation.
\end{itemize}

\textbf{OUTPUT FORMAT (STRICT)}\\
Return ONLY the following JSON structure, with no preamble, explanation, or markdown:

\begin{minted}[fontsize=\small, baselinestretch=0.9, frame=single]{text}
json
{"report_sections": [
   {"section_name": "Name of the section as given in SECTIONS TO BE GENERATED",
    "section_text": "Generated text for the section in string"},
    ...]}
\end{minted}

\vspace{-0.2cm}
\textbf{SECTIONS TO BE GENERATED}\\
\text{[SECTION\_NAMES]}
\begin{minted}[ baselinestretch=0.9, frame=single]{text}
e.g. ['EEG DESCRIPTION/DETAILS']
\end{minted}
\vspace{-0.2cm}
\textbf{PATIENT HISTORY AND EEG DESCRIPTION}\\
\text{[PATIENT\_HISTORY\_AND\_EEG\_DESCRIPTION]}
\begin{minted}[fontsize=\small, baselinestretch=0.9, frame=single]{text}
e.g. age: 77.0, gender: Female, indication: ***** y.o. female with history 
of afib, l mca stroke, and headaches, presenting with episodes of altered 
consciousness concerning for seizure vs. syncope.   
none  pertinent medications: keppra, neurontin, seroquel.  none
\end{minted}

Now generate the EEG report. 
    
    \end{tcolorbox}
    \caption{Prompt for unimodal + text and EEG features as input baselines.}
    \label{fig:unimodal_text_and_eeg_features_prompt}
\end{figure}

%% file: TABLES/ELM_hyperparameters.tex
\begin{table}[h]
    \centering
    \caption{Hyperparameters of the alignment (projector) modules.}

    \resizebox{\linewidth}{!}{
    \begin{tabular}{l l c}
        \hline
        \textbf{Module} & \textbf{Hyperparameter} & \textbf{Values} \\
        \hline
        
        \multirow{4}{*}{Linear Projector} 
            & Projector Input Dimension (CBraMod Embedding Dimension) & 200 \\
            & Projector Output Dimension (Qwen-3 4B Embedding Dimension) &  2560 \\
            & Bias & True \\
            & \# Trainable parameters & 522,240 \\
        \hline

        \multirow{7}{*}{Perceiver Projector} 
            & \# Query tokens & $256$  \\
            & Embedding Dimension & 200 \\
            & \# Attention heads & $8$  \\
            & \# Perceiver layers (cross attention followed by feed-forward layer)& $2$  \\
            & Dropout & 0.1 \\
            & Feed-forward multiplier ($\times$) & 2 \\
            & Final linear projection & Linear(200, 2560) \\
            & \# Trainable parameters & 1,219,040 \\
        \hline

        \multirow{8}{*}{Sequence Context Compression (SCC)} 
            & Transformer Encoder Type & Linear attention transformer \\
            & Transformer \# heads & $8$  \\
            & Transformer \# layers (depth) & $1$  \\
            & \# Query tokens  & $256$  \\
            & \# Perceiver layers & $1$  \\
            & \# Trainable parameters  & 1,378,440 \\
        \hline
        
        \multirow{5}{*}{Sequence Context Alignment (SCA)} 
            & Transformer Encoder Type & Linear attention transformer \\
            & \# Heads & $8$ \\
            & \# Transformer Layers & $2$  \\
            & Final linear projection & Linear(200, 2560) \\
            & \# Trainable parameters & 1,486,240 \\

        \hline
    \end{tabular}
    }
    \vspace{-0.3cm}
    \label{tab:elm_params}
\end{table}

%% file: TABLES/ELM_training_parameters.tex
\begin{wraptable}[5]{r}{0.6\linewidth}
\vspace{-1.2cm}
    \centering
    \caption{Training details}
    \begin{tabular}{lc}
        \hline
        \textbf{Hyperparameter}& \textbf{Values}  \\
        \hline
          Batch size & 4 \\
          Gradient accumulation steps  & 4 \\
          Optimizer & AdamW \\
          Learning rate & 1e-4 \\
          Weight decay & 0.01\\
          $\beta_1$ & 0.9\\
          $\beta_2$ & 0.99\\
          LR scheduler & Linear scheduler with warmup \\
          \# of training epochs & 10 \\
          Warm-up ratio & 0.1 \\
          Mixed precision & bf16 \\

        \hline
    \end{tabular}
    \label{tab:elm_training}
    \vspace{-0.5cm}
\end{wraptable}

%% file: FIG/ELM_prompts.tex
\begin{figure}[ht]
    \centering
    \begin{tcolorbox}[
    colback=gray!5,
    colframe=promptblue,
    title={\mbox{\textbf{ELM Prompt}}},
    boxrule=0.8pt,
        arc=2mm,
        left=8pt, right=8pt, top=8pt, bottom=8pt,   
    ]

\textbf{Input :} EEG projected tokens prepended to text tokens.\\

\textbf{EEG CHANNELS}
\begin{minted}[fontsize=\small, baselinestretch=0.9, frame=single]{text}
['C3', 'C4', 'O1', 'O2', 'Cz', 'F3', 'F4', 'F7', 'F8', 'Fz', 'Fp1', 'Fp2', 'Fpz', 
'P3', 'P4', 'Pz', 'T3', 'T4', 'T5', 'T6', 'A1', 'A2']
\end{minted}

You are an expert clinical neurophysiologist specializing in EEG interpretation and clinical report generation.\\

\textbf{TASK}\\
Your task is to generate the specified sections (**SECTIONS TO BE GENERATED**) of a formal clinical EEG report using the above provided data of EEG recording sessions and the following information:
\begin{itemize}
\vspace{-0.2cm}
    \item Patient history
    \vspace{-0.2cm}
    \item EEG description
    \vspace{-0.2cm}
    \item EEG Channels
\end{itemize}

\textbf{EEG SECTION DESCRIPTIONS} \\
\text{[STANDARDIZED\_SECTION\_DESCRIPTIONS]}
\begin{minted}[fontsize=\small, baselinestretch=0.9, frame=single]{text}
e.g. EEG DESCRIPTION/DETAILS: Detailed narrative of EEG findings including 
background activity, sleep stages, physiologic variants, and abnormalities 
observed during the recording period.
\end{minted}
\vspace{-0.2cm}
\textbf{GUIDELINES}
\begin{itemize}
\vspace{-0.2cm}
    \item Generate only the sections listed in **SECTIONS TO BE GENERATED**.
    \vspace{-0.2cm}
    \item Do NOT generate any additional sections.
    \vspace{-0.2cm}
    \item Do NOT repeat the same section more than once.
    \vspace{-0.2cm}
    \item Only generate the output in the JSON format and do not include any other text or explanation.
\end{itemize}

\textbf{OUTPUT FORMAT (STRICT)}\\
Return ONLY the following JSON structure, with no preamble, explanation, or markdown:

\begin{minted}[ baselinestretch=0.9, frame=single]{text}
json
{"report_sections": [
    {"section_name": "Name of the section as given in SECTIONS TO BE 
                      GENERATED",
     "section_text": "Generated text for the section in string"},
    ... ]}
\end{minted}

\vspace{-0.2cm}
\textbf{SECTIONS TO BE GENERATED}\\
\text{[SECTION\_NAMES]}
\begin{minted}[ baselinestretch=0.9, frame=single]{text}
e.g. ['EEG DESCRIPTION/DETAILS']
\end{minted}
\vspace{-0.2cm}
\textbf{PATIENT HISTORY AND EEG DESCRIPTION}\\
\text{[PATIENT\_HISTORY\_AND\_EEG\_DESCRIPTION]}
\begin{minted}[fontsize=\small, baselinestretch=0.9, frame=single]{text}
e.g. age: 77.0, gender: Female, indication: ***** y.o. female with history 
of afib, l mca stroke, and headaches, presenting with episodes of altered 
consciousness concerning for seizure vs. syncope.   
none  pertinent medications: keppra, neurontin, seroquel.  none
\end{minted}

Now generate the EEG report.

    \medskip
    
    \end{tcolorbox}
    \caption{Prompt for CELM.}
    \label{fig:ELM_prompt}
\end{figure}

%% file: TABLES/app_baseline_results_S0001_with_patient_history.tex
\begin{table}[ht]

\centering
\caption{Report generation performance comparison on samples with clinical context on S0001.
We provide results for a complete set of evaluation metrics under two input settings:
(i) \emph{Unimodal + Text Only Input}, where language models generate reports solely from clinical context text, and
(ii) \emph{Unimodal + Text + EEG Features Input}, where EEG-derived features are additionally provided.
The table compares multiple strong baselines, including general-purpose and medical LLMs, against our \method.
The best results are highlighted in \textcolor{highlightorange}{orange}, and the best baseline per category is highlighted in \textcolor{highlightblue}{blue}.
}
\label{tab:app_baseline_results_S0001_with_PH}
\begin{adjustbox}{max width=\textwidth}

\setlength{\tabcolsep}{4pt} 
\renewcommand{\arraystretch}{1.2}
\begin{tabular}{lcccccccc}
\toprule

\textbf{Method}   & \textbf{BLEU-1}     &\textbf{BLEU-4}  &\textbf{ROUGE-1} &\textbf{ROUGE-2}  & \textbf{ROUGE-L}  & \textbf{ROUGE-LSUM}    & \textbf{BERTScore} &\textbf{METEOR}  \\
\midrule
 &\multicolumn{8}{c}{\textbf{Unimodal + Text Only Input}} \\
\cmidrule(lr){2-9}

Gemma-3-1b-it& 0.1897 $\pm$ 0.1340& 0.0351 $\pm$ 0.0694& 0.2112 $\pm$ 0.1408& 0.0431 $\pm$ 0.0985& 0.1356 $\pm$ 0.1129& 0.1381 $\pm$ 0.1137& 0.6461 $\pm$ 0.3014& 0.1567 $\pm$ 0.1238\\
Gemma-3n-e2b-it& 0.1967 $\pm$ 0.1592& 0.0641 $\pm$ 0.1402& 0.2410 $\pm$ 0.1722& 0.0702 $\pm$ 0.1939& 0.1657 $\pm$ 0.1826& 0.1687 $\pm$ 0.1824& 0.7848 $\pm$ 0.0765& 0.1875 $\pm$ 0.1725\\
Gemma-3-4b-it& 0.2372 $\pm$ 0.1508& 0.0641 $\pm$ 0.1349& 0.2857 $\pm$ 0.1590& \cellcolor{highlightblue!35} 0.0821 $\pm$ 0.1893& \cellcolor{highlightblue!35} 0.1893 $\pm$ 0.1749& \cellcolor{highlightblue!35} 0.1929 $\pm$ 0.1747& 0.7941 $\pm$ 0.0922& 0.2147 $\pm$ 0.1668\\
Gemma-3n-e4b-it& 0.1793 $\pm$ 0.1685& \cellcolor{highlightblue!35} 0.0642 $\pm$ 0.1486& 0.2416 $\pm$ 0.1778& 0.0763 $\pm$ 0.2003& 0.1659 $\pm$ 0.1892& 0.1687 $\pm$ 0.1890& 0.7847 $\pm$ 0.0747& 0.1816 $\pm$ 0.1786\\
Medgemma-4b-it& 0.1502 $\pm$ 0.1725& 0.0485 $\pm$ 0.1383& 0.2196 $\pm$ 0.1924& 0.0654 $\pm$ 0.1954& 0.1569 $\pm$ 0.1908& 0.1593 $\pm$ 0.1910& 0.6683 $\pm$ 0.2994& 0.1529 $\pm$ 0.1835\\
Llama-3.2-1b-instruct& 0.0337 $\pm$ 0.0667& 0.0059 $\pm$ 0.0166& 0.0919 $\pm$ 0.0744& 0.0053 $\pm$ 0.0217& 0.0615 $\pm$ 0.0500& 0.0629 $\pm$ 0.0515& 0.5580 $\pm$ 0.3284& 0.0529 $\pm$ 0.0519\\
Llama-3.2-3b-instruct& 0.1951 $\pm$ 0.1630& 0.0625 $\pm$ 0.1487& 0.2224 $\pm$ 0.1755& 0.0669 $\pm$ 0.1860& 0.1511 $\pm$ 0.1760& 0.1550 $\pm$ 0.1759& 0.7116 $\pm$ 0.2311& 0.1843 $\pm$ 0.1864\\
Llama-3.1-8b-instruct& 0.2128 $\pm$ 0.1702& 0.0598 $\pm$ 0.1444& 0.2511 $\pm$ 0.1872& 0.0753 $\pm$ 0.1967& 0.1689 $\pm$ 0.1889& 0.1722 $\pm$ 0.1892& 0.6994 $\pm$ 0.2605& 0.1839 $\pm$ 0.1788\\
Meta-llama-3-8b-instruct& 0.2249 $\pm$ 0.1602& 0.0620 $\pm$ 0.1418& 0.2673 $\pm$ 0.1731& 0.0765 $\pm$ 0.1895& 0.1758 $\pm$ 0.1785& 0.1795 $\pm$ 0.1787& 0.7392 $\pm$ 0.2055& 0.1946 $\pm$ 0.1809\\
Qwen3-1.7b& 0.1335 $\pm$ 0.1481& 0.0213 $\pm$ 0.0450& 0.1378 $\pm$ 0.1557& 0.0285 $\pm$ 0.0704& 0.0835 $\pm$ 0.1054& 0.0855 $\pm$ 0.1075& 0.3856 $\pm$ 0.3941& 0.1102 $\pm$ 0.1316\\
Qwen3-4b-instruct-2507& \cellcolor{highlightblue!35} 0.2795 $\pm$ 0.1243& 0.0593 $\pm$ 0.1100& \cellcolor{highlightblue!35} 0.3038 $\pm$ 0.1460& 0.0763 $\pm$ 0.1672& 0.1872 $\pm$ 0.1623& 0.1916 $\pm$ 0.1622& \cellcolor{highlightblue!35} 0.7975 $\pm$ 0.0852& \cellcolor{highlightblue!35} 0.2418 $\pm$ 0.1440\\

\midrule

& \multicolumn{8}{c}{\textbf{Unimodal + Text + EEG Features Input}} \\
\cmidrule(lr){2-9}

Gemma-3-1b-it& 0.0068 $\pm$ 0.0442& 0.0019 $\pm$ 0.0273& 0.0086 $\pm$ 0.0524& 0.0022 $\pm$ 0.0367& 0.0063 $\pm$ 0.0449& 0.0064 $\pm$ 0.0451& 0.0311 $\pm$ 0.1512& 0.0064 $\pm$ 0.0462\\
Gemma-3n-e2b-it& 0.0083 $\pm$ 0.0504& 0.0029 $\pm$ 0.0330& 0.0116 $\pm$ 0.0605& 0.0032 $\pm$ 0.0403& 0.0080 $\pm$ 0.0500& 0.0083 $\pm$ 0.0507& 0.0415 $\pm$ 0.1756& 0.0088 $\pm$ 0.0543\\
Gemma-3-4b-it& 0.0085 $\pm$ 0.0532& 0.0025 $\pm$ 0.0299& 0.0093 $\pm$ 0.0594& 0.0032 $\pm$ 0.0402& 0.0062 $\pm$ 0.0466& 0.0065 $\pm$ 0.0474& 0.0247 $\pm$ 0.1384& 0.0078 $\pm$ 0.0558\\
Gemma-3n-e4b-it& 0.0101 $\pm$ 0.0348& 0.0020 $\pm$ 0.0067& 0.0397 $\pm$ 0.0606& 0.0014 $\pm$ 0.0072& 0.0259 $\pm$ 0.0395& 0.0262 $\pm$ 0.0401& 0.2498 $\pm$ 0.3570& 0.0231 $\pm$ 0.0392\\
Medgemma-4b-it& 0.1618 $\pm$ 0.1730& 0.0488 $\pm$ 0.1400& 0.2100 $\pm$ 0.1919& 0.0611 $\pm$ 0.1870& 0.1455 $\pm$ 0.1858& 0.1482 $\pm$ 0.1862& 0.6172 $\pm$ 0.3319& 0.1529 $\pm$ 0.1832\\
Llama-3.2-1b-instruct& 0.0310 $\pm$ 0.0615& 0.0053 $\pm$ 0.0119& 0.0820 $\pm$ 0.0727& 0.0037 $\pm$ 0.0150& 0.0547 $\pm$ 0.0485& 0.0556 $\pm$ 0.0493& 0.4978 $\pm$ 0.3562& 0.0473 $\pm$ 0.0479\\
Llama-3.2-3b-instruct& 0.1692 $\pm$ 0.1414& \cellcolor{highlightblue!35} 0.0528 $\pm$ 0.1147& 0.1844 $\pm$ 0.1611& 0.0543 $\pm$ 0.1575& 0.1271 $\pm$ 0.1529& 0.1342 $\pm$ 0.1536& 0.6441 $\pm$ 0.2898& 0.1751 $\pm$ 0.1871\\
Llama-3.1-8b-instruct& 0.2040 $\pm$ 0.1450& 0.0508 $\pm$ 0.1177& \cellcolor{highlightblue!35} 0.2151 $\pm$ 0.1693& \cellcolor{highlightblue!35} 0.0617 $\pm$ 0.1695& \cellcolor{highlightblue!35} 0.1469 $\pm$ 0.1690& \cellcolor{highlightblue!35} 0.1499 $\pm$ 0.1693& 0.6930 $\pm$ 0.2289& 0.1844 $\pm$ 0.1647\\
Qwen3-1.7b& \cellcolor{highlightblue!35} 0.2047 $\pm$ 0.1351& 0.0331 $\pm$ 0.0611& 0.2051 $\pm$ 0.1453& 0.0422 $\pm$ 0.0971& 0.1286 $\pm$ 0.1150& 0.1317 $\pm$ 0.1163& 0.6063 $\pm$ 0.3215& 0.1683 $\pm$ 0.1304\\
Qwen3-4b-instruct-2507& 0.1809 $\pm$ 0.0986& 0.0288 $\pm$ 0.0555& 0.2058 $\pm$ 0.1192& 0.0402 $\pm$ 0.0969& 0.1275 $\pm$ 0.1082& 0.1303 $\pm$ 0.1087& \cellcolor{highlightblue!35} 0.7249 $\pm$ 0.1840& \cellcolor{highlightblue!35} 0.2067 $\pm$ 0.1079\\

\midrule

& \multicolumn{8}{c}{\textbf{Multimodal + Linear Projector + CBraMod}} \\
\cmidrule(lr){2-9}

Gemma-3-4b-it& 0.2201 $\pm$ 0.1349& \cellcolor{highlightblue!35} 0.0552 $\pm$ 0.1087 & 0.2688 $\pm$ 0.1448& \cellcolor{highlightblue!35} 0.0726 $\pm$ 0.1652 & \cellcolor{highlightblue!35} 0.1803 $\pm$ 0.1558 & \cellcolor{highlightblue!35} 0.1834 $\pm$ 0.1556 & 0.7904 $\pm$ 0.0798& 0.2020 $\pm$ 0.1579\\
Medgemma-4b-it& 0.0392 $\pm$ 0.1478& 0.0260 $\pm$ 0.1212& 0.0494 $\pm$ 0.1806& 0.0356 $\pm$ 0.1668& 0.0445 $\pm$ 0.1764& 0.0447 $\pm$ 0.1765& 0.0886 $\pm$ 0.2601& 0.0441 $\pm$ 0.1719\\
Llama-3.2-3B-instruct& 0.0312 $\pm$ 0.1055& 0.0146 $\pm$ 0.0816& 0.0568 $\pm$ 0.1311& 0.0182 $\pm$ 0.1059& 0.0446 $\pm$ 0.1193& 0.0448 $\pm$ 0.1192& 0.2402 $\pm$ 0.3583& 0.0396 $\pm$ 0.1172\\
Llama-3.1-8B-instruct& 0.0042 $\pm$ 0.0312& 0.0017 $\pm$ 0.0190& 0.0063 $\pm$ 0.0428& 0.0029 $\pm$ 0.0284& 0.0050 $\pm$ 0.0368& 0.0051 $\pm$ 0.0370& 0.0189 $\pm$ 0.1169& 0.0050 $\pm$ 0.0369\\
Qwen3-4B-instruct-2507& \cellcolor{highlightblue!35} 0.2647 $\pm$ 0.0908 & 0.0420 $\pm$ 0.0643& \cellcolor{highlightblue!35} 0.2819 $\pm$ 0.1132 & 0.0595 $\pm$ 0.1139& 0.1676 $\pm$ 0.1183& 0.1720 $\pm$ 0.1187& \cellcolor{highlightblue!35} 0.7921 $\pm$ 0.0417 & \cellcolor{highlightblue!35} 0.2444 $\pm$ 0.1093 \\

\midrule



\textbf{CELM-SCC}& \textbf{0.3383} $\pm$ 0.1936& \textbf{0.1345} $\pm$ 0.1860& \textbf{0.3843} $\pm$ 0.1876& \textbf{0.1651} $\pm$ 0.1960& \textbf{0.2699} $\pm$ 0.1867& \textbf{0.2732} $\pm$ 0.1865& \textbf{0.7949} $\pm$ 0.1783& \textbf{0.2889} $\pm$ 0.1866\\

\textbf{CELM}& \cellcolor{highlightorange!35}\textbf{ 0.4823} $\pm$ 0.1920 & \cellcolor{highlightorange!35}\textbf{ 0.2831} $\pm$ 0.1952 & \cellcolor{highlightorange!35}\textbf{ 0.5565} $\pm$ 0.1683 & \cellcolor{highlightorange!35}\textbf{ 0.3458} $\pm$ 0.2036 & \cellcolor{highlightorange!35}\textbf{ 0.4687} $\pm$ 0.1857 & \cellcolor{highlightorange!35}\textbf{ 0.4702} $\pm$ 0.1852 & \cellcolor{highlightorange!35}\textbf{ 0.8697} $\pm$ 0.0901 & \cellcolor{highlightorange!35}\textbf{ 0.4734} $\pm$ 0.1941 \\






\bottomrule

\end{tabular}
\end{adjustbox}

\end{table}

%% file: TABLES/app_baseline_results_S0002_with_patient_history.tex
\begin{table}[ht]

\centering
\caption{Report generation performance comparison on samples with clinical context on S0002.
We provide results for a complete set of evaluation metrics under two input settings:
(i) \emph{Unimodal + Text Only Input}, where language models generate reports solely from clinical context text, and
(ii) \emph{Unimodal + Text + EEG Features Input}, where EEG-derived features are additionally provided.
The table compares multiple strong baselines, including general-purpose and medical LLMs, against our \method.
The best results are highlighted in \textcolor{highlightorange}{orange}, and the best baseline per category is highlighted in \textcolor{highlightblue}{blue}.
}
\label{tab:app_baseline_results_S0002_with_PH}
\begin{adjustbox}{max width=\textwidth}

\setlength{\tabcolsep}{4pt} 
\renewcommand{\arraystretch}{1.2}
\begin{tabular}{lcccccccc}
\toprule

\textbf{Method}   & \textbf{BLEU-1}     &\textbf{BLEU-4}  &\textbf{ROUGE-1} &\textbf{ROUGE-2}  & \textbf{ROUGE-L}  & \textbf{ROUGE-LSUM}    & \textbf{BERTScore} &\textbf{METEOR}  \\
\midrule
 &\multicolumn{8}{c}{\textbf{Unimodal + Text Only Input}} \\
\cmidrule(lr){2-9}

Gemma-3-1b-it& 0.1509 $\pm$ 0.0898& 0.0158 $\pm$ 0.0121& 0.2330 $\pm$ 0.0876& 0.0302 $\pm$ 0.0210& 0.1233 $\pm$ 0.0467& 0.1259 $\pm$ 0.0495& 0.7058 $\pm$ 0.2423& 0.1319 $\pm$ 0.0565\\
Gemma-3n-e2b-it& 0.0925 $\pm$ 0.0769& 0.0115 $\pm$ 0.0230& 0.2318 $\pm$ 0.0600& 0.0274 $\pm$ 0.0346& 0.1224 $\pm$ 0.0377& 0.1254 $\pm$ 0.0408& 0.7683 $\pm$ 0.0560& 0.1181 $\pm$ 0.0472\\
Gemma-3-4b-it& 0.1538 $\pm$ 0.0990& 0.0163 $\pm$ 0.0180& \cellcolor{highlightblue!35} 0.2775 $\pm$ 0.0558& 0.0354 $\pm$ 0.0335& \cellcolor{highlightblue!35} 0.1407 $\pm$ 0.0336& \cellcolor{highlightblue!35} 0.1440 $\pm$ 0.0369& 0.7851 $\pm$ 0.0730& 0.1506 $\pm$ 0.0514\\
Gemma-3n-e4b-it& 0.0539 $\pm$ 0.0664& 0.0083 $\pm$ 0.0243& 0.1998 $\pm$ 0.0649& 0.0281 $\pm$ 0.0358& 0.1152 $\pm$ 0.0392& 0.1175 $\pm$ 0.0407& 0.7635 $\pm$ 0.0525& 0.0993 $\pm$ 0.0490\\
Medgemma-4b-it& 0.0865 $\pm$ 0.0855& 0.0109 $\pm$ 0.0194& 0.2280 $\pm$ 0.0734& 0.0341 $\pm$ 0.0341& 0.1315 $\pm$ 0.0433& 0.1345 $\pm$ 0.0458& 0.7576 $\pm$ 0.1375& 0.1198 $\pm$ 0.0542\\
Llama-3.2-1b-instruct& 0.0340 $\pm$ 0.0700& 0.0036 $\pm$ 0.0073& 0.0954 $\pm$ 0.0869& 0.0082 $\pm$ 0.0135& 0.0581 $\pm$ 0.0483& 0.0596 $\pm$ 0.0509& 0.5139 $\pm$ 0.3501& 0.0494 $\pm$ 0.0504\\
Llama-3.2-3b-instruct& 0.1651 $\pm$ 0.1110& 0.0186 $\pm$ 0.0256& 0.2346 $\pm$ 0.0877& 0.0321 $\pm$ 0.0351& 0.1221 $\pm$ 0.0478& 0.1271 $\pm$ 0.0517& 0.7261 $\pm$ 0.1919& 0.1430 $\pm$ 0.0689\\
Llama-3.1-8b-instruct& 0.1629 $\pm$ 0.0970& 0.0190 $\pm$ 0.0237& 0.2655 $\pm$ 0.0766& \cellcolor{highlightblue!35} 0.0424 $\pm$ 0.0356& 0.1368 $\pm$ 0.0444& 0.1404 $\pm$ 0.0476& 0.7502 $\pm$ 0.1531& 0.1480 $\pm$ 0.0581\\
Meta-llama-3-8b-instruct& 0.1441 $\pm$ 0.0847& 0.0176 $\pm$ 0.0232& 0.2588 $\pm$ 0.0586& 0.0403 $\pm$ 0.0330& 0.1354 $\pm$ 0.0374& 0.1390 $\pm$ 0.0411& 0.7702 $\pm$ 0.0817& 0.1416 $\pm$ 0.0481\\
Qwen3-1.7b& 0.0886 $\pm$ 0.1095& 0.0097 $\pm$ 0.0124& 0.1189 $\pm$ 0.1296& 0.0142 $\pm$ 0.0191& 0.0633 $\pm$ 0.0686& 0.0652 $\pm$ 0.0716& 0.3650 $\pm$ 0.3893& 0.0735 $\pm$ 0.0821\\
Qwen3-4b-instruct-2507& \cellcolor{highlightblue!35} 0.1912 $\pm$ 0.0807& \cellcolor{highlightblue!35} 0.0218 $\pm$ 0.0197& 0.2690 $\pm$ 0.0448& 0.0398 $\pm$ 0.0298& 0.1317 $\pm$ 0.0295& 0.1350 $\pm$ 0.0324& \cellcolor{highlightblue!35} 0.7867 $\pm$ 0.0178& \cellcolor{highlightblue!35} 0.1622 $\pm$ 0.0434\\

\midrule

& \multicolumn{8}{c}{\textbf{Unimodal + Text + EEG Features Input}} \\
\cmidrule(lr){2-9}

Gemma-3-1b-it& 0.0957 $\pm$ 0.0925& 0.0109 $\pm$ 0.0203& 0.1755 $\pm$ 0.1157& 0.0239 $\pm$ 0.0316& 0.0980 $\pm$ 0.0653& 0.1007 $\pm$ 0.0684& 0.5785 $\pm$ 0.3384& 0.0954 $\pm$ 0.0692\\
Gemma-3n-e2b-it& 0.0750 $\pm$ 0.0771& 0.0093 $\pm$ 0.0226& 0.2190 $\pm$ 0.0840& 0.0350 $\pm$ 0.0350& 0.1256 $\pm$ 0.0516& 0.1281 $\pm$ 0.0539& 0.7175 $\pm$ 0.2153& 0.1082 $\pm$ 0.0546\\
Gemma-3-4b-it& 0.2089 $\pm$ 0.0895& 0.0222 $\pm$ 0.0186& \cellcolor{highlightblue!35} 0.2886 $\pm$ 0.0458& \cellcolor{highlightblue!35} 0.0446 $\pm$ 0.0331& \cellcolor{highlightblue!35} 0.1451 $\pm$ 0.0315& \cellcolor{highlightblue!35} 0.1491 $\pm$ 0.0356& \cellcolor{highlightblue!35} 0.7883 $\pm$ 0.0418& 0.1738 $\pm$ 0.0476\\
Gemma-3n-e4b-it& 0.0169 $\pm$ 0.0486& 0.0018 $\pm$ 0.0052& 0.0357 $\pm$ 0.0903& 0.0049 $\pm$ 0.0138& 0.0194 $\pm$ 0.0490& 0.0207 $\pm$ 0.0526& 0.1073 $\pm$ 0.2678& 0.0180 $\pm$ 0.0462\\
Medgemma-4b-it& 0.0888 $\pm$ 0.0910& 0.0110 $\pm$ 0.0195& 0.1829 $\pm$ 0.1128& 0.0248 $\pm$ 0.0349& 0.1008 $\pm$ 0.0641& 0.1029 $\pm$ 0.0660& 0.5979 $\pm$ 0.3260& 0.1000 $\pm$ 0.0702\\
Llama-3.2-1b-instruct& 0.0312 $\pm$ 0.0668& 0.0035 $\pm$ 0.0073& 0.1016 $\pm$ 0.0773& 0.0090 $\pm$ 0.0142& 0.0620 $\pm$ 0.0420& 0.0634 $\pm$ 0.0438& 0.5903 $\pm$ 0.3062& 0.0517 $\pm$ 0.0448\\
Llama-3.2-3b-instruct& 0.1779 $\pm$ 0.1101& 0.0207 $\pm$ 0.0224& 0.2309 $\pm$ 0.0903& 0.0350 $\pm$ 0.0349& 0.1217 $\pm$ 0.0484& 0.1364 $\pm$ 0.0575& 0.7377 $\pm$ 0.1557& 0.1540 $\pm$ 0.0743\\
Llama-3.1-8b-instruct& 0.1960 $\pm$ 0.0959& \cellcolor{highlightblue!35} 0.0227 $\pm$ 0.0251& 0.2256 $\pm$ 0.0904& 0.0365 $\pm$ 0.0364& 0.1198 $\pm$ 0.0542& 0.1236 $\pm$ 0.0574& 0.6895 $\pm$ 0.2171& 0.1509 $\pm$ 0.0682\\
Qwen3-1.7b& 0.1890 $\pm$ 0.1048& 0.0185 $\pm$ 0.0118& 0.2049 $\pm$ 0.1021& 0.0236 $\pm$ 0.0179& 0.1077 $\pm$ 0.0530& 0.1110 $\pm$ 0.0568& 0.6345 $\pm$ 0.2944& 0.1413 $\pm$ 0.0724\\
Qwen3-4b-instruct-2507& \cellcolor{highlightblue!35} 0.2483 $\pm$ 0.0425& 0.0217 $\pm$ 0.0098& 0.2638 $\pm$ 0.0341& 0.0334 $\pm$ 0.0179& 0.1283 $\pm$ 0.0180& 0.1322 $\pm$ 0.0250& 0.7704 $\pm$ 0.0125& \cellcolor{highlightblue!35} 0.2017 $\pm$ 0.0285\\

\midrule

& \multicolumn{8}{c}{\textbf{Multimodal + Linear Projector + CBraMod}} \\
\cmidrule(lr){2-9}

Gemma-3-4b-it& 0.2683 $\pm$ 0.0751& 0.0253 $\pm$ 0.0224& \cellcolor{highlightblue!35} 0.3084 $\pm$ 0.0491 & 0.0491 $\pm$ 0.0328& \cellcolor{highlightblue!35} 0.1500 $\pm$ 0.0331 & \cellcolor{highlightblue!35} 0.1539 $\pm$ 0.0374 & \cellcolor{highlightblue!35} 0.7879 $\pm$ 0.0729 & 0.2002 $\pm$ 0.0482\\
Medgemma-4b-it& 0.0533 $\pm$ 0.0795& 0.0058 $\pm$ 0.0084& 0.1491 $\pm$ 0.1349& 0.0202 $\pm$ 0.0215& 0.0863 $\pm$ 0.0772& 0.0878 $\pm$ 0.0788& 0.4492 $\pm$ 0.3899& 0.0729 $\pm$ 0.0701\\
Llama-3.2-3B-instruct& 0.1529 $\pm$ 0.1010& 0.0169 $\pm$ 0.0217& 0.2425 $\pm$ 0.0837& 0.0344 $\pm$ 0.0333& 0.1327 $\pm$ 0.0486& 0.1360 $\pm$ 0.0511& 0.7333 $\pm$ 0.1900& 0.1398 $\pm$ 0.0610\\
Llama-3.1-8B-instruct& 0.0169 $\pm$ 0.0492& 0.0018 $\pm$ 0.0051& 0.0385 $\pm$ 0.0823& 0.0041 $\pm$ 0.0114& 0.0231 $\pm$ 0.0480& 0.0238 $\pm$ 0.0500& 0.1533 $\pm$ 0.3011& 0.0209 $\pm$ 0.0458\\
Qwen3-4B-instruct-2507& \cellcolor{highlightblue!35} 0.2775 $\pm$ 0.0485 & \cellcolor{highlightblue!35} 0.0283 $\pm$ 0.0121 & 0.2998 $\pm$ 0.0425& \cellcolor{highlightblue!35} 0.0507 $\pm$ 0.0240 & 0.1429 $\pm$ 0.0239& 0.1473 $\pm$ 0.0306& 0.7852 $\pm$ 0.0153& \cellcolor{highlightblue!35} 0.2143 $\pm$ 0.0349 \\

\midrule



\textbf{CELM-SCC}& \textbf{0.3767} $\pm$ 0.1557& \textbf{0.1671} $\pm$ 0.1054& \textbf{0.4487} $\pm$ 0.1283& \textbf{0.2149} $\pm$ 0.1109& \textbf{0.3232} $\pm$ 0.1263& \textbf{0.3268} $\pm$ 0.1258& \textbf{0.8229} $\pm$ 0.1341& \textbf{0.3232}$\pm$ 0.1261\\

\textbf{CELM}& \cellcolor{highlightorange!35}\textbf{ 0.5695} $\pm$ 0.1702 & \cellcolor{highlightorange!35}\textbf{ 0.4145} $\pm$ 0.1639 & \cellcolor{highlightorange!35}\textbf{ 0.6408} $\pm$ 0.1494 & \cellcolor{highlightorange!35}\textbf{ 0.4805} $\pm$ 0.1603 & \cellcolor{highlightorange!35}\textbf{ 0.5757} $\pm$ 0.1624 & \cellcolor{highlightorange!35}\textbf{ 0.5772} $\pm$ 0.1613 & \cellcolor{highlightorange!35}\textbf{ 0.8755} $\pm$ 0.1521 & \cellcolor{highlightorange!35}\textbf{ 0.5597} $\pm$ 0.1728 \\






\bottomrule

\end{tabular}
\end{adjustbox}

\end{table}

%% file: TABLES/app_baseline_results_S0002_without_patient_history.tex
\begin{table}[ht]

\centering
\caption{Zero-context report generation performance on S0002.
The best results are highlighted in \textcolor{highlightorange}{orange}, and the best baseline is highlighted in \textcolor{highlightblue}{blue}.
}
\label{tab:app_baseline_results_S0002_without_PH}
\begin{adjustbox}{max width=\textwidth}

\setlength{\tabcolsep}{4pt} 
\renewcommand{\arraystretch}{1.2}
\begin{tabular}{lcccccccc}
\toprule

\textbf{Method}   & \textbf{BLEU-1}     &\textbf{BLEU-4}  &\textbf{ROUGE-1} &\textbf{ROUGE-2}  & \textbf{ROUGE-L}  & \textbf{ROUGE-LSUM}    & \textbf{BERTScore} &\textbf{METEOR}  \\
\midrule

& \multicolumn{8}{c}{\textbf{Unimodal + Text + EEG Features Input}} \\
\cmidrule(lr){2-9}

Gemma-3-4b-it& 0.0544 $\pm$ 0.0904& 0.0062 $\pm$ 0.0111& 0.0907 $\pm$ 0.1259& 0.0132 $\pm$ 0.0213& 0.0476 $\pm$ 0.0658& 0.0491 $\pm$ 0.0685& 0.2800 $\pm$ 0.3788& 0.0528 $\pm$ 0.0755\\
Medgemma-4b-it& 0.0423 $\pm$ 0.0627& 0.0052 $\pm$ 0.0074& 0.1256 $\pm$ 0.1004& 0.0142 $\pm$ 0.0171& 0.0744 $\pm$ 0.0580& 0.0754 $\pm$ 0.0588& 0.5090 $\pm$ 0.3670& 0.0668 $\pm$ 0.0563\\
Llama-3.2-3b-instruct& 0.1375 $\pm$ 0.1060& 0.0136 $\pm$ 0.0103& 0.2035 $\pm$ 0.0874& \cellcolor{highlightblue!35} 0.0285 $\pm$ 0.0199& 0.1105 $\pm$ 0.0441& 0.1142 $\pm$ 0.0475& \cellcolor{highlightblue!35} 0.7042 $\pm$ 0.2133& 0.1245 $\pm$ 0.0633\\
Qwen3-4b-instruct-2507& \cellcolor{highlightblue!35} 0.2211 $\pm$ 0.0834& \cellcolor{highlightblue!35} 0.0185 $\pm$ 0.0082& \cellcolor{highlightblue!35} 0.2307 $\pm$ 0.0838& 0.0274 $\pm$ 0.0166& \cellcolor{highlightblue!35} 0.1152 $\pm$ 0.0421& \cellcolor{highlightblue!35} 0.1178 $\pm$ 0.0451& 0.6919 $\pm$ 0.2319& \cellcolor{highlightblue!35} 0.1719 $\pm$ 0.0635\\

\midrule

& \multicolumn{8}{c}{\textbf{Multimodal + Linear Projector + CBraMod}} \\
\cmidrule(lr){2-9}

Gemma-3-4b-it& 0.2457 $\pm$ 0.0855& 0.0231 $\pm$ 0.0101& \cellcolor{highlightblue!35} 0.2942 $\pm$ 0.0621& 0.0458 $\pm$ 0.0178& \cellcolor{highlightblue!35} 0.1440 $\pm$ 0.0303& \cellcolor{highlightblue!35} 0.1475 $\pm$ 0.0343& 0.7767 $\pm$ 0.1330& 0.1901 $\pm$ 0.0508\\
Medgemma-4b-it& 0.0343 $\pm$ 0.0454& 0.0041 $\pm$ 0.0055& 0.1785 $\pm$ 0.1146& 0.0186 $\pm$ 0.0170& 0.1035 $\pm$ 0.0645& 0.1060 $\pm$ 0.0672& 0.5798 $\pm$ 0.3438& 0.0769 $\pm$ 0.0529\\
Llama-3.2-3B-instruct& 0.0762 $\pm$ 0.0935& 0.0081 $\pm$ 0.0099& 0.1339 $\pm$ 0.1223& 0.0164 $\pm$ 0.0197& 0.0749 $\pm$ 0.0677& 0.0769 $\pm$ 0.0702& 0.4451 $\pm$ 0.3843& 0.0748 $\pm$ 0.0708\\
Qwen3-4B-instruct-2507& \cellcolor{highlightblue!35} 0.2635 $\pm$ 0.0399& \cellcolor{highlightblue!35} 0.0278 $\pm$ 0.0081& 0.2918 $\pm$ 0.0311& \cellcolor{highlightblue!35} 0.0479 $\pm$ 0.0168& 0.1430 $\pm$ 0.0162& 0.1468 $\pm$ 0.0238& \cellcolor{highlightblue!35} 0.7827 $\pm$ 0.0136& \cellcolor{highlightblue!35} 0.2100 $\pm$ 0.0332\\

\midrule



\textbf{CELM-SCC}& \textbf{0.3097 $\pm$ 0.1511}& \textbf{0.1065 $\pm$ 0.0753}& \textbf{0.3864 $\pm$ 0.1199}& \textbf{0.1484 $\pm$ 0.0780}& \textbf{0.2499 $\pm$ 0.0957}& \textbf{0.2530 $\pm$ 0.0962}& \textbf{0.7940 $\pm$ 0.1658}& \textbf{0.2601 $\pm$ 0.1040}\\
\textbf{CELM}& \cellcolor{highlightorange!35}\textbf{ 0.4976} $\pm$ 0.2062 & \cellcolor{highlightorange!35}\textbf{ 0.3313} $\pm$ 0.1761 & \cellcolor{highlightorange!35}\textbf{ 0.5625} $\pm$ 0.2024 & \cellcolor{highlightorange!35}\textbf{ 0.3931} $\pm$ 0.1840 & \cellcolor{highlightorange!35}\textbf{ 0.4880} $\pm$ 0.2013 & \cellcolor{highlightorange!35}\textbf{ 0.4896} $\pm$ 0.2008 & \cellcolor{highlightorange!35}\textbf{ 0.8113} $\pm$ 0.2517 & \cellcolor{highlightorange!35}\textbf{ 0.4854} $\pm$ 0.2075 \\






\bottomrule

\end{tabular}
\end{adjustbox}

\end{table}

%% file: TABLES/main_eeg_projector_ablation.tex
\begin{table}[ht]

\centering
\caption{Alignment module ablation on S0002 dataset. The best results are highlighted in \textcolor{highlightorange}{orange}
}
\label{tab:app_results_projector_ablation}
\begin{adjustbox}{max width=\linewidth}

\setlength{\tabcolsep}{4pt} 
\renewcommand{\arraystretch}{1.0}
\begin{tabular}{lcccccccc}
\toprule

\textbf{Projector}   &  \textbf{BLEU-1}     &\textbf{BLEU-4}  &\textbf{ROUGE-1} &\textbf{ROUGE-2}  & \textbf{ROUGE-L}  & \textbf{ROUGE-LSUM}    & \textbf{BERTScore} &\textbf{METEOR} \\
\midrule

& \multicolumn{8}{c}{\textbf{EEG Only}} \\
\cmidrule(lr){2-9}

Linear Projector & 0.2644 $\pm$ 0.0471& 0.0247 $\pm$ 0.0063& 0.2928 $\pm$ 0.0314& 0.0447 $\pm$ 0.0141& 0.1392 $\pm$ 0.0161& 0.1392 $\pm$ 0.0161& 0.7828 $\pm$ 0.0152& 0.2072 $\pm$ 0.0382\\

Perceiver Projector & 0.2636 $\pm$ 0.0696& 0.0249 $\pm$ 0.0098& 0.2897 $\pm$ 0.0344& 0.0404 $\pm$ 0.0146& 0.1377 $\pm$ 0.0165& 0.1377 $\pm$ 0.0165& 0.7888 $\pm$ 0.0162& 0.1968 $\pm$ 0.0385\\

SCC Projector & 0.2991 $\pm$ 0.1276& 0.0912 $\pm$ 0.0590& 0.3793 $\pm$ 0.0892& 0.1378 $\pm$ 0.0666& 0.2441 $\pm$ 0.0771& 0.2441 $\pm$ 0.0771& \cellcolor{highlightorange!35}\textbf{ 0.8192} $\pm$ 0.0299& 0.2574 $\pm$ 0.0891\\

SCA Projector & \cellcolor{highlightorange!35}\textbf{ 0.4652} $\pm$ 0.1884& \cellcolor{highlightorange!35}\textbf{ 0.2666} $\pm$ 0.1354& \cellcolor{highlightorange!35}\textbf{ 0.5248} $\pm$ 0.1869& \cellcolor{highlightorange!35}\textbf{ 0.3271} $\pm$ 0.1494& \cellcolor{highlightorange!35}\textbf{ 0.4339} $\pm$ 0.1767& \cellcolor{highlightorange!35}\textbf{ 0.4339} $\pm$ 0.1767& 0.7990 $\pm$ 0.2432& \cellcolor{highlightorange!35}\textbf{ 0.4390} $\pm$ 0.1869\\

\midrule
& \multicolumn{8}{c}{\textbf{With Patient History}} \\
\cmidrule(lr){2-9}
Linear Projector & 0.2775 $\pm$ 0.0485& 0.0283 $\pm$ 0.0121& 0.2998 $\pm$ 0.0425& 0.0507 $\pm$ 0.0240& 0.1429 $\pm$ 0.0239& 0.1473 $\pm$ 0.0306& 0.7852 $\pm$ 0.0153& 0.2143 $\pm$ 0.0349\\

Perceiver Projector & 0.2672 $\pm$ 0.0556& 0.0273 $\pm$ 0.0192& 0.2879 $\pm$ 0.0378& 0.0461 $\pm$ 0.0300& 0.1384 $\pm$ 0.0281& 0.1422 $\pm$ 0.0321& 0.7878 $\pm$ 0.0159& 0.1955 $\pm$ 0.0393\\

SCC Projector & 0.3767 $\pm$ 0.1557& 0.1671 $\pm$ 0.1054& 0.4487 $\pm$ 0.1283& 0.2149 $\pm$ 0.1109& 0.3232 $\pm$ 0.1263& 0.3268 $\pm$ 0.1258& 0.8229 $\pm$ 0.1341& 0.3232 $\pm$ 0.1261\\

SCA Projector & \cellcolor{highlightorange!35}\textbf{ 0.5695} $\pm$ 0.1702& \cellcolor{highlightorange!35}\textbf{ 0.4145} $\pm$ 0.1639& \cellcolor{highlightorange!35}\textbf{ 0.6408} $\pm$ 0.1494& \cellcolor{highlightorange!35}\textbf{ 0.4805} $\pm$ 0.1603& \cellcolor{highlightorange!35}\textbf{ 0.5757} $\pm$ 0.1624& \cellcolor{highlightorange!35}\textbf{ 0.5772} $\pm$ 0.1613& \cellcolor{highlightorange!35}\textbf{ 0.8755} $\pm$ 0.1521& \cellcolor{highlightorange!35}\textbf{ 0.5597} $\pm$ 0.1728\\

\midrule
& \multicolumn{8}{c}{\textbf{All}} \\
\cmidrule(lr){2-9}
Linear Projector & 0.2762 $\pm$ 0.0485& 0.0279 $\pm$ 0.0117& 0.2991 $\pm$ 0.0416& 0.0501 $\pm$ 0.0233& 0.1426 $\pm$ 0.0232& 0.1465 $\pm$ 0.0296& 0.7849 $\pm$ 0.0153& 0.2135 $\pm$ 0.0353\\

Perceiver Projector & 0.2669 $\pm$ 0.0571& 0.0271 $\pm$ 0.0184& 0.2881 $\pm$ 0.0375& 0.0455 $\pm$ 0.0288& 0.1384 $\pm$ 0.0271& 0.1417 $\pm$ 0.0309& 0.7879 $\pm$ 0.0160& 0.1956 $\pm$ 0.0392\\

SCC Projector & 0.3689 $\pm$ 0.1548& 0.1594 $\pm$ 0.1042& 0.4417 $\pm$ 0.1267& 0.2071 $\pm$ 0.1098& 0.3152 $\pm$ 0.1246& 0.3185 $\pm$ 0.1243& 0.8226 $\pm$ 0.1275& 0.3166 $\pm$ 0.1245\\

SCA Projector & \cellcolor{highlightorange!35}\textbf{ 0.5590} $\pm$ 0.1749& \cellcolor{highlightorange!35}\textbf{ 0.3996} $\pm$ 0.1673& \cellcolor{highlightorange!35}\textbf{ 0.6292} $\pm$ 0.1574& \cellcolor{highlightorange!35}\textbf{ 0.4650} $\pm$ 0.1657& \cellcolor{highlightorange!35}\textbf{ 0.5614} $\pm$ 0.1692& \cellcolor{highlightorange!35}\textbf{ 0.5628} $\pm$ 0.1685& \cellcolor{highlightorange!35}\textbf{ 0.8678} $\pm$ 0.1650& \cellcolor{highlightorange!35}\textbf{ 0.5475} $\pm$ 0.1779\\

\bottomrule

\end{tabular}
\end{adjustbox}

\end{table}

%% file: TABLES/app_eeg_encoder_ablation.tex
\begin{table}[ht]

\centering
\caption{EEG encoder ablation study
}

\label{tab:app_eeg_encoder_ablation}
\begin{adjustbox}{max width=\textwidth}

\setlength{\tabcolsep}{4pt} 
\renewcommand{\arraystretch}{1.2}
\begin{tabular}{lcccccccc}
\toprule

\textbf{EEG Encoder}   & \textbf{BLEU-1}     &\textbf{BLEU-4}  &\textbf{ROUGE-1} &\textbf{ROUGE-2}  & \textbf{ROUGE-L}  & \textbf{ROUGE-LSUM}    & \textbf{BERTScore} &\textbf{METEOR}  \\
\midrule
 Labram& 0.4654 $\pm$ 0.2272& 0.3188 $\pm$ 0.1904& 0.5355 $\pm$ 0.2312& 0.3809 $\pm$ 0.2027& 0.4673 $\pm$ 0.2233& 0.4689 $\pm$ 0.2231& 0.7781 $\pm$ 0.2931& 0.4600 $\pm$ 0.2259\\
Cbramod& \cellcolor{highlightorange!35}\textbf{ 0.5590} $\pm$ 0.1749 & \cellcolor{highlightorange!35}\textbf{ 0.3996} $\pm$ 0.1673 & \cellcolor{highlightorange!35}\textbf{ 0.6292} $\pm$ 0.1574 & \cellcolor{highlightorange!35}\textbf{ 0.4650} $\pm$ 0.1657 & \cellcolor{highlightorange!35}\textbf{ 0.5614} $\pm$ 0.1692 & \cellcolor{highlightorange!35}\textbf{ 0.5628} $\pm$ 0.1685 & \cellcolor{highlightorange!35}\textbf{ 0.8678} $\pm$ 0.1650 & \cellcolor{highlightorange!35}\textbf{ 0.5475} $\pm$ 0.1779 \\






\bottomrule

\end{tabular}
\end{adjustbox}

\vspace{-0.5cm}
\end{table}

%% file: TABLES/app_section_wise_performace.tex
\begin{table}[ht]

\centering
\caption{Performance by report section on S0001 dataset
}
\label{tab:app_section_wise_performance}
\begin{adjustbox}{max width=\textwidth}

\setlength{\tabcolsep}{4pt} 
\renewcommand{\arraystretch}{1.2}
\begin{tabular}{llccccccc}
\toprule

\textbf{Report Section}& \textbf{Method}   & \textbf{BLEU-1}     &\textbf{BLEU-4}  &\textbf{ROUGE-1} &\textbf{ROUGE-2}  & \textbf{ROUGE-L}  & \textbf{ROUGE-LSUM}    &\textbf{METEOR}  \\
\midrule

\multirow{5}{*}{EEG Description/Details} & gemma-3n-E4B-it & 0.0095 $\pm$ 0.0344 & 0.0017 $\pm$ 0.0057 & 0.0396 $\pm$ 0.0604 & 0.0012 $\pm$ 0.0057 & 0.0256 $\pm$ 0.0389 & 0.0259 $\pm$ 0.0395  & 0.0228 $\pm$ 0.0385 \\
& medgemma-4b-it & 0.1624 $\pm$ 0.1729 & 0.0489 $\pm$ 0.1401 & 0.2108 $\pm$ 0.1917 & 0.0612 $\pm$ 0.1872 & 0.1460 $\pm$ 0.1858 & 0.1487 $\pm$ 0.1861  & 0.1535 $\pm$ 0.1831 \\
& Llama-3.1-8B-Instruct & 0.2045 $\pm$ 0.1452 & 0.0501 $\pm$ 0.1179 & 0.2143 $\pm$ 0.1691 & 0.0603 $\pm$ 0.1694 & 0.1454 $\pm$ 0.1687 & 0.1485 $\pm$ 0.1690  & 0.1838 $\pm$ 0.1645 \\
& Qwen3-4B-Instruct-2507 & 0.1818 $\pm$ 0.0988 & 0.0285 $\pm$ 0.0555 & 0.2059 $\pm$ 0.1190 & 0.0393 $\pm$ 0.0969 & 0.1271 $\pm$ 0.1082 & 0.1300 $\pm$ 0.1087  & 0.2069 $\pm$ 0.1080 \\
\cmidrule{2-9}
& \textbf{CELM-SCC} &\cellcolor{highlightorange!35}0.3381 $\pm$ 0.1900 & \cellcolor{highlightorange!35}0.1291 $\pm$ 0.1832 & \cellcolor{highlightorange!35}0.3826 $\pm$ 0.1854 & \cellcolor{highlightorange!35}0.1601 $\pm$ 0.1935 & \cellcolor{highlightorange!35}0.2650 $\pm$ 0.1841 & \cellcolor{highlightorange!35}0.2687 $\pm$ 0.1840  & \cellcolor{highlightorange!35}0.2875 $\pm$ 0.1833 \\
& \textbf{CELM} & \cellcolor{highlightorange!35}0.4907 $\pm$ 0.1834 & \cellcolor{highlightorange!35}0.2841 $\pm$ 0.1911 & \cellcolor{highlightorange!35}0.5631 $\pm$ 0.1607 & \cellcolor{highlightorange!35}0.3483 $\pm$ 0.1987 & \cellcolor{highlightorange!35}0.4727 $\pm$ 0.1813 & \cellcolor{highlightorange!35}0.4745 $\pm$ 0.1809  & \cellcolor{highlightorange!35}0.4798 $\pm$ 0.1873 \\
\midrule

\multirow{5}{*}{Epileptiform Abnormalities} & gemma-3n-E4B-it & 0.0212 $\pm$ 0.0508 & 0.0081 $\pm$ 0.0201 & 0.0256 $\pm$ 0.0642 & 0.0048 $\pm$ 0.0181 & 0.0212 $\pm$ 0.0529 & 0.0217 $\pm$ 0.0539  & 0.0193 $\pm$ 0.0563 \\
& medgemma-4b-it & 0.0015 $\pm$ 0.0117 & 0.0008 $\pm$ 0.0064 & 0.0043 $\pm$ 0.0330 & 0.0034 $\pm$ 0.0265 & 0.0043 $\pm$ 0.0330 & 0.0043 $\pm$ 0.0330  & 0.0030 $\pm$ 0.0234 \\
& Llama-3.1-8B-Instruct & 0.0786 $\pm$ 0.1185 & 0.0343 $\pm$ 0.0596 & 0.0954 $\pm$ 0.1437 & 0.0422 $\pm$ 0.0860 & 0.0858 $\pm$ 0.1337 & 0.0863 $\pm$ 0.1347  & 0.0992 $\pm$ 0.1523 \\
& Qwen3-4B-Instruct-2507 & 0.0574 $\pm$ 0.0919 & 0.0150 $\pm$ 0.0240 & 0.0772 $\pm$ 0.1201 & 0.0273 $\pm$ 0.0473 & 0.0568 $\pm$ 0.0808 & 0.0578 $\pm$ 0.0830  & 0.0759 $\pm$ 0.0942 \\
\cmidrule{2-9}
& \textbf{CELM-SCC} &\cellcolor{highlightorange!35}0.2870 $\pm$ 0.3051 & \cellcolor{highlightorange!35}0.1862 $\pm$ 0.2173 & \cellcolor{highlightorange!35}0.3392 $\pm$ 0.2614 & \cellcolor{highlightorange!35}0.2104 $\pm$ 0.2319 & \cellcolor{highlightorange!35}0.3233 $\pm$ 0.2597 & \cellcolor{highlightorange!35}0.3221 $\pm$ 0.2598  & \cellcolor{highlightorange!35}0.2717 $\pm$ 0.2560 \\
& \textbf{CELM} & \cellcolor{highlightorange!35}0.3172 $\pm$ 0.3472 & \cellcolor{highlightorange!35}0.2061 $\pm$ 0.2641 & \cellcolor{highlightorange!35}0.3881 $\pm$ 0.2967 & \cellcolor{highlightorange!35}0.2401 $\pm$ 0.2796 & \cellcolor{highlightorange!35}0.3600 $\pm$ 0.2902 & \cellcolor{highlightorange!35}0.3607 $\pm$ 0.2897  & \cellcolor{highlightorange!35}0.3222 $\pm$ 0.3107 \\
\midrule

\multirow{5}{*}{Background Activity} & gemma-3n-E4B-it & 0.0440 $\pm$ 0.0909 & 0.0058 $\pm$ 0.0113 & 0.0954 $\pm$ 0.1133 & 0.0071 $\pm$ 0.0175 & 0.0584 $\pm$ 0.0679 & 0.0584 $\pm$ 0.0679  & 0.0534 $\pm$ 0.0685 \\
& medgemma-4b-it & 0.0000 $\pm$ 0.0000 & 0.0000 $\pm$ 0.0000 & 0.0000 $\pm$ 0.0000 & 0.0000 $\pm$ 0.0000 & 0.0000 $\pm$ 0.0000 & 0.0000 $\pm$ 0.0000  & 0.0000 $\pm$ 0.0000 \\
& Llama-3.1-8B-Instruct & 0.0318 $\pm$ 0.0706 & 0.0047 $\pm$ 0.0102 & 0.1093 $\pm$ 0.1004 & 0.0102 $\pm$ 0.0181 & 0.0759 $\pm$ 0.0690 & 0.0759 $\pm$ 0.0690  & 0.0443 $\pm$ 0.0433 \\
& Qwen3-4B-Instruct-2507 & 0.2172 $\pm$ 0.1394 & 0.0302 $\pm$ 0.0202 & 0.1984 $\pm$ 0.1289 & 0.0396 $\pm$ 0.0306 & 0.1158 $\pm$ 0.0752 & 0.1158 $\pm$ 0.0752  & 0.1492 $\pm$ 0.0969 \\
\cmidrule{2-9}
& \textbf{CELM-SCC} &\cellcolor{highlightorange!35}0.1494 $\pm$ 0.1142 & \cellcolor{highlightorange!35}0.0266 $\pm$ 0.0281 & \cellcolor{highlightorange!35}0.2442 $\pm$ 0.0629 & \cellcolor{highlightorange!35}0.0441 $\pm$ 0.0252 & \cellcolor{highlightorange!35}0.1670 $\pm$ 0.0521 & \cellcolor{highlightorange!35}0.1670 $\pm$ 0.0521  & \cellcolor{highlightorange!35}0.1355 $\pm$ 0.0524 \\
& \textbf{CELM} & \cellcolor{highlightorange!35}0.1825 $\pm$ 0.1225 & \cellcolor{highlightorange!35}0.0332 $\pm$ 0.0244 & \cellcolor{highlightorange!35}0.2703 $\pm$ 0.0536 & \cellcolor{highlightorange!35}0.0701 $\pm$ 0.0476 & \cellcolor{highlightorange!35}0.1935 $\pm$ 0.0530 & \cellcolor{highlightorange!35}0.1935 $\pm$ 0.0530  & \cellcolor{highlightorange!35}0.1786 $\pm$ 0.0655 \\
\midrule

\multirow{5}{*}{Interictal Epileptiform Abnormalities} & gemma-3n-E4B-it & 0.0207 $\pm$ 0.0414 & 0.0096 $\pm$ 0.0192 & 0.0255 $\pm$ 0.0318 & 0.0075 $\pm$ 0.0150 & 0.0255 $\pm$ 0.0318 & 0.0255 $\pm$ 0.0318  & 0.0159 $\pm$ 0.0182 \\
& medgemma-4b-it & 0.0000 $\pm$ 0.0000 & 0.0000 $\pm$ 0.0000 & 0.0000 $\pm$ 0.0000 & 0.0000 $\pm$ 0.0000 & 0.0000 $\pm$ 0.0000 & 0.0000 $\pm$ 0.0000  & 0.0000 $\pm$ 0.0000 \\
& Llama-3.1-8B-Instruct & 0.0891 $\pm$ 0.0899 & 0.0125 $\pm$ 0.0126 & 0.1200 $\pm$ 0.1008 & 0.0228 $\pm$ 0.0221 & 0.0672 $\pm$ 0.0550 & 0.0708 $\pm$ 0.0581  & 0.0863 $\pm$ 0.0781 \\
& Qwen3-4B-Instruct-2507 & 0.1594 $\pm$ 0.0755 & 0.0251 $\pm$ 0.0092 & 0.1856 $\pm$ 0.1088 & 0.0334 $\pm$ 0.0218 & 0.0985 $\pm$ 0.0502 & 0.1048 $\pm$ 0.0539  & 0.1253 $\pm$ 0.0472 \\
\cmidrule{2-9}
& \textbf{CELM-SCC} &\cellcolor{highlightorange!35}0.0100 $\pm$ 0.0133 & \cellcolor{highlightorange!35}0.0027 $\pm$ 0.0039 & \cellcolor{highlightorange!35}0.0739 $\pm$ 0.0444 & \cellcolor{highlightorange!35}0.0228 $\pm$ 0.0140 & \cellcolor{highlightorange!35}0.0635 $\pm$ 0.0361 & \cellcolor{highlightorange!35}0.0666 $\pm$ 0.0338  & \cellcolor{highlightorange!35}0.0457 $\pm$ 0.0217 \\
& \textbf{CELM} & \cellcolor{highlightorange!35}0.0765 $\pm$ 0.0815 & \cellcolor{highlightorange!35}0.0135 $\pm$ 0.0129 & \cellcolor{highlightorange!35}0.1166 $\pm$ 0.1240 & \cellcolor{highlightorange!35}0.0236 $\pm$ 0.0232 & \cellcolor{highlightorange!35}0.0787 $\pm$ 0.0779 & \cellcolor{highlightorange!35}0.0787 $\pm$ 0.0779  & \cellcolor{highlightorange!35}0.0650 $\pm$ 0.0559 \\
\midrule

\multirow{5}{*}{Events/Seizures} & gemma-3n-E4B-it & 0.0316 $\pm$ 0.0632 & 0.0086 $\pm$ 0.0172 & 0.0497 $\pm$ 0.0696 & 0.0012 $\pm$ 0.0016 & 0.0286 $\pm$ 0.0341 & 0.0291 $\pm$ 0.0341  & 0.0364 $\pm$ 0.0610 \\
& medgemma-4b-it & 0.0000 $\pm$ 0.0000 & 0.0000 $\pm$ 0.0000 & 0.0000 $\pm$ 0.0000 & 0.0000 $\pm$ 0.0000 & 0.0000 $\pm$ 0.0000 & 0.0000 $\pm$ 0.0000  & 0.0000 $\pm$ 0.0000 \\
& Llama-3.1-8B-Instruct & 0.0448 $\pm$ 0.0551 & 0.0225 $\pm$ 0.0302 & 0.0544 $\pm$ 0.0495 & 0.0124 $\pm$ 0.0211 & 0.0528 $\pm$ 0.0497 & 0.0528 $\pm$ 0.0497  & 0.0212 $\pm$ 0.0201 \\
& Qwen3-4B-Instruct-2507 & 0.0511 $\pm$ 0.0465 & 0.0167 $\pm$ 0.0138 & 0.0955 $\pm$ 0.0374 & 0.0040 $\pm$ 0.0071 & 0.0749 $\pm$ 0.0315 & 0.0758 $\pm$ 0.0300  & 0.0677 $\pm$ 0.0648 \\
\cmidrule{2-9}
& \textbf{CELM-SCC} &\cellcolor{highlightorange!35}0.0481 $\pm$ 0.0674 & \cellcolor{highlightorange!35}0.0257 $\pm$ 0.0353 & \cellcolor{highlightorange!35}0.1617 $\pm$ 0.1375 & \cellcolor{highlightorange!35}0.0004 $\pm$ 0.0009 & \cellcolor{highlightorange!35}0.1571 $\pm$ 0.1414 & \cellcolor{highlightorange!35}0.1578 $\pm$ 0.1407  & \cellcolor{highlightorange!35}0.0290 $\pm$ 0.0309 \\
& \textbf{CELM} & \cellcolor{highlightorange!35}0.0110 $\pm$ 0.0220 & \cellcolor{highlightorange!35}0.0090 $\pm$ 0.0181 & \cellcolor{highlightorange!35}0.1606 $\pm$ 0.1248 & \cellcolor{highlightorange!35}0.0024 $\pm$ 0.0033 & \cellcolor{highlightorange!35}0.1512 $\pm$ 0.1322 & \cellcolor{highlightorange!35}0.1531 $\pm$ 0.1305  & \cellcolor{highlightorange!35}0.0288 $\pm$ 0.0119 \\
\midrule

\multirow{5}{*}{Seizures} & gemma-3n-E4B-it & 0.0168 $\pm$ 0.0352 & 0.0032 $\pm$ 0.0070 & 0.0405 $\pm$ 0.0588 & 0.0005 $\pm$ 0.0013 & 0.0256 $\pm$ 0.0343 & 0.0300 $\pm$ 0.0409  & 0.0187 $\pm$ 0.0273 \\
& medgemma-4b-it & 0.0000 $\pm$ 0.0000 & 0.0000 $\pm$ 0.0000 & 0.0000 $\pm$ 0.0000 & 0.0000 $\pm$ 0.0000 & 0.0000 $\pm$ 0.0000 & 0.0000 $\pm$ 0.0000  & 0.0000 $\pm$ 0.0000 \\
& Llama-3.1-8B-Instruct & 0.0273 $\pm$ 0.0489 & 0.0169 $\pm$ 0.0328 & 0.0294 $\pm$ 0.0432 & 0.0010 $\pm$ 0.0026 & 0.0279 $\pm$ 0.0427 & 0.0294 $\pm$ 0.0432  & 0.0286 $\pm$ 0.0542 \\
& Qwen3-4B-Instruct-2507 & 0.0388 $\pm$ 0.0558 & 0.0121 $\pm$ 0.0180 & 0.0790 $\pm$ 0.1025 & 0.0085 $\pm$ 0.0149 & 0.0502 $\pm$ 0.0552 & 0.0544 $\pm$ 0.0606  & 0.0464 $\pm$ 0.0543 \\
\cmidrule{2-9}
& \textbf{CELM-SCC} &\cellcolor{highlightorange!35}0.0631 $\pm$ 0.1319 & \cellcolor{highlightorange!35}0.0551 $\pm$ 0.1180 & \cellcolor{highlightorange!35}0.0773 $\pm$ 0.1057 & \cellcolor{highlightorange!35}0.0000 $\pm$ 0.0000 & \cellcolor{highlightorange!35}0.0725 $\pm$ 0.1075 & \cellcolor{highlightorange!35}0.0735 $\pm$ 0.1072  & \cellcolor{highlightorange!35}0.0336 $\pm$ 0.0502 \\
& \textbf{CELM} & \cellcolor{highlightorange!35}0.0272 $\pm$ 0.0489 & \cellcolor{highlightorange!35}0.0253 $\pm$ 0.0471 & \cellcolor{highlightorange!35}0.1390 $\pm$ 0.2061 & \cellcolor{highlightorange!35}0.0005 $\pm$ 0.0012 & \cellcolor{highlightorange!35}0.1285 $\pm$ 0.2088 & \cellcolor{highlightorange!35}0.1329 $\pm$ 0.2081  & \cellcolor{highlightorange!35}0.0419 $\pm$ 0.0545 \\
\midrule

\multirow{5}{*}{Impression/Interpretation} & gemma-3n-E4B-it & 0.0172 $\pm$ 0.0725 & 0.0108 $\pm$ 0.0622 & 0.0237 $\pm$ 0.0949 & 0.0125 $\pm$ 0.0876 & 0.0207 $\pm$ 0.0923 & 0.0207 $\pm$ 0.0923  & 0.0193 $\pm$ 0.0990 \\
& medgemma-4b-it & 0.0000 $\pm$ 0.0000 & 0.0000 $\pm$ 0.0000 & 0.0000 $\pm$ 0.0000 & 0.0000 $\pm$ 0.0000 & 0.0000 $\pm$ 0.0000 & 0.0000 $\pm$ 0.0000  & 0.0000 $\pm$ 0.0000 \\
& Llama-3.1-8B-Instruct & 0.0809 $\pm$ 0.1339 & 0.0294 $\pm$ 0.0774 & 0.1148 $\pm$ 0.1863 & 0.0489 $\pm$ 0.1334 & 0.0883 $\pm$ 0.1654 & 0.0895 $\pm$ 0.1663  & 0.0876 $\pm$ 0.1499 \\
& Qwen3-4B-Instruct-2507 & 0.0885 $\pm$ 0.1173 & 0.0201 $\pm$ 0.0328 & 0.1069 $\pm$ 0.1399 & 0.0308 $\pm$ 0.0570 & 0.0675 $\pm$ 0.0910 & 0.0693 $\pm$ 0.0945  & 0.0996 $\pm$ 0.1172 \\
\cmidrule{2-9}
& \textbf{CELM-SCC} &\cellcolor{highlightorange!35}0.2194 $\pm$ 0.2899 & \cellcolor{highlightorange!35}0.1465 $\pm$ 0.2993 & \cellcolor{highlightorange!35}0.2687 $\pm$ 0.2946 & \cellcolor{highlightorange!35}0.1531 $\pm$ 0.3116 & \cellcolor{highlightorange!35}0.2190 $\pm$ 0.2929 & \cellcolor{highlightorange!35}0.2215 $\pm$ 0.2929  & \cellcolor{highlightorange!35}0.2206 $\pm$ 0.2996 \\
& \textbf{CELM} & \cellcolor{highlightorange!35}0.2493 $\pm$ 0.3481 & \cellcolor{highlightorange!35}0.2103 $\pm$ 0.3515 & \cellcolor{highlightorange!35}0.3581 $\pm$ 0.3381 & \cellcolor{highlightorange!35}0.2473 $\pm$ 0.3793 & \cellcolor{highlightorange!35}0.3358 $\pm$ 0.3434 & \cellcolor{highlightorange!35}0.3372 $\pm$ 0.3428  & \cellcolor{highlightorange!35}0.2829 $\pm$ 0.3428 \\




\bottomrule

\end{tabular}
\end{adjustbox}

\end{table}

%% file: TABLES/cross_site_generalization_experiment.tex
\begin{table}[ht]
\centering
\caption{Cross-site generalization of \method{} on S0001. We compare a model trained and
tested on S0001 (in-site) with a model trained on S0002 and tested on S0001 (cross-site).
We report mean $\pm$ standard deviation.}
\label{tab:cross_site}
\begin{adjustbox}{max width=\linewidth}
\setlength{\tabcolsep}{4pt}
\renewcommand{\arraystretch}{1.0}
\begin{tabular}{lcccc}
\toprule
\textbf{Setting} & \textbf{BLEU-1} & \textbf{ROUGE-1} & \textbf{METEOR} & \textbf{BERTScore} \\
\midrule
Train and test on S0001 & \cellcolor{highlightorange!35}\textbf{0.4823} $\pm$ 0.1920 & \cellcolor{highlightorange!35}\textbf{0.5565} $\pm$ 0.1683 & \cellcolor{highlightorange!35}\textbf{0.4734} $\pm$ 0.1941 & \cellcolor{highlightorange!35}\textbf{0.8697} $\pm$ 0.0901 \\
Train on S0002, test on S0001 & 0.3411 $\pm$ 0.1646 & 0.3585 $\pm$ 0.1673 & 0.3215 $\pm$ 0.1712 & 0.8015 $\pm$ 0.1584 \\
\bottomrule
\end{tabular}
\end{adjustbox}
\end{table}

%% file: checklist.tex
\clearpage
\section*{NeurIPS Paper Checklist}

\begin{enumerate}

\item {\bf Claims}
    \item[] Question: Do the main claims made in the abstract and introduction accurately reflect the paper's contributions and scope?
    \item[] Answer: \answerYes{} 
    \item[] Justification: The claims and contributions mentioned in the abstract and introduction accurately reflect the paper's contribution and scope. 

\item {\bf Limitations}
    \item[] Question: Does the paper discuss the limitations of the work performed by the authors?
    \item[] Answer: \answerYes{} 
    \item[] Justification: Limitations and implications of our work are discussed under Section~\ref{sec:discussion}.

\item {\bf Theory assumptions and proofs}
    \item[] Question: For each theoretical result, does the paper provide the full set of assumptions and a complete (and correct) proof?
    \item[] Answer: \answerNA{} 
    \item[] Justification: The paper does not include theoretical proofs. 

    \item {\bf Experimental result reproducibility}
    \item[] Question: Does the paper fully disclose all the information needed to reproduce the main experimental results of the paper to the extent that it affects the main claims and/or conclusions of the paper (regardless of whether the code and data are provided or not)?
    \item[] Answer: \answerYes{} 
    \item[] Justification: All the necessary details to reproduce our work are provided in the main text, along with extended details in the Appendix. Additionally, we provide our code for the proposed model and benchmark construction pipeline at \url{https://github.com/Jathurshan0330/CELM}. The data utilized for this work is publicly available at \url{https://bdsp.io/content/harvard-eeg-db/4.1/}, which can be accessed after completing the necessary steps to obtain credentialed access.

\item {\bf Open access to data and code}
    \item[] Question: Does the paper provide open access to the data and code, with sufficient instructions to faithfully reproduce the main experimental results, as described in supplemental material?
    \item[] Answer: \answerYes{}
    \item[] Justification:  We provide our code for the proposed model and benchmark construction pipeline at \url{https://github.com/Jathurshan0330/CELM} for reproducibility. The data utilized for this work is publicly available at \url{https://bdsp.io/content/harvard-eeg-db/4.1/}, which can be accessed after completing the necessary steps to obtain credentialed access.

\item {\bf Experimental setting/details}
    \item[] Question: Does the paper specify all the training and test details (e.g., data splits, hyperparameters, how they were chosen, type of optimizer) necessary to understand the results?
    \item[] Answer: \answerYes{} 
    \item[] Justification: Provided in the main text and appendix.

\item {\bf Experiment statistical significance}
    \item[] Question: Does the paper report error bars suitably and correctly defined or other appropriate information about the statistical significance of the experiments?
    \item[] Answer: \answerYes{} 
    \item[] Justification:  We report standard deviations and error bars where relevant, and assess statistical significance with paired Wilcoxon signed-rank tests.

\item {\bf Experiments compute resources}
    \item[] Question: For each experiment, does the paper provide sufficient information on the computer resources (type of compute workers, memory, time of execution) needed to reproduce the experiments?
    \item[] Answer: \answerYes 
    \item[] Justification: Provided in Appendix 
    
\item {\bf Code of ethics}
    \item[] Question: Does the research conducted in the paper conform, in every respect, with the NeurIPS Code of Ethics \url{https://neurips.cc/public/EthicsGuidelines}?
    \item[] Answer: \answerYes{} 
    \item[] Justification: Yes, it follows the NeurIPS Code of Ethics.

\item {\bf Broader impacts}
    \item[] Question: Does the paper discuss both potential positive societal impacts and negative societal impacts of the work performed?
    \item[] Answer: \answerYes{} 
    \item[] Justification: Included in Section~\ref{sec:discussion}
    
\item {\bf Safeguards}
    \item[] Question: Does the paper describe safeguards that have been put in place for responsible release of data or models that have a high risk for misuse (e.g., pre-trained language models, image generators, or scraped datasets)?
    \item[] Answer: \answerYes{}
    \item[] Justification: We release only code; no clinical data or derived reports are redistributed. Model checkpoints are not publicly released, and access will be gated under the BDSP Credentialed Health Data License 1.5.0 in the future. We provide a model card (Appendix~\ref{app:modelcard}) documenting intended use, out-of-scope uses, and known failure modes, and we state that \method{} is for research use only (Section~\ref{sec:discussion}). All data processing used locally hosted LLMs, as required by the data use agreement.

\item {\bf Licenses for existing assets}
    \item[] Question: Are the creators or original owners of assets (e.g., code, data, models), used in the paper, properly credited and are the license and terms of use explicitly mentioned and properly respected?
    \item[] Answer: \answerYes{} 
    \item[] Justification: The creators of the dataset used in this study are cited. 

\item {\bf New assets}
    \item[] Question: Are new assets introduced in the paper well documented and is the documentation provided alongside the assets?
    \item[] Answer: \answerYes{} 
    \item[] Justification: We release the benchmark construction code and the model code at
\url{https://github.com/Jathurshan0330/CELM} under the MIT License, with documentation and a model
card (Appendix~\ref{app:modelcard}).

\item {\bf Crowdsourcing and research with human subjects}
    \item[] Question: For crowdsourcing experiments and research with human subjects, does the paper include the full text of instructions given to participants and screenshots, if applicable, as well as details about compensation (if any)? 
    \item[] Answer: \answerYes{} 
    \item[] Justification: The paper does not include crowdsourcing or human-subjects research
on participants. It does include an expert evaluation of model outputs
(Section~\ref{sec:human_expert_eval_study}). Appendix~\ref{app:expert} gives the full task instructions, survey design, evaluator background and recruitment. Participation was
voluntary and uncompensated.

\item {\bf Institutional review board (IRB) approvals or equivalent for research with human subjects}
    \item[] Question: Does the paper describe potential risks incurred by study participants, whether such risks were disclosed to the subjects, and whether Institutional Review Board (IRB) approvals (or an equivalent approval/review based on the requirements of your country or institution) were obtained?
    \item[] Answer: \answerNA{} 
    \item[] Justification: No new data were collected from patients. We perform a
retrospective analysis and modeling of a de-identified, credentialed-access database under its data use agreement (Ethics Statement).

\item {\bf Declaration of LLM usage}
    \item[] Question: Does the paper describe the usage of LLMs if it is an important, original, or non-standard component of the core methods in this research? Note that if the LLM is used only for writing, editing, or formatting purposes and does \emph{not} impact the core methodology, scientific rigor, or originality of the research, declaration is not required.
    \item[] Answer: \answerYes{} 
    \item[] Justification: The proposed method utilizes local LLMs in the proposed framework. Additionally, LLMs were used to refine and correct grammar in the writing.

\end{enumerate}